\documentclass[journal]{IEEEtran}

\usepackage[all]{xy}
\usepackage{subfig}
\usepackage[table]{xcolor}
\usepackage{amsfonts}
\usepackage{amsmath}
\usepackage{amssymb}
\usepackage{array}
\usepackage{cite}
\usepackage{color}
\usepackage{colortbl}
\usepackage{float}
\usepackage{flushend}
\usepackage{graphicx}
\usepackage{hyperref}
\usepackage{inputenc}
\usepackage{lscape}
\usepackage{multirow}
\usepackage{rotating}
\usepackage{soul}
\usepackage{subfig}
\usepackage{times}
\usepackage{url}
\usepackage{wrapfig}
\usepackage{xspace}

\DeclareGraphicsExtensions{.pdf,.png}

\def\Real{\mathbb{R}}

\def\X{\ensuremath{\mathbf{X}}}

\def\Y{\ensuremath{\mathbf{Y}}}

\def\x{\ensuremath{\mathbf{x}}}
\def\y{\ensuremath{\mathbf{y}}}

\newcommand{\blue}[1]{\textcolor[rgb]{0,0,1}{#1}}           

\hypersetup{
    bookmarks=true,         
    unicode=false,          
    pdftoolbar=true,        
    pdfmenubar=true,        
    pdffitwindow=false,     
    pdfstartview={FitH},    
    pdftitle={RBIG},    
    pdfauthor={Jose A. Padron},     
    pdfsubject={},   
    pdfcreator={JAP}   
    pdfproducer={JAP}, 
    pdfkeywords={}, 
    pdfnewwindow=true,      
    colorlinks=true,       
    linkcolor={blue},          
    citecolor=blue,        
    filecolor=blue,      
    urlcolor=blue           
}

\title{Unsupervised Anomaly and Change Detection with Multivariate Gaussianization}

\author{Jos\'e A. Padr\'on-Hidalgo, Valero Laparra, and Gustau Camps-Valls,~\IEEEmembership{Fellow,~IEEE}

\thanks{{\bf \Large Paper published in IEEE Transactions on Geoscience and Remote Sensing, vol. 60, pp. 1-10, 2022, Art no. 5513010, doi: 10.1109/TGRS.2021.3116186.}}
\thanks{
\indent Image Processing Laboratory (IPL) \newline
Universitat de Val\`encia, Catedr\'atico A. Escardino - 46980 Paterna, Val\`encia (iSpain). E-mail: {gustau.camps@uv.es}
}\thanks{Research funded by the European Research Council (ERC) under the ERC-CoG-2014 SEDAL project (grant agreement 647423) and the Spanish Ministry of Economy, Industry and Competitiveness under the `Network of Excellence' program (grant code TEC2016-81900-REDT) and the projects TEC2016-77741-R, DPI2017-89867-C2-2-R, RTI2018-096765-A-I00 and PID2019-109026RB-I00. 
Jose A. Padr\'on was supported by the Grisolia grant from Generalitat Valenciana (GVA) with code GRISOLIA/2016/100.}
}

\begin{document}
\maketitle
 
\begin{abstract}
Anomaly detection is a field of intense research in remote sensing image processing. Identifying low probability events in remote sensing images is a challenging problem given the high-dimensionality of the data,  especially when no (or little) information about the anomaly is available a priori. While plenty of methods are available, the vast majority of them do not scale well to large datasets and require the  choice of some (very often critical) hyperparameters. Therefore, unsupervised and computationally efficient detection methods become strictly necessary, especially now with the data deluge problem. In this paper, we propose an unsupervised method for detecting anomalies and changes in remote sensing images by means of a multivariate Gaussianization methodology that allows to estimate multivariate densities accurately, a long-standing problem in statistics and machine learning. The methodology transforms arbitrarily complex multivariate data into a multivariate Gaussian distribution. Since the transformation is differentiable, by applying the change of variables formula one can estimate the probability at any point of the original domain. The assumption is straightforward: pixels with low estimated probability are considered anomalies. Our method can describe any multivariate distribution, makes an efficient use of memory and computational resources, and is parameter-free. We show the efficiency of the method in experiments involving both anomaly detection and change detection in different remote sensing image sets. For anomaly detection we propose two approaches. The first using directly the Gaussianization transform and the second using an hybrid model that combines Gaussianization and the Reed-Xiaoli (RX) method typically used in anomaly detection. {For change detection we take advantage of the Gaussianization transform and attribute the change to pixels with low probability compared to the first image, instead of those with high difference value typically employed in remote sensing.}    
Results show that our approach outperforms other linear and nonlinear methods in terms of detection power in both anomaly and change detection scenarios, showing robustness and scalability to dimensionality and sample sizes.
\end{abstract}


\maketitle

\section{Introduction}
Remote Sensing (RS) has become a powerful tool to develop applications for Earth monitoring ~\cite{Camps2014, Plaza, Camps2011}. Earth observation (EO) satellite missions, such as Sentinels-2 and Landsat-8 are able to replace the hard and costly work of the man on the ground. Also, the use of very high resolution (VHR) satellite imagery (e.g. QuickBird and the Worldview constellation) is becoming increasingly important for remote sensing applications, and it makes possible the detection of dangerous events such as extreme precipitations, heat waves, latent fires, droughts, floods or urbanization. The vast amount of data available from different sensors makes it urgent to have automatic methods to detect these events. A good and quite standard approach nowadays  to tackle this problem considers statistical models that allow us to detect anomalies and changes on the Earth cover. 

\begin{figure*}[t!]
\begin{center}
\includegraphics[width=18cm]{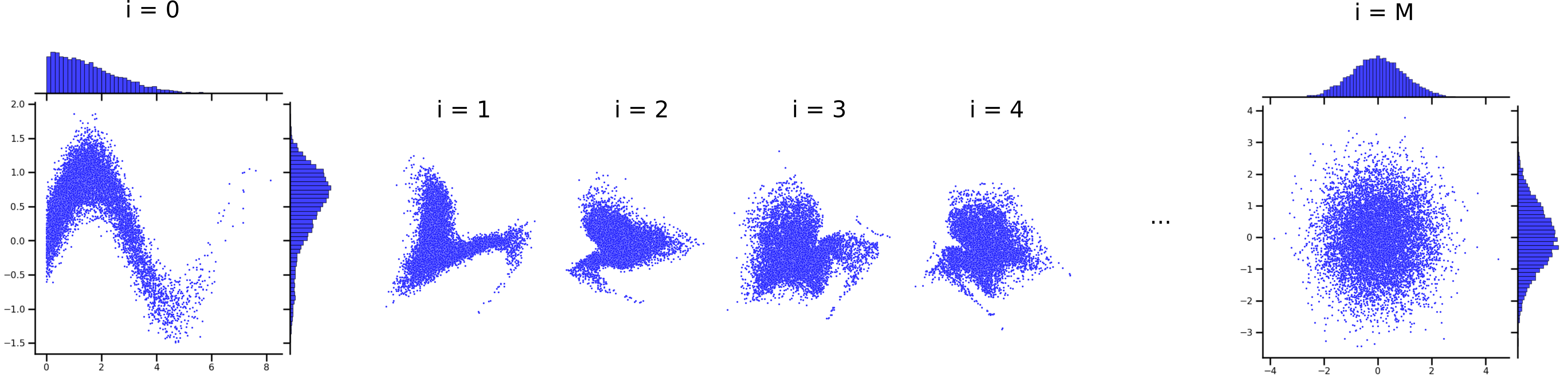}
\end{center}
\caption{{Illustration of the RBIG procedure on toy data. Original data ($i=0$) is transformed by using multiple iterations ($i=1,2,3,4...$). After enough iterations ($i=M$) the data follows a Gaussian distribution, in this particular case $M=150$. This example {has been} performed using the toolbox that can be found in} \cite{RBIG4AD_web}.}
\label{fig:fig_RBIG}
\end{figure*}

Statistical methods for anomaly detection (AD) focus on detecting small portions of the image which do not belong to the background of the scene~\cite{AD2}. Anomalies are considered a group of weird (low probability) pixels which significantly differ from their neighbors. AD is a challenging task and many variants have been proposed in the literature, such as neighbor based, clustering, classification, etc \cite{Matteoli2010,Salem,SahIn_AD,Liu2019,Qu2,Wang1}. 
However, among all of them, the Reed-Xiaoli (RX) approach \cite{Reed90tassp} is still the most widely used method for AD since the Gaussian distribution assumption is a reasonable approach in several cases, it is unsupervised, fast and easy to implement. The RX method allows us to detect the anomalous samples compared to the background using the well-known Mahalanobis distance. {RX can be used as either a {\em local}  or {\em global} approach~\cite{Qu}. In this study, all the methods are used in a global setting for the sake of simplicity. However, the proposed method is general enough and could be easily adapted to both local and dual-window approaches.} Since the Gaussian assumption is not flexible enough in most cases, variants of the RX has been developed to cope with higher-order feature relations. One option which obtains good results is based on the theory of reproducing kernels in Hilbert spaces, which extend the RX approach to the kernel RX  (KRX)~\cite{Kwon_KRX,Padron_RX,Zhou1}. However, the KRX algorithm has not been widely adopted in practice because, being a kernel method, the memory and computational cost increase with the number of pixels cubically and quadratically respectively, and more importantly the selection of the kernel parameters is critical to achieve a good performance. While unsupervised approaches to fit the kernel parameter exist, they achieve a sub-optimal performance and hence supervised approaches have to be used. 
In this manuscript we propose to use a different, more straightforward approach to the problem of anomaly detection based on multivariate Gaussianization transformation. {The proposed method is able to provide an estimation of the probability of multidimensional data. The basic idea is simple: the method finds an invertible and differentiable transformation that converts the multivariate distribution of the input data into a multivariate Gaussian and applies the change of variable formula under transformations to estimate the density in the input domain. Therefore, the probability of each data point in the original domain can be estimated by using two ingredients: the probability of the data in the transform domain and the determinant of the Jacobian of the transformation applied to the point. It is important to stress the fact that both ingredients are needed to estimate the density in the input domain. That is, the method is not transforming the data to a Gaussian domain to estimate the probability there, but uses the Gaussian domain as a convenient intermediate stage to estimate the density in the original one. Actually, the probability in the Gaussian domain must be multiplied by the determinant of the Jacobian. As we will see, the operations involved to estimate these  transformations are easy and fast, the method scales well with data dimensionality and does not require additional information to fit any parameter~\cite{rbig,Johnson20grsm}. {The parameters required for RBIG computation are the ones to fit the marginal density estimation. In our case, we use the simplest case based on computing histograms with the default parameters in the toolbox (see \cite{RBIG4AD_web}) and no extra fitting is used in any case.}
Note that the proposed method estimates the data density using all available data. In the AD setting, a naive application would incidentally use the anomalies too, which is not desirable. To address this issue, we propose a two-step procedure that finds a trade-off between the Gaussian assumption (very rigid) aimed to discard the potential anomalies, and the proposed method (very flexible) aimed to better estimate the background density. This two-step approach improves the results of the two components when used separately, and yields more robust density estimates, which ultimately leads to a powerful, automatic, unsupervised, algorithm for anomaly detection.} 

{Detecting changes in images automatically is extremely important because it allows us to improve predictions and our understanding of events. The usual approach to change detection (CD) in remote sensing is based on arithmetic operations over the image before and after the possible change \cite{Lu2004,Liu2019}. The most usual approach to detect changes is based on thresholding or classifying the difference image. This approach relies on previous (and quite critical) stages like atmospheric correction or image co-registration. Here we instead illustrate our method on the CD approach of finding {\em statistical} (probabilistic) differences between the signals before and after the event \cite{vandenburg2020evaluation}.
Other settings for change detection with RBIG could be defined by exploiting differences, ratios or other convenient transformations of the pair of images. Our approach can be considered as particular case of the anomaly detection problem, where the change class is the target class to be detected.} 

As for AD, a similar problem occurs when using statistical methods for CD~\cite{Lu04,Wu}: one aims to learn the distribution of the original image and analyze the statistical differences of the pixels in the new incoming image.  
Likewise, the RX and KRX extensions have been proposed to deal with CD problems. However, they show the same drawbacks as in AD. Here we describe how the multivariate Gaussianization can be used also in CD problems. Note that in both cases, AD and CD, the proposed statistical method is used to evaluate the pixel's probability so that one can classify them as anomalies or changes, respectively.

The remainder of the manuscript is organized as follow. Section \ref{sec:rbig} summarizes the Gaussianization transformation in general, and how to adapt it to anomaly and change detection. In section \ref{sec:experiments}, we illustrate its performance in three experiments, involving simulated anomalies, \blue{real} AD and CD examples with a database of multi- and hyperspectral images. Results show that the proposed approach is robust and flexible enough to be applied in different AD and CD scenarios, and obtains better performance than other simple and robust methods (like the RX) and more flexible and adaptable ones, like the KRX. Section \ref{sec:conclusions} concludes the paper with some remarks and further work.

\section{Multivariate Gaussianization for Detection}\label{sec:rbig}

{ Here we use the rotation-based iterative Gaussianization (RBIG), a nonparametric method for density estimation of multivariate distributions proposed in~\cite{rbig,Johnson20grsm}. RBIG is rooted in the idea of Gaussianization, introduced in the seminal work by~\cite{Friedman87} and further developed in~\cite{Chen00},} which consists of seeking for a transformation $G_x$ that converts a multivariate dataset $\X\in\Real^{\ell\times d}$ in domain $X$ to a domain where the mapped data $\Y\in\Real^{\ell\times d}$ follows a multivariate normal distribution in domain $Y$, i.e. $p_Y(\y) \sim \mathcal{N}({0},{I})$: 

\begin{eqnarray}
\begin{array}{llll}\label{Gauss}
      G_{x}: & \x\in \Real^d & \mapsto & \y\in \Real^d \\ 
      &   \sim p_X(\x) & & p_Y(\y) \sim {\mathcal N}({0},{I}_d),
\end{array}
\end{eqnarray}
where inputs and mapped data points have the same dimensionality $\x,\y\in\Real^d$, ${0}$ is a vector of zeros (for the means) and ${I}_d$ is the identity matrix for the covariance of dimension $d$.
Using the change of variable formula one can estimate the probability of a point $\x$ in the original domain: 
\begin{equation}
p_X(\x) = p_Y(\y) |J_{G_x}(\y)|,
\label{eq:change_variables}
\end{equation}
{where $p_X(\x)$ is the probability distribution of the original data point $\x$, $p_Y(\y)$ is the probability of the data point transformed to the Gaussian domain (note that by definition $p_Y\sim {\mathcal N}({0},{I}_d)$ and can be easily computed), and $|J_{G_x}(\y)|$ is the determinant of the Jacobian of the transformation $G_x$ in the point $\y$.}

{It is important to stress the important role of the term $|J_{G_x}(\y)|$: Note that one would be tempted to see the Gaussianization as a sort of pre-processing step that converts data to a convenient (Gaussian) domain where one could then  apply methods that assume Gaussian distribution. This is, however, incorrect since the term $|J_{G_x}(\y)|$ has to be included in the density estimation, and it is computed for each data point. For this formula to work, $G_x$ has to be differentiable, i.e. the $|J_{G_x}(\y)|>0, \forall \y$. The Gaussianization method we propose in this paper, RBIG, obtains a transformation $G_x$ that fulfills this property, cf. \cite{rbig}. Therefore, RBIG can be easily applied to estimate the probability of data points in the original domain, $p_X(\x)$.}

RBIG is an iterative algorithm, where in each iteration, $n$, two steps are applied: 1) a set of $d$ marginal Gaussianizations to each of the variables, $\boldsymbol{\Psi} = [\Phi_1,\ldots,\Phi_d]$, and 2) a linear rotation, ${R}\in\Real^{d\times d}$: 
\begin{equation}
\x[i+1] = {R}[i] \cdot \boldsymbol{\Psi}[i](\x[i]),~~~i=1,\ldots,M
\label{rbig_iter}
\end{equation}
where $M$ is the number of steps (iterations) in the sequence, $i=1,\ldots,M$. The final transformation $G_x$ is the composition of all performed transformations through iterations. {This procedure is illustrated with a toy example in Fig.~\ref{fig:fig_RBIG}. In \cite{rbig}
we showed that with enough iterations $M$ the method converges and the transformed data ultimately follows a standardized Gaussian, i.e. $p_Y(\y) \sim \mathcal{N}({0},{I}_d)$, taking $\y = \x[M]$.}  

{The computational load can be divided on two steps: the first one when the method is trained on a representative training set, and the second when is applied to unseen test dataset. On the one hand, the training time will depend on the the number of iterations and the employed type of linear transformation (see \cite{Laparra11tnnls} for a discussion on the different options). A reasonable number of iterations cannot be set before hand since it will be depend on the non-gaussianity of the data. However, it can be automatically found with the stopping criteria proposed in \cite{Laparra11tnnls}. On the other hand, when applying the method, the computation of the Jacobian will depend only on the number of iterations. It is important to note that, since the linear transformations are restricted to be rotations, the determinant of the Jacobian for them is always equal to $1$. Therefore the only part to be computed is the Jacobian of the marginal transformation, which is equal to the probability of the data before applying the transformation \cite{Laparra11tnnls}. Therefore, the application of the method is in principle computationally easier than the training step. However, note that if one wants to obtain the full Jacobian, not only the determinant as it is needed to compute the probability, the amount of data to be store is huge which can make the method difficult to apply. The use of RBIG in the framework of Normalizing Flows or density destructors can be useful in this sense since it is devoted to find transformation with easily computable Jacobians \cite{Johnson_2020_DD}. An implementation of RBIG in this framework can be found in \cite{RBIG4AD_web}.}

\begin{figure}[t!]
\setlength{\tabcolsep}{2pt}
\begin{center}
\begin{tabular}{cc}
Real Image & Original Data \\
  \\[0.1cm]
\includegraphics[trim=0.5cm 0.5cm 0.5cm 2cm, width = 4cm, height= 3.2cm]{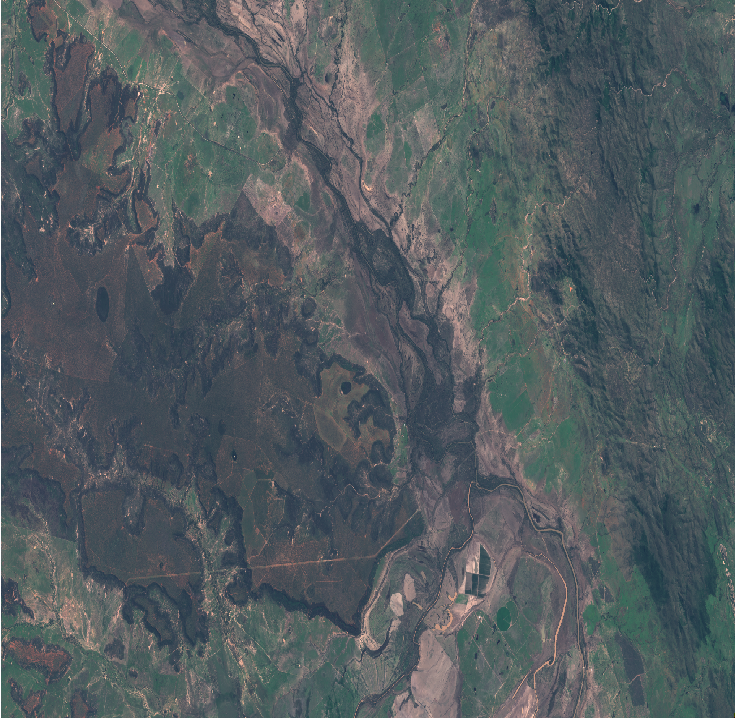}&
\includegraphics[trim=0.5cm 1.6cm 0.2cm 0.5cm, width=5cm, height= 4cm]{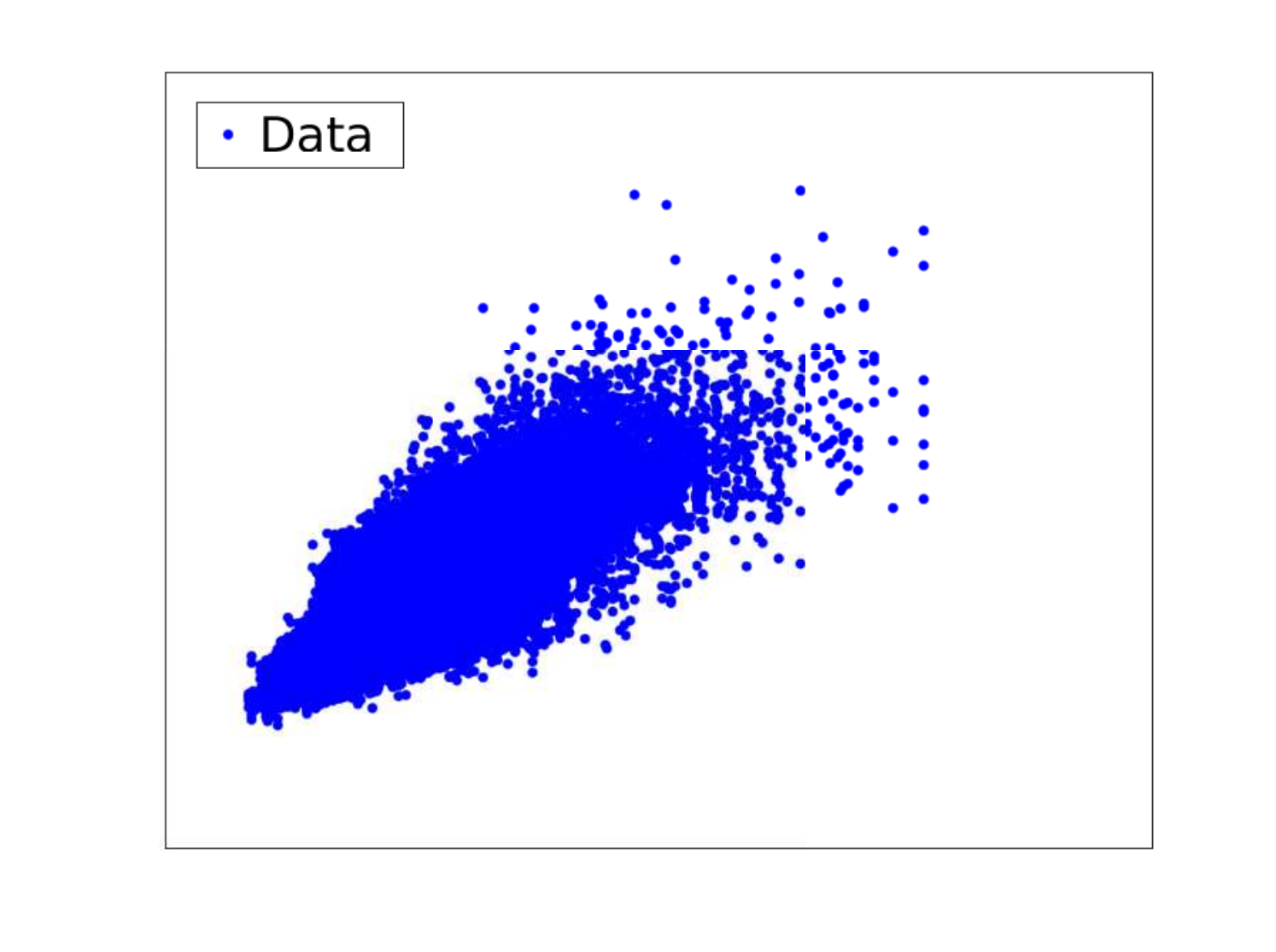} \\[0.1cm]\\
Gaussianized data & Synthesized data\\
[0.1cm]\\
\includegraphics[trim=0.0cm 0.4cm 0.0cm 0.0cm, width=4.2cm, height= 4cm]{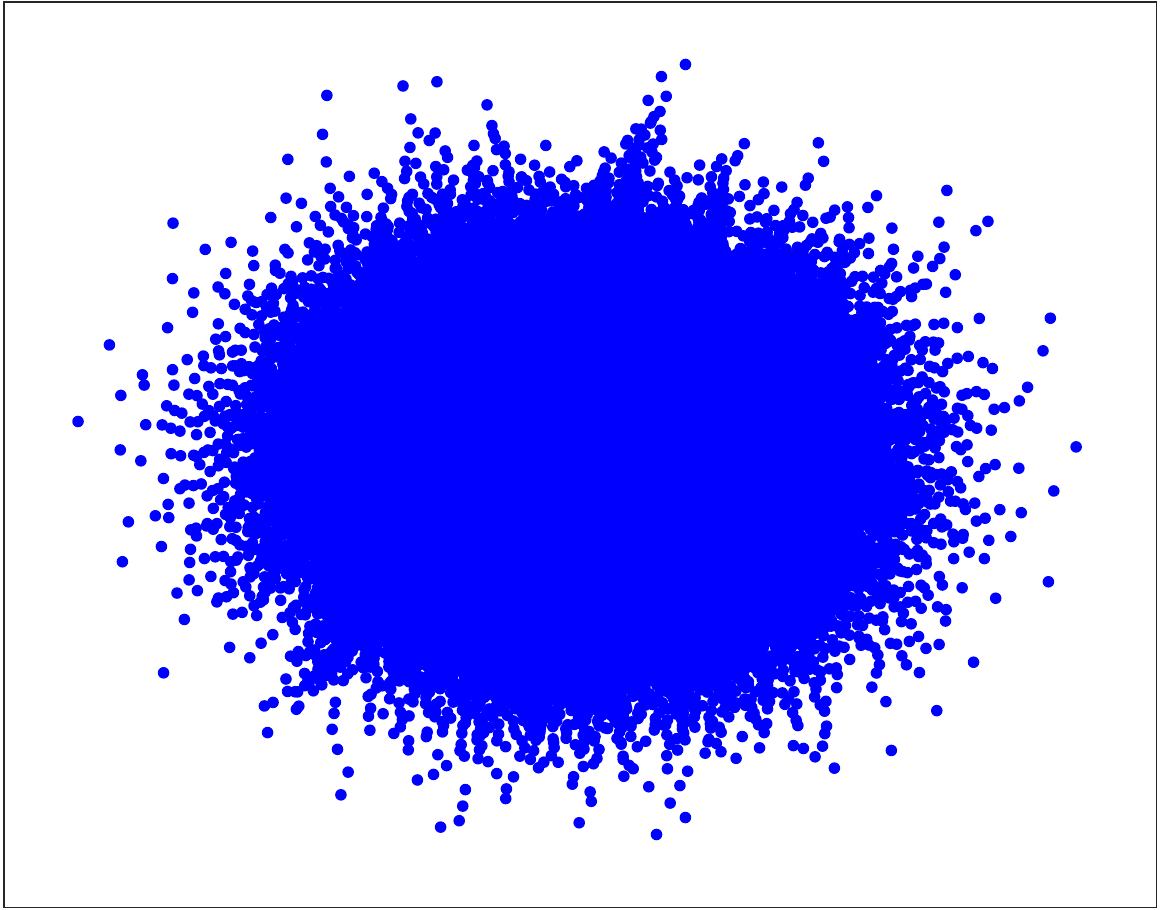}&
\includegraphics[trim=0.4cm 1.65cm 0.5cm 0.5cm, width=5cm, height= 4cm]{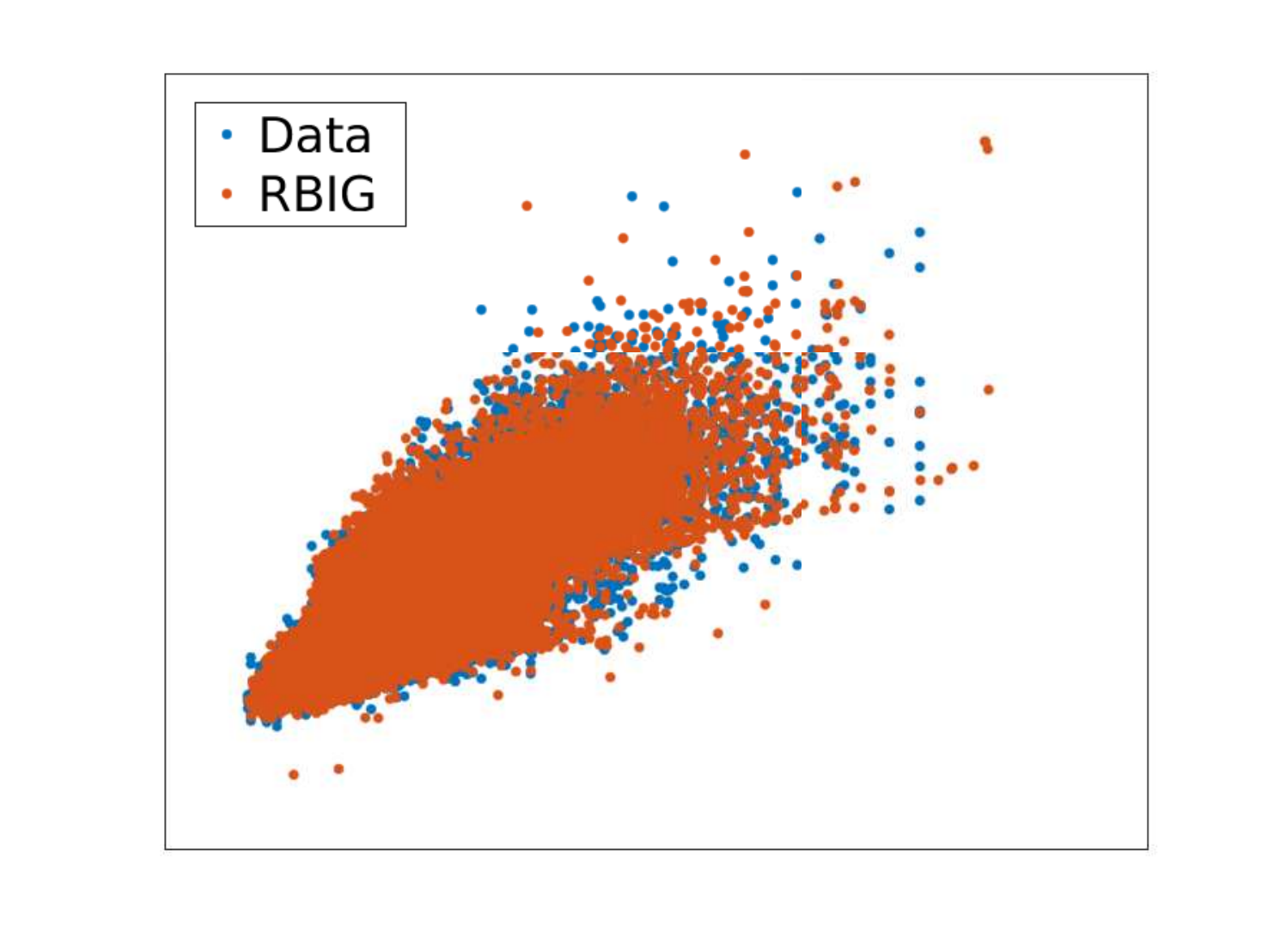}
\\

\end{tabular}
\end{center}
\caption{Illustration of synthesized data using the RBIG methodology in a real Sentinel-2 image. Top-left: RGB composite of the original image. Top-right: representation of the first two bands of the original image. Bottom-left: first two dimensions of the Gaussianized data. Bottom-right: first two bands of the original image data (blue) and randomly generated data inverted using the learned Gaussianization transformation (orange).}
\label{fig:toy1}
\end{figure}

An illustration of how RBIG can be adapted to describe the distribution of remote sensing data is shown in Fig.~\ref{fig:toy1}. In this example we take data from the Sentinel-2 image {\small \sf Australia} (see Table~\ref{table:database} for details), which has $d=12$ bands, and use RBIG to Gaussianize its pixel's distribution. We can see that the Gaussianized data follows a Gaussian distribution. Besides we apply the inverse of the learned Gaussianization transformation to randomly generated Gaussian points obtaining synthetic new data that follows a deemed similar distribution as the original one. This illustrates the invertibility property of RBIG, which allows us to estimate densities in the original domain and use the well-known relation between probability and anomaly to derive unsupervised density-based anomaly and change detectors. { Now we describe how 
Gaussianization can be applied to both anomaly detection (AD) and change detection problems (CD). The point-wise density estimate can be trivially linked to a degree of anomaly in a particular image; the higher the probability the lower the anomalousness. Likewise, the point-wise density estimate can be used to assess the probability of change between images; ones uses the first image to estimate the  background, and applies the transform to the second image.} 

\subsection{RBIG for Detection of Anomalies}

One of the most successful methods applied to the problem of anomaly detection is the Reed-Xiaoli (RX) method \cite{Reed90tassp}, a successful type of matched filter. The idea behind the RX method can be interpreted in probabilistic terms \cite{Padron_RX}; intuitively, a data point is more anomalous when it has less probability to appear:
\begin{equation}
A(\x) \propto \frac{1}{p_{X}(\x)}.
\label{eq:AD_prob}
\end{equation}
Actually, when the distribution is assumed to be Gaussian, $p_{X} \sim p_G$, this relation defines the RX method anomaly detector, i.e. $A(\x) \sim A_{\text{RX}}(\x)$. Actually $A_{\text{RX}}(\x)$ is equivalent to the Mahalanobis distance between the data point and the mean, i.e. $A_{\text{RX}}(\x) = (\x-\mu)^\top\boldsymbol{\Sigma}^{-1}(\x-\mu)$, where $p(\x)\sim{\mathcal N}({0},\boldsymbol{\Sigma})$. 

While RX has been widely used, it has the limitations inherent to the Gaussian distribution assumption. The use of kernel methods has been proposed to generalize the RX method to the nonlinear and non-Gaussian case \cite{Kwon_KRX,Padron_RX}. {Conceptually, the kernel RX (KRX) method can be understood as a way to propose a Gaussian distribution where the covariance is defined in a higher dimensional Hilbert feature space. Taking as a reference the RX method, this translates into replacing the covariance matrix with a kernel matrix that estimates the similarities between samples \cite{CampsValls09wiley,Rojo18dspkm}.} In practice, this implies that correlation is substituted by a non-linear (kernel) similarity measure. The anomaly detected using the KRX method can be formulated as:
\begin{equation}
A_{\text{KRX}}(\x) \propto \frac{1}{p_K(\x)},
\end{equation}
where $p_K(\x)$ is the distribution induced by using the kernel function instead of the covariance.
The KRX is an elegant extension of the RX, yet it has the problem of fitting kernel parameters and the high computational cost (as one has to invert a kernel matrix, which has cubic cost with the number of points $\ell$). Whereas some heuristics exist in the literature to fit the kernel parameters, in practice one only achieves the full potential of the KRX approach by fitting the parameters after cross-validation \cite{Padron_RX}. This requires having access to labeled data as anomalous versus non-anomalous classes, which is not a very realistic and not even practical setting. In this work, we approach the more useful and practical, yet more challenging, problem of unsupervised anomaly detection (i.e. no labeled data available), and therefore in our comparisons we will fit the kernel method parameter using the most successful (and sensible) heuristic to set the Gaussian kernel lengthscale $\sigma$ as the average of all distances among $\X$.

As an alternative to linear measures of anomalousness like in RX, or nonlinear yet implicit feature transformations with parameters to tune like in KRX, we here propose a more straightforward approach to estimate the probability density function with RBIG (sec. \ref{sec:rbig}). {Here we stick to the idea of assuming a Gaussian distribution in a transform domain. In particular, RBIG forces the true distribution in the transform domain to be a multivariate Gaussian. However, note that the distribution in the original domain is estimated not only assuming Gaussian distribution in the transform domain but also including the determinant of the Jacobian of the transformation (see eq. \ref{eq:change_variables}). This gives us a non-parametric, parameter-free and efficient estimation of the data distribution in the original domain. RBIG is an unsupervised method by construction so it does not require labeled data, and scales linearly with the data.} By using RBIG to compute $p_X$, we obtain the method proposed in this work:
\begin{equation}
A_{\text{RBIG}}(\x) \propto \frac{1}{p_{\text{RBIG}}(\x)}.
\end{equation}

An important aspect to take into account is the intrinsic characteristics of the data used to estimate the density, which has implications in the quality of the estimation. {When the distribution contains even a moderate number of anomalies, an accurate density estimate will cast anomalies as regular points, i.e. non-anomalous.}
This vastly depends on the flexibility of the class of models used. 
When the model is rigid like in the RX case, this is not a problem since it cannot be adapted to the anomalies. 
For the KRX one can control this effect by tuning the kernel lengthscale and the regularization term, but as explained before requires labeled data. 
This is an important aspect to take into account mostly in the anomaly detection scenario, where all data (included the anomalous samples) are used to estimate the density. 
Therefore we propose to use an hybrid model that combines the (too rigid) RX model with the (too flexible) RBIG model. {First, we apply RX to discard data more likely to be anomalous. The remaining points are then used to apply RBIG. This tries to avoid using anomalous data in RBIG, which after all is intended to learn the background or pervasive data distribution.} The number of data points selected as non-anomalous in the first step will define the trade-off between flexibility and rigidity.

\subsection{RBIG for change detection}

{Remote sensing change detection algorithms have mainly relied on thresholding or classifying the difference (or ratio) between a pair of previously co-registered images \cite{Liu2019}. Multitemporal registration and radiometric and atmospheric corrections are thus very important aspects that strongly impact the detection performance. A minimum error in image co-registration for example leads to false detections. In this paper we illustrate and motivate the use of RBIG following an alternative probabilistic approach. Basically, the idea is to characterize statistically the original image probability, and attribute the change to pixels in the second image with low probability. When applied pixel-wise (as we do here) this has the drawback of not detecting some changes that are not statistically noticeable. For instance, in two images with the same car in different locations of a road, the method would not detect the car as a change.  Such unfortunate cases could be however addressed by extending the idea easily by working in a sliding window instead of a pixel-wise approach. The statistical approach has the advantage of reducing the importance of perfect co-registration since the method operates in the geometric space defined by the image, not in the spatial domain explicitly. Thus, small errors of co-registration would not penalize the performance of the proposed approach.}

The idea to exploit RBIG for change detection is using data coming from the first image $\X_1$ only to estimate the probability model and then evaluating the probability (or change score, $C$) for each point in the second image $\X_2$, as follows:
\begin{equation}
C(\x_2) \propto \frac{1}{p_{X_1}(\x_2)}.
\label{eq:CD}
\end{equation}
As for the anomaly detection case, we can use different models to estimate $p_{X_1}$. The most widely used is the Gaussian model. As in the previous section, when assuming a Gaussian distribution for the input data, the RX method can be used here too, i.e. $C_{\text{RX}}(\y)$. 

Likewise, kernel methods have been proposed to alleviate the strict assumption of Gaussian distribution \cite{Padron_RX}
. While different configurations were proposed in order to take into account only the anomalous changes, here we use the configuration designed for change detection. Following the idea in equation \eqref{eq:CD}, the data of the first image ($\X_1$) is used to estimate the kernel and then the method is evaluated in the second image:
\begin{equation}
C_{\text{KRX}}(\x_2) \propto \dfrac{1}{p_{\text{K}(X_1)}(\x_2)}.
\label{eq:CD_krx}
\end{equation}
Equivalently, we can use RBIG to estimate the probability of the first image and evaluate the probability in the second one:
\begin{equation}
C_{\text{RBIG}}(\x_2) \propto \dfrac{1}{p_{\text{RBIG}(X_1)}(\x_2)}.
\label{eq:CD_rbig}
\end{equation}
It is important to note that, in this case, the data used to estimate the probability density does not contain anomalies (changes in this setting) so the hybrid model is not needed here.

\section{Experimental Results}\label{sec:experiments}

This section analyzes the performance of the proposed RBIG method for anomaly and change detection. In order to assess the robustness we performed tests in both simulated and real scenes of varying dimensionality and sample size. 
We evaluate the detection power of the methods quantitatively through the Receiver Operating Characteristic (ROC) and Precision-Recall (PR) curves, along with the Area Under the Curve (AUC) scores. Besides, we provide examples of detection maps of each method to evaluate their quality by visual inspection.

We have performed three experiments. The first experiment is designed to illustrate the effect of the evaluated in an anomaly detection (AD) toy example. The second experiment deals with AD problem in different real scenarios: detection of air planes, latent fires, vehicles, and urbanization (roofs). The third experiment is related to evaluate the methods in change detection (CD) problems involving floods, fires and droughts. Table \ref{table:database} summarizes the different data sets used in the experiments. 

{We added two standard methods in the experimental results for comparison: (1) a kernel-based method known as the support vector domain descriptor (SVDD) \cite{Tax99}, where one seeks to embrace all data points into a hypersphere in Hilbert space, and (2) the classical kernel density estimation (KDE), where a Gaussian kernel is used to define the distribution. All methods were used in the same way to the anomaly detection (Eq. \eqref{eq:AD_prob}) and the change detection problems (Eq.~\eqref{eq:CD}).
In order to ease the reproducibility, we provide MATLAB code implementations of the all methods, as well as database with the labeled images used in the second and third experiments in \cite{RBIG4AD_web}.}

\subsection{Experiment 1: Simulated Anomalies}

\begin{figure}[h]
\begin{center}
\begin{tabular}{cc}
\subfloat[RX]{\includegraphics[width=4cm]{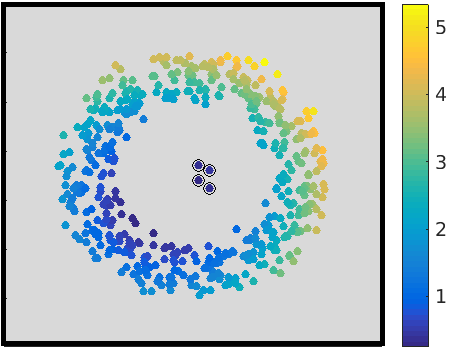}}&
\subfloat[KRX]{\includegraphics[width=4.3cm]{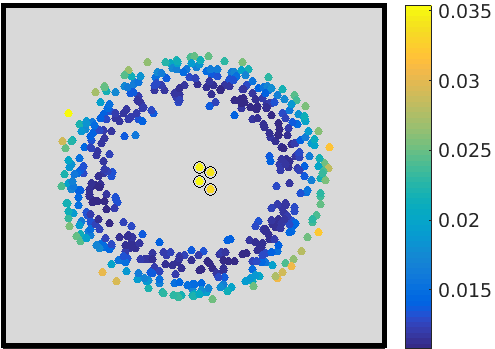}}\\
\subfloat[RBIG]{\includegraphics[width=4cm]{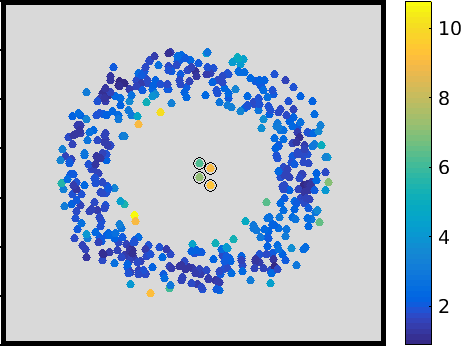}}&
\subfloat[HYBRID]{\includegraphics[width=4cm]{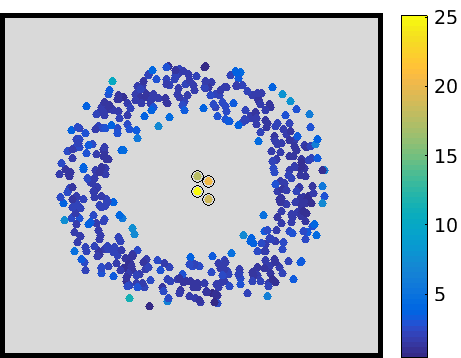}}
\end{tabular}
\end{center}
\caption{Synthetic experiment to illustrate the methods  performance when detecting anomalies. The color bar shows the intensity in terms of anomaly score from dark blue (less) to yellow (more). The image (a) correspond to RX detector, image (b) is the kernel version of RX, (c) represent the RBIG method and (d) showcase the hybrid model. }
\label{fig:toyRBIG}
\end{figure}

The aim of this experiment is to illustrate the behavior of the proposed methods in challenging distributions exhibiting highly nonlinear feature relations. We designed a two-dimensional dataset where the non-anomalous data is in a circumference and the anomalous data in the middle. 
Figure~\ref{fig:toyRBIG} shows the performance of the different methods. The RX method assumption does not hold (the data is clearly non-Gaussian), hence it shows poor performance. The performance of KRX is better than RX but some false detections emerge in the outer circle, mainly related to the difficulty to select a reasonable kernel parameter. 
The direct application of RBIG easily identifies the anomalous points since they are far from the more dense (most probable) region. The proposed hybrid model further refines the detection since the density is estimated from pervasive data yielded by RX only. 

\subsection{Experiment 2: Anomaly Detection in Real Scenarios}

We performed tests in four real examples. 
Table~\ref{table:database} summarizes relevant attributes of the datasets such as sensors, spatial and spectral resolution.

\subsubsection{Data collection}
\begin{figure*}[t!]
\setlength{\tabcolsep}{2pt}
\begin{center}
  \begin{tabular}{cccccc}
       
       \includegraphics[width = 3cm ,height = 3cm]{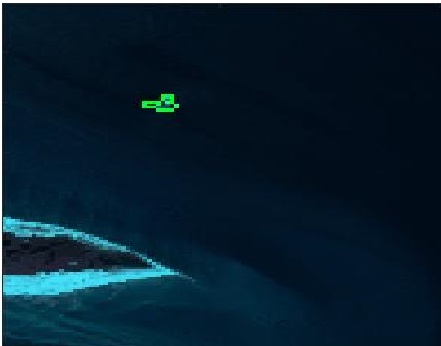}&
       
       \includegraphics[width = 3cm ,height = 3cm]{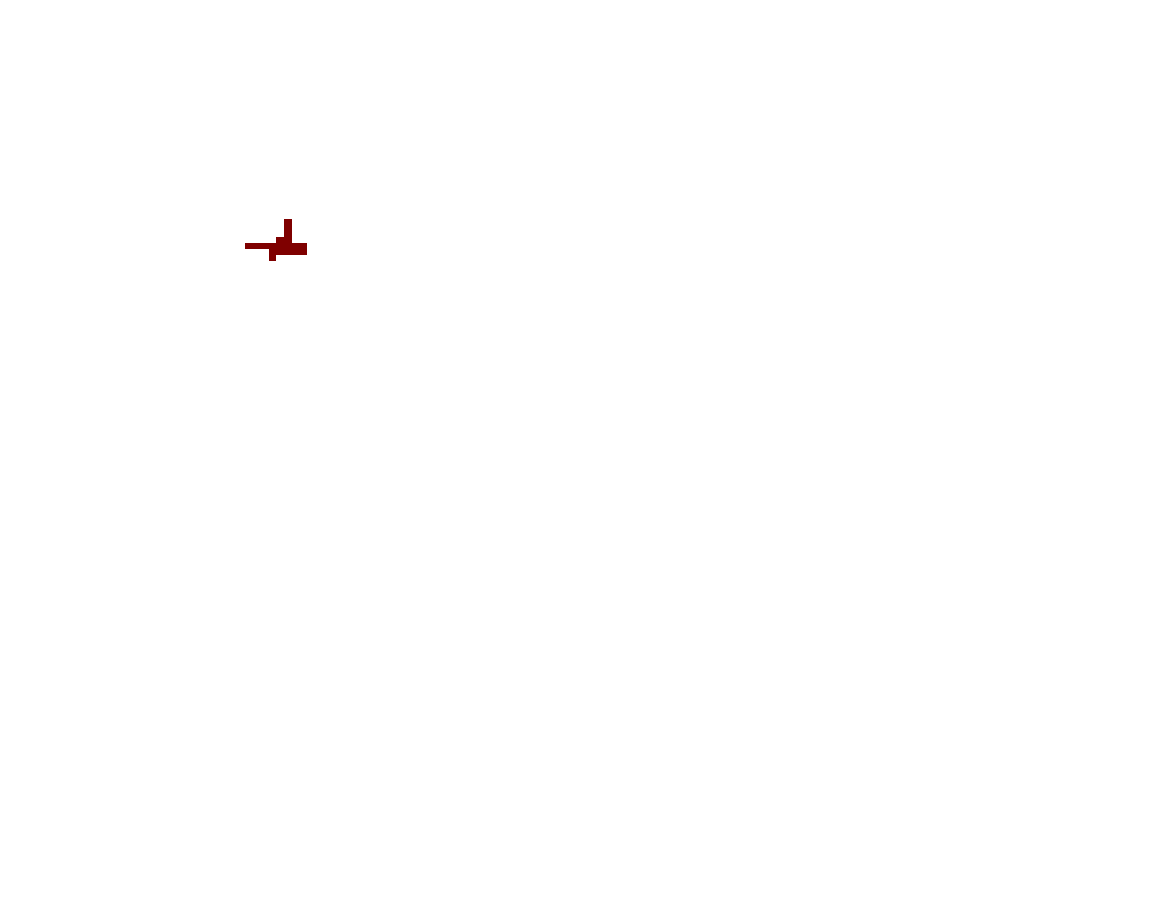}&
       
       \includegraphics[width = 3cm ,height = 3cm]{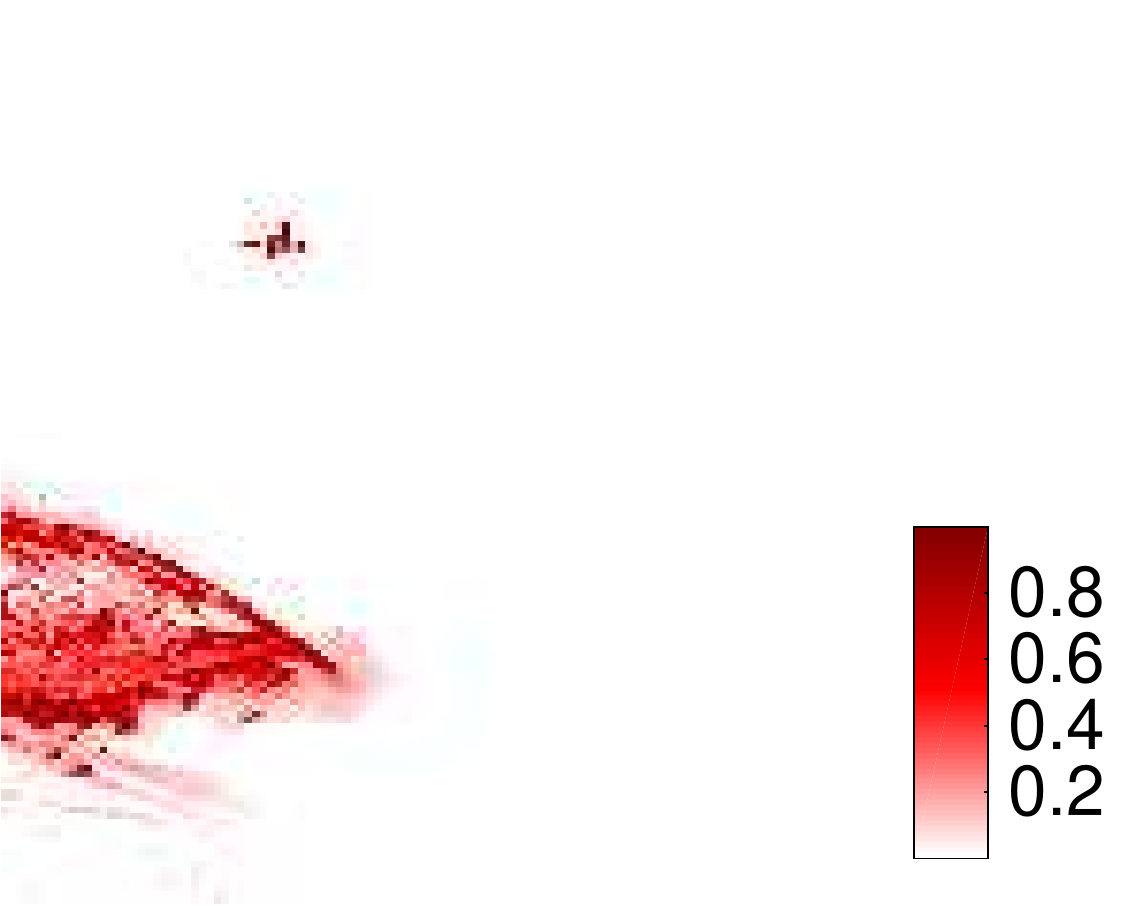}&
       \includegraphics[width = 3cm ,height = 3cm]{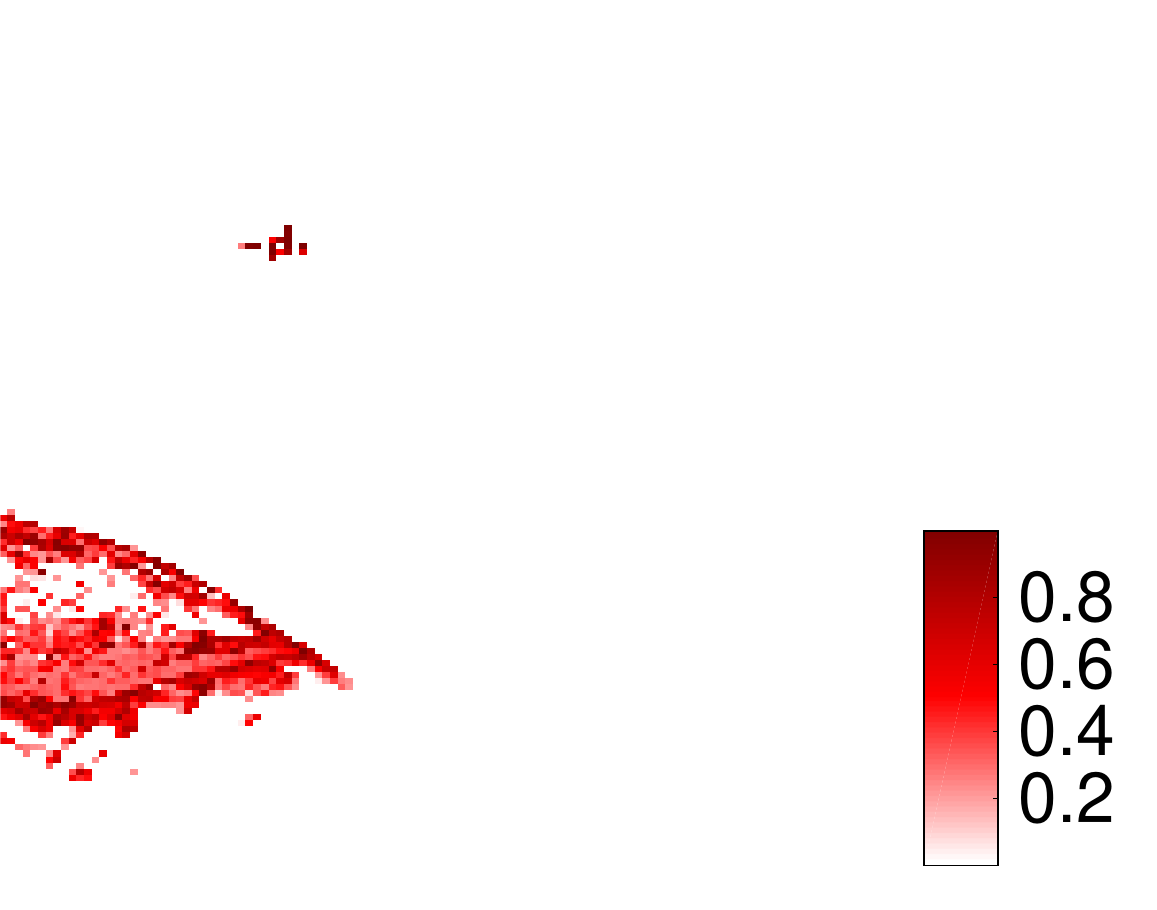}&
       \includegraphics[width = 3cm ,height = 3cm]{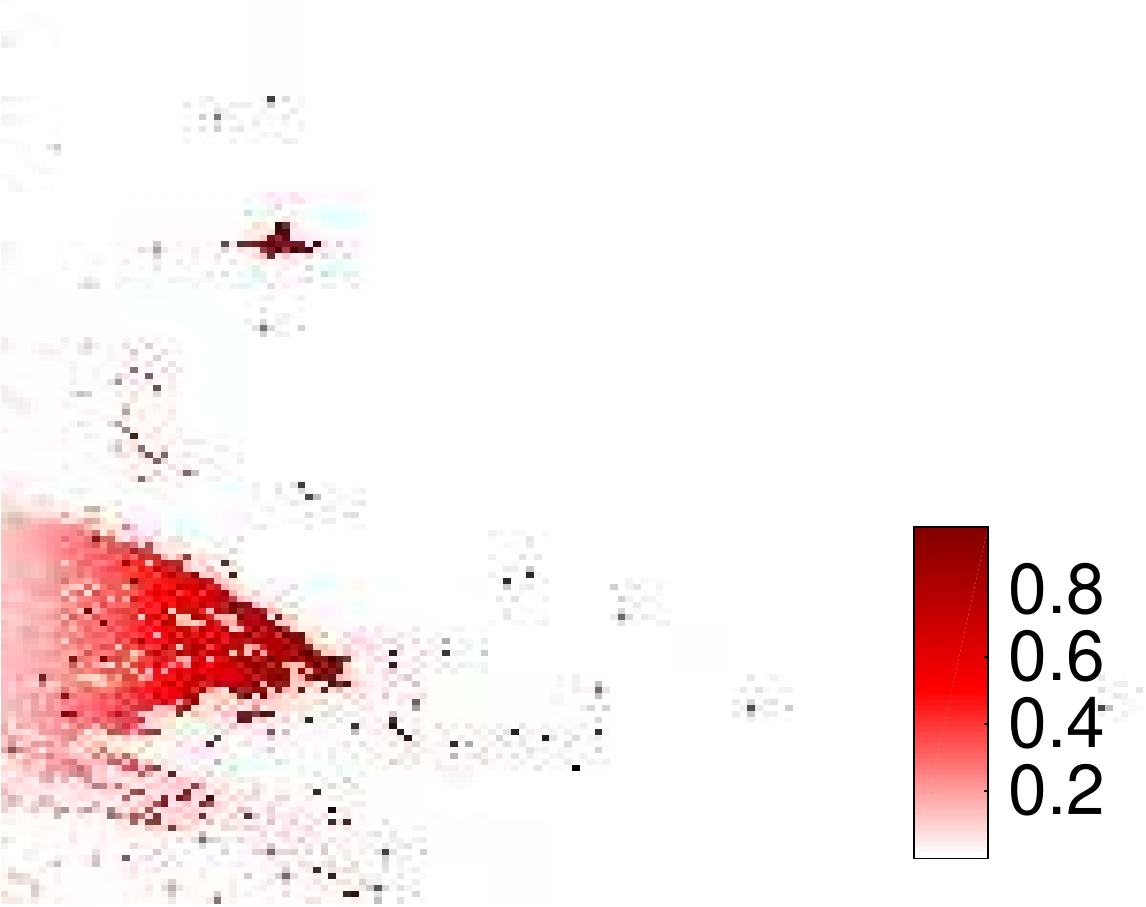}&
       \includegraphics[width = 3cm ,height = 3cm]{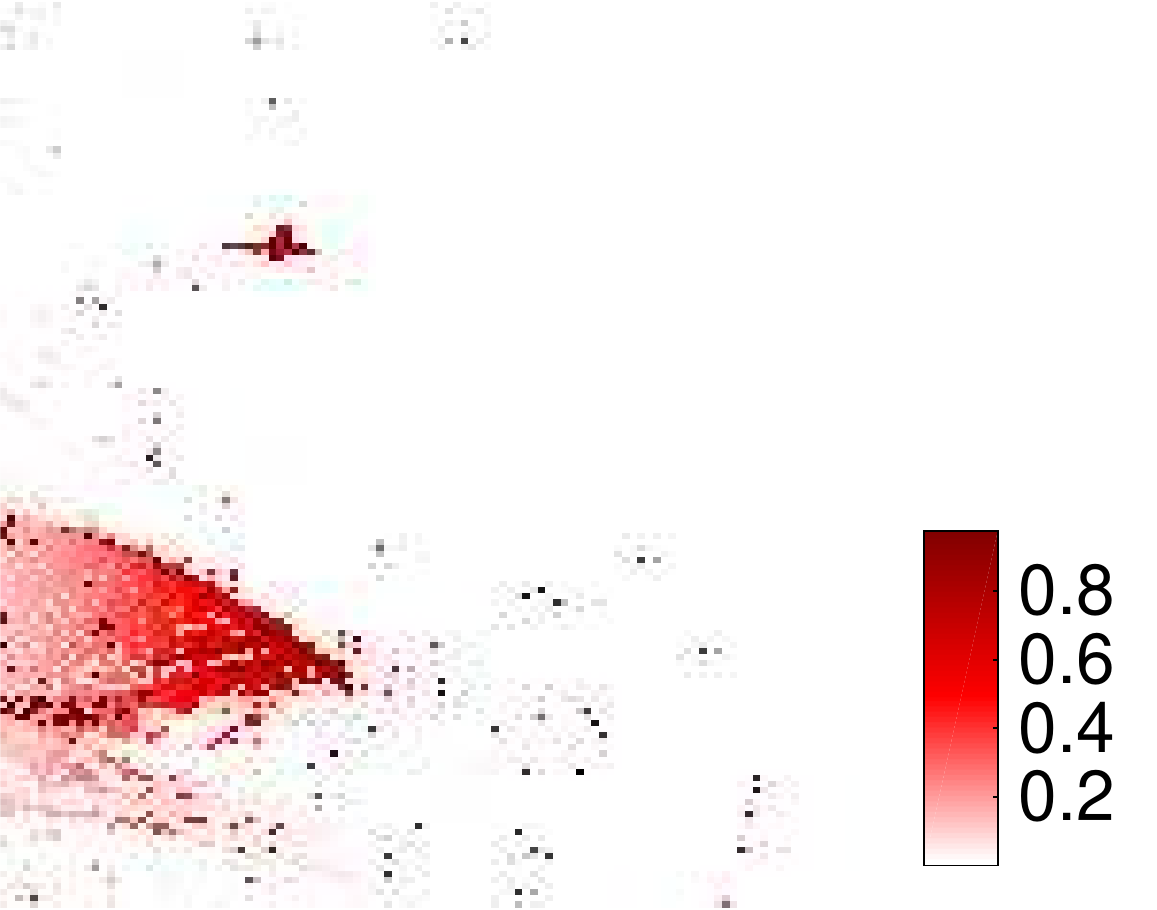}\\
       Cat-Island & GT  & RX (0.96) &  KDE (0.97) &  RBIG (0.99)& HYBRID (0.99) \\

       \includegraphics[width = 3cm ,height = 3cm]{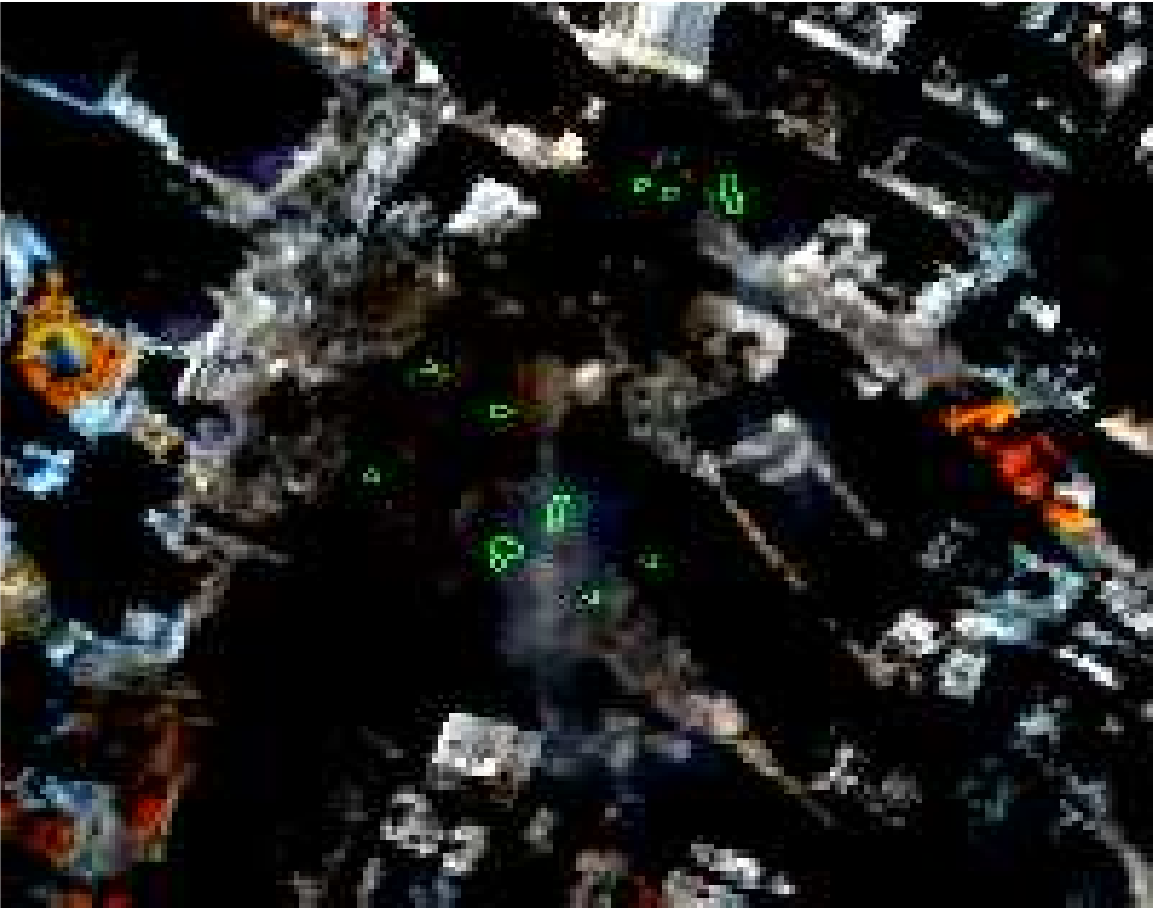}&
       \includegraphics[width = 3cm ,height = 3cm]{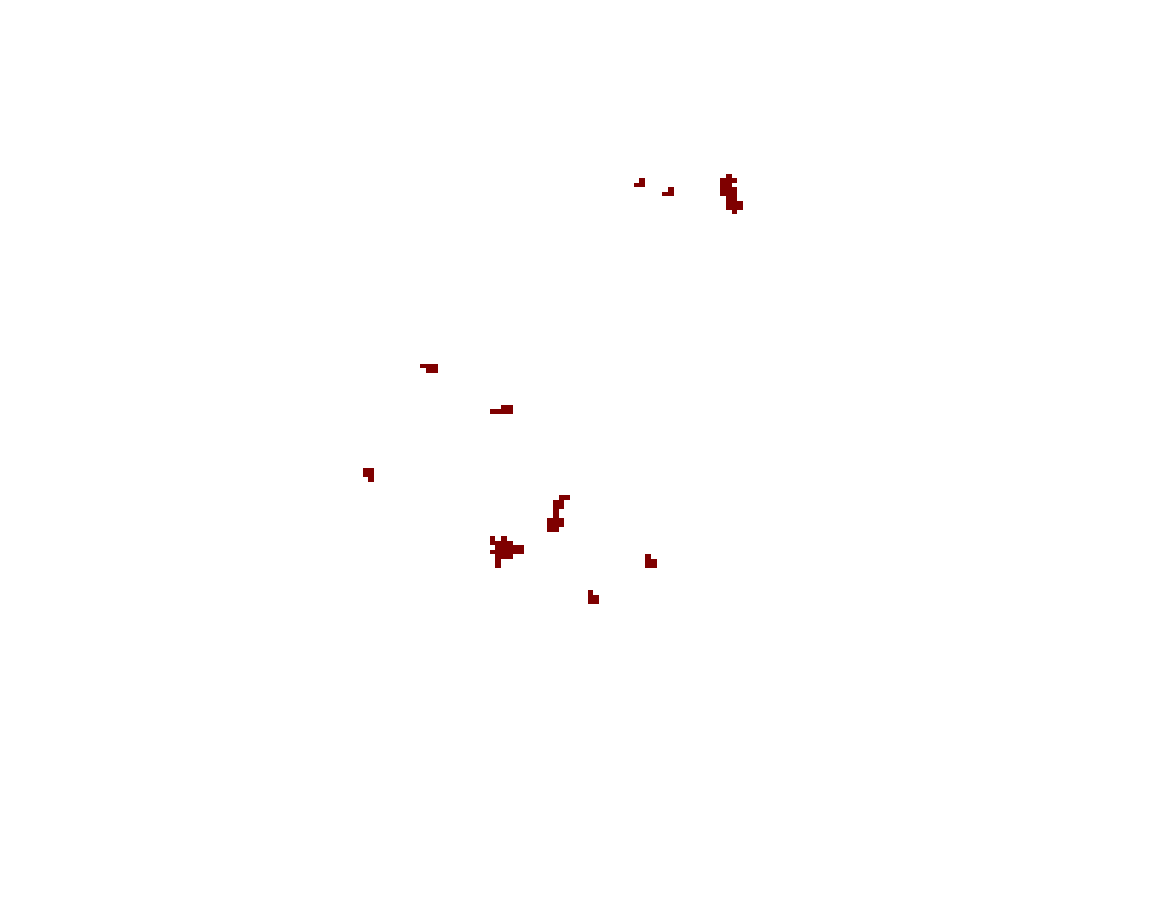}&
       \includegraphics[width = 3cm ,height = 3cm]{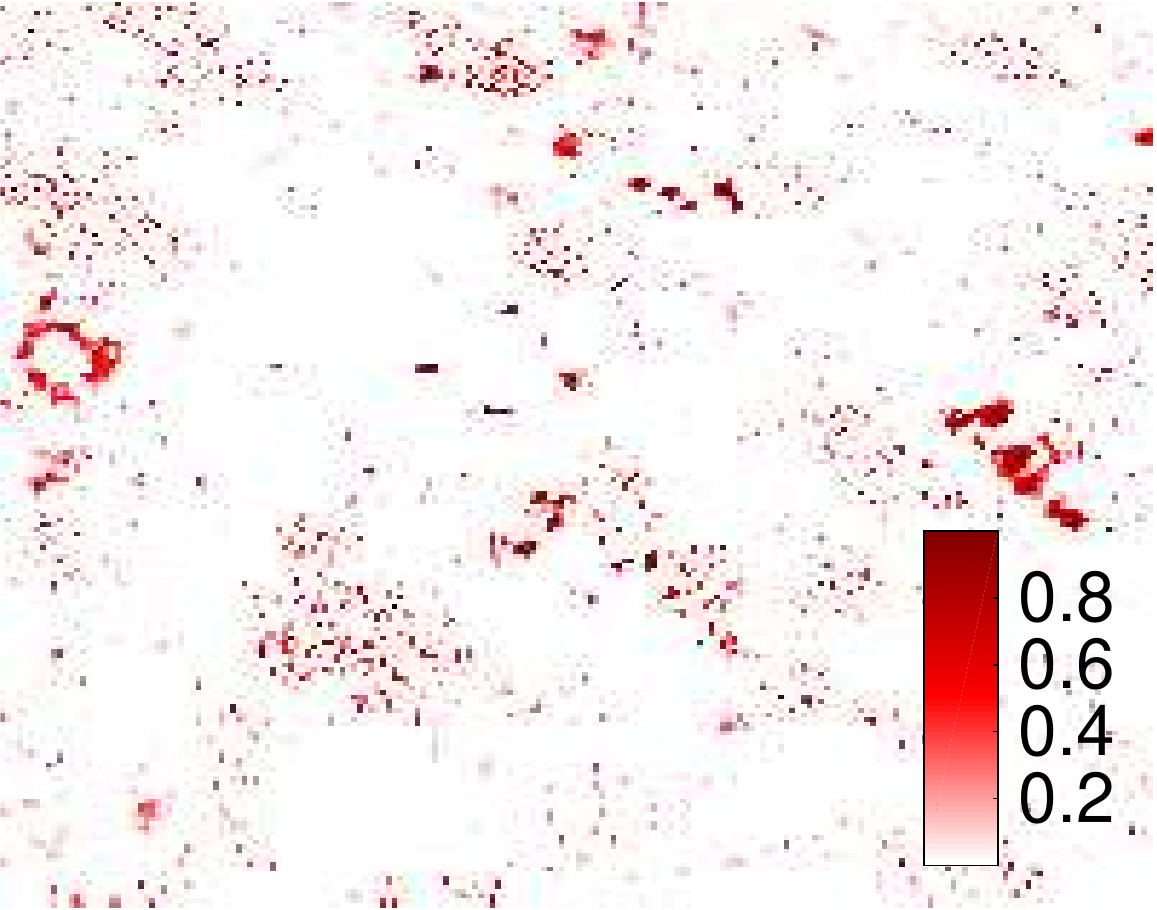}&
       \includegraphics[width = 3cm ,height = 3cm]{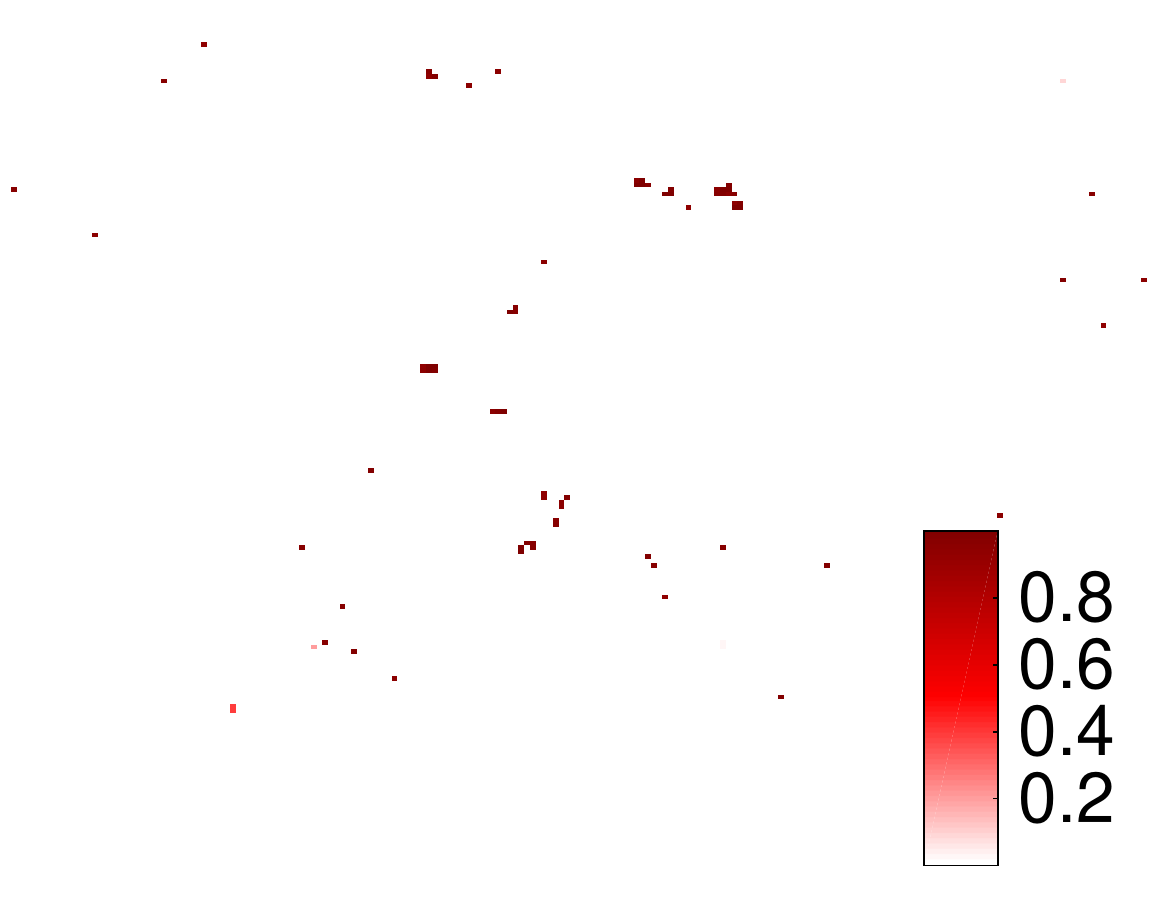}&
       \includegraphics[width = 3cm ,height = 3cm]{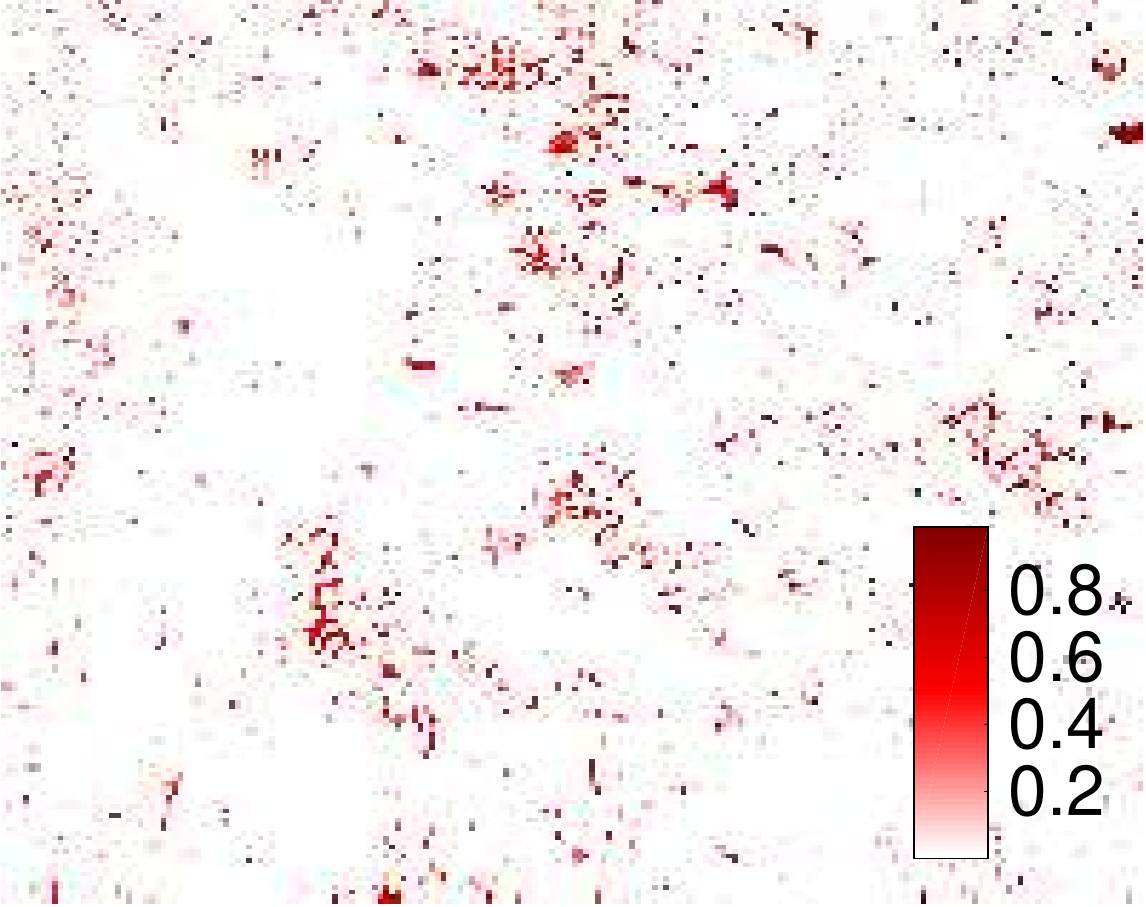}&
       \includegraphics[width = 3cm ,height = 3cm]{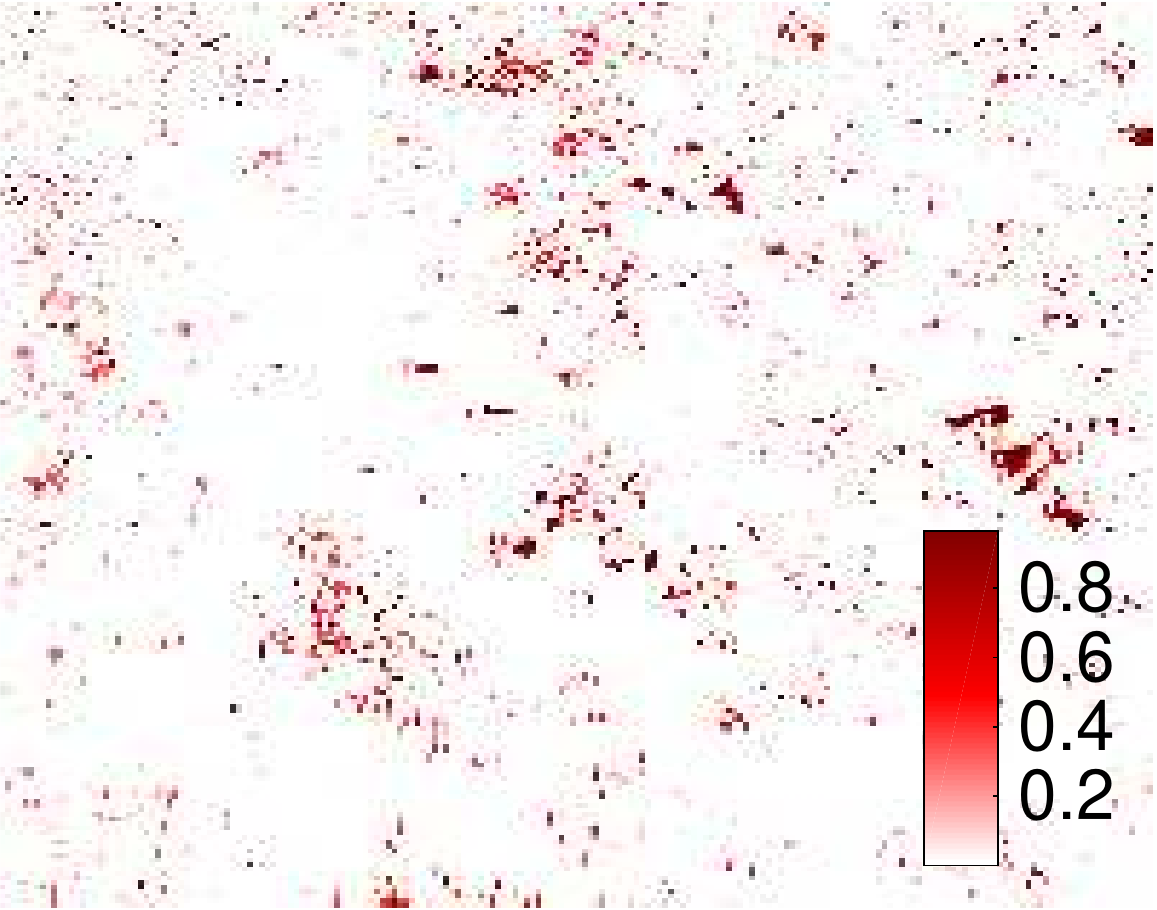}\\
        WTC & GT &  RX (0.95) &  KDE (0.95) &  RBIG (0.95) &  HYBRID (0.95)\\

\includegraphics[width = 3cm , height = 3cm]{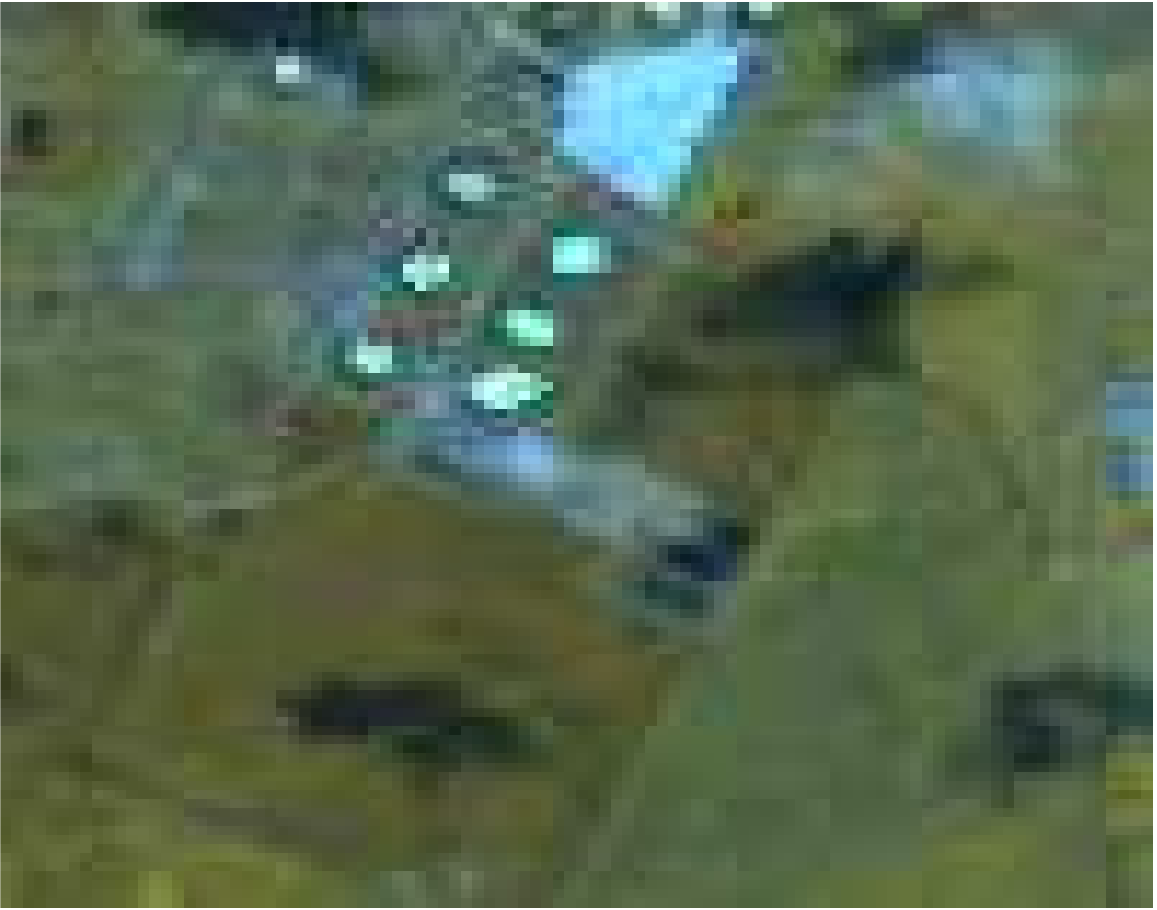}
&
\includegraphics[width = 3cm, height = 3cm]{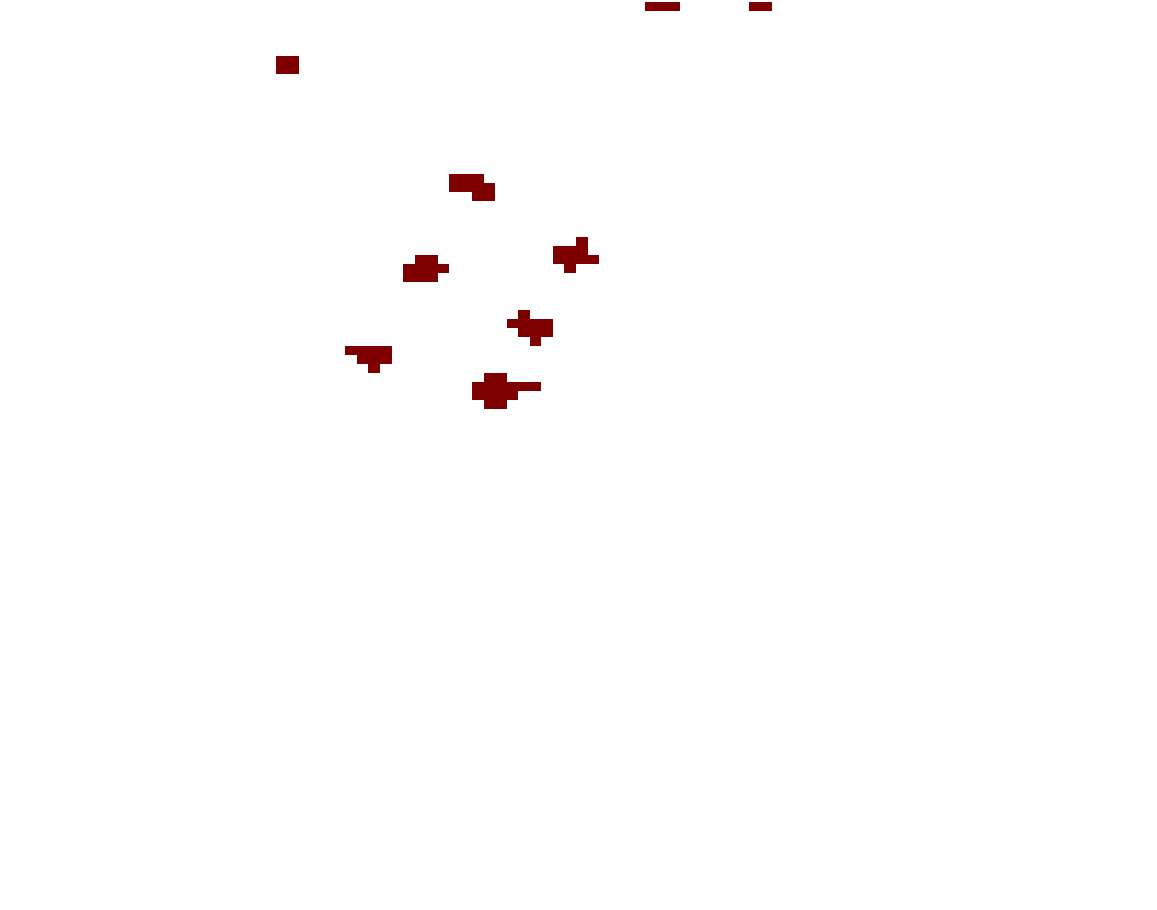}& 
\includegraphics[width = 3cm, height = 3cm]{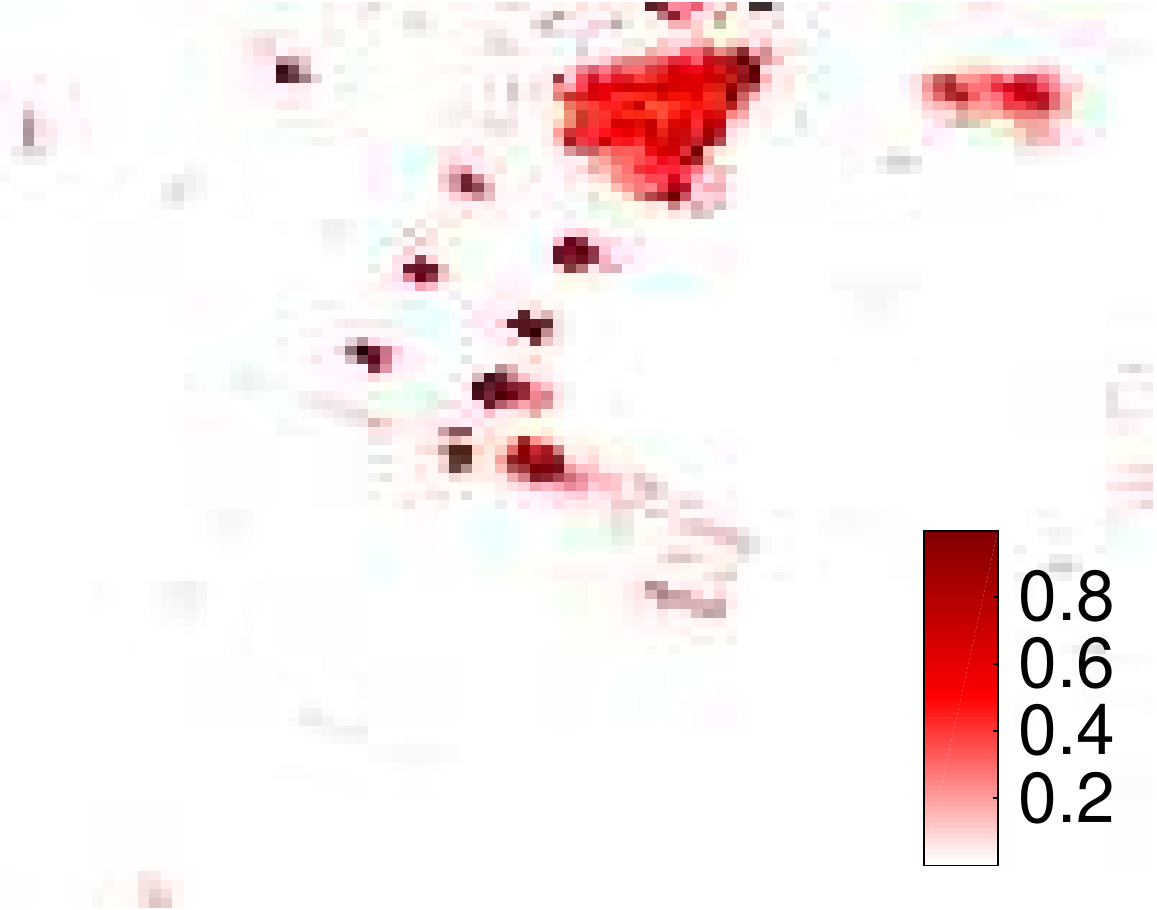}&
\includegraphics[width = 3cm, height = 3cm]{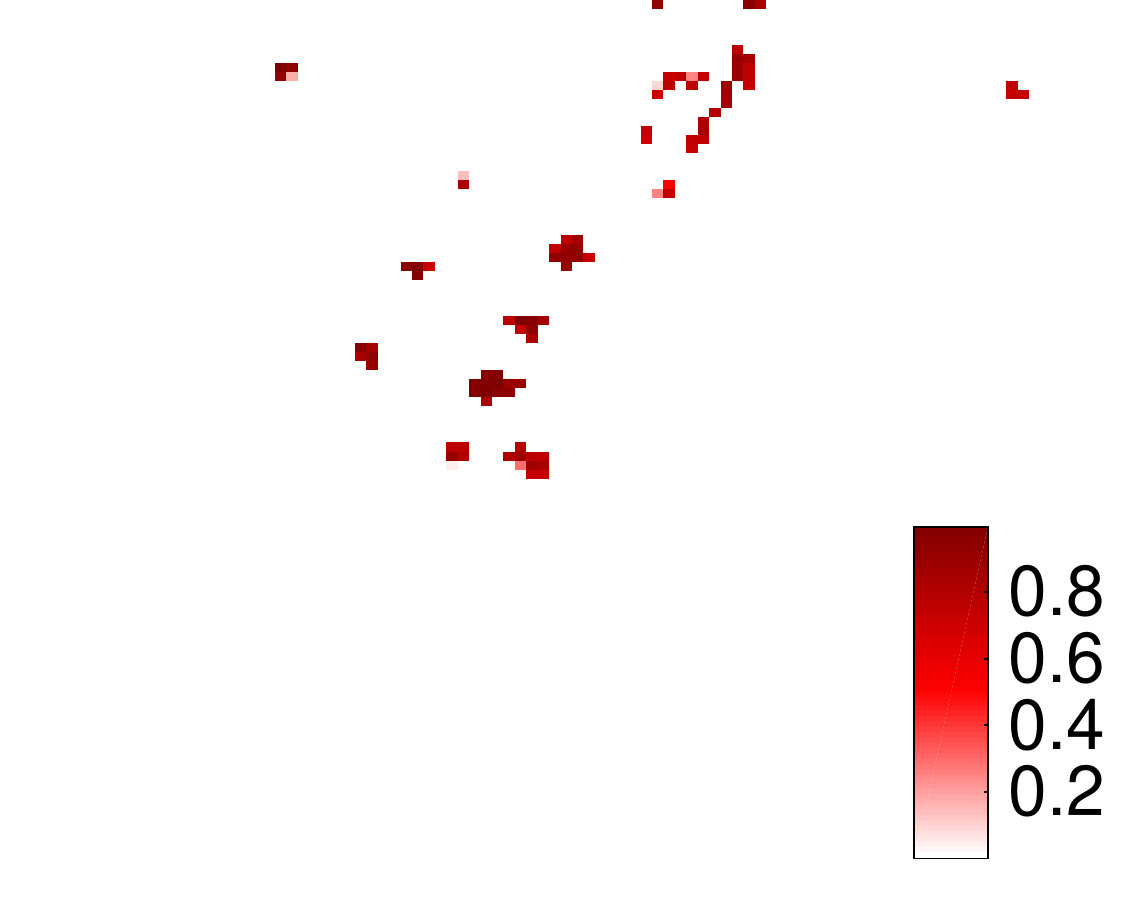}&
\includegraphics[width = 3cm , height = 3cm]{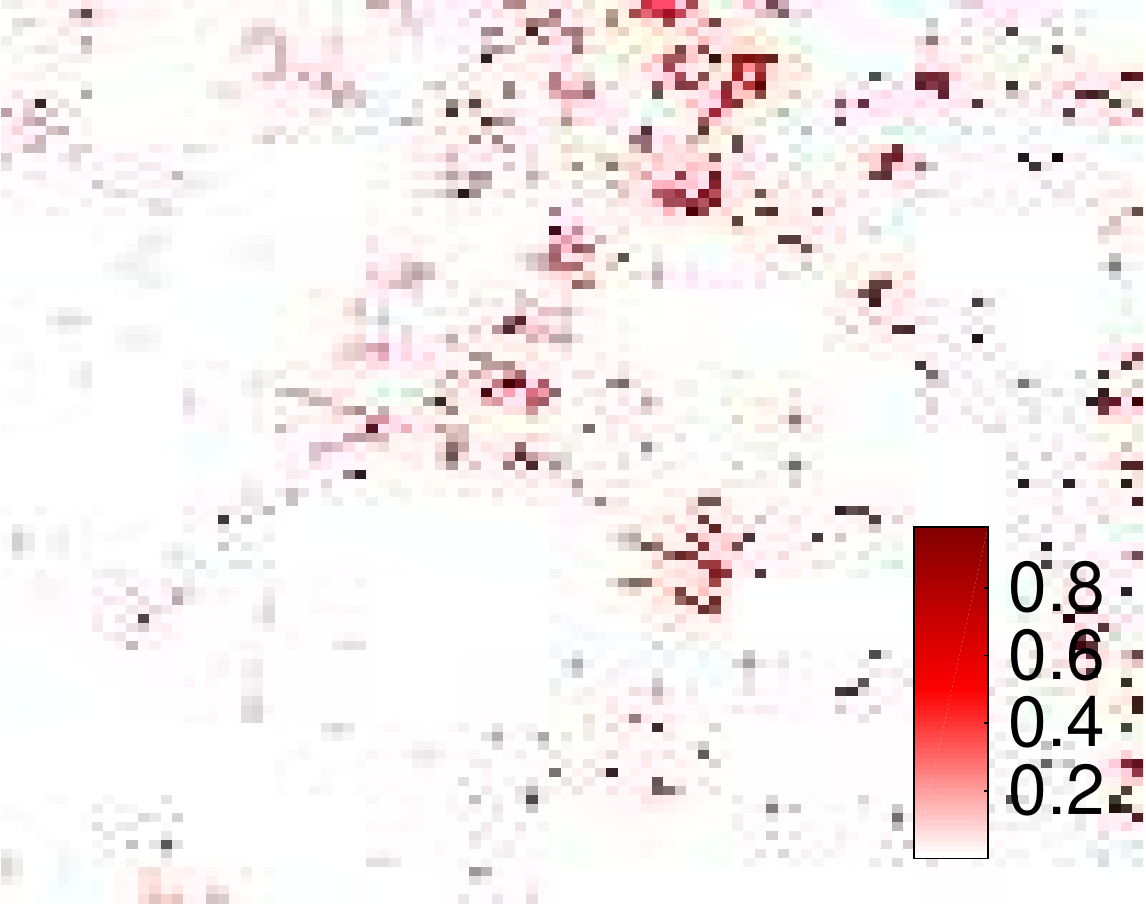}&
\includegraphics[width = 3cm , height = 3cm]{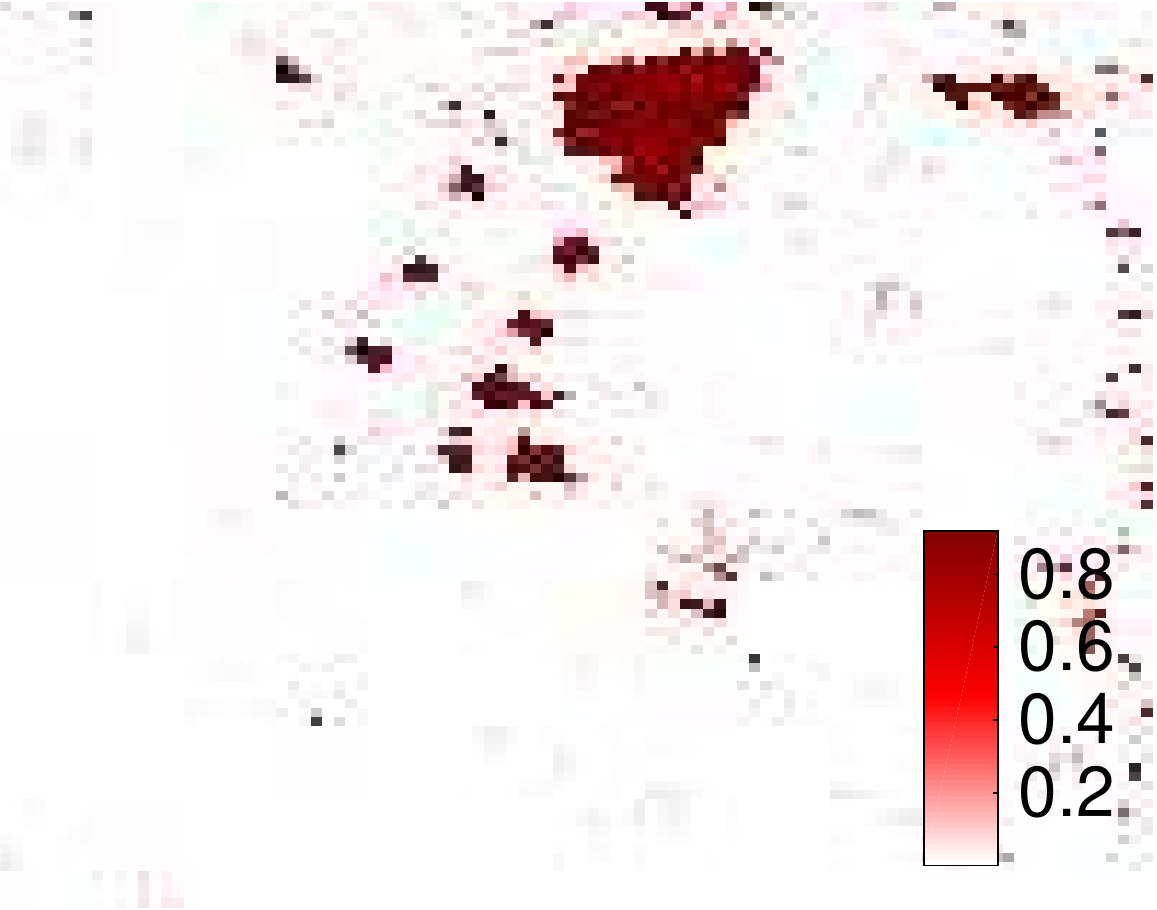}\\
  Texas-Coast & GT &  RX (0.99) &  KDE (0.99) &  RBIG (0.94) &  HYBRID (0.99)\\

\includegraphics[width = 3cm , height = 3cm]{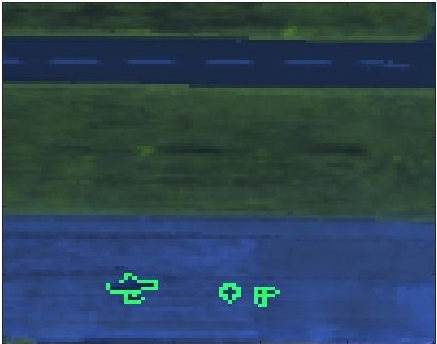}&
\includegraphics[width = 3cm , height = 3cm]{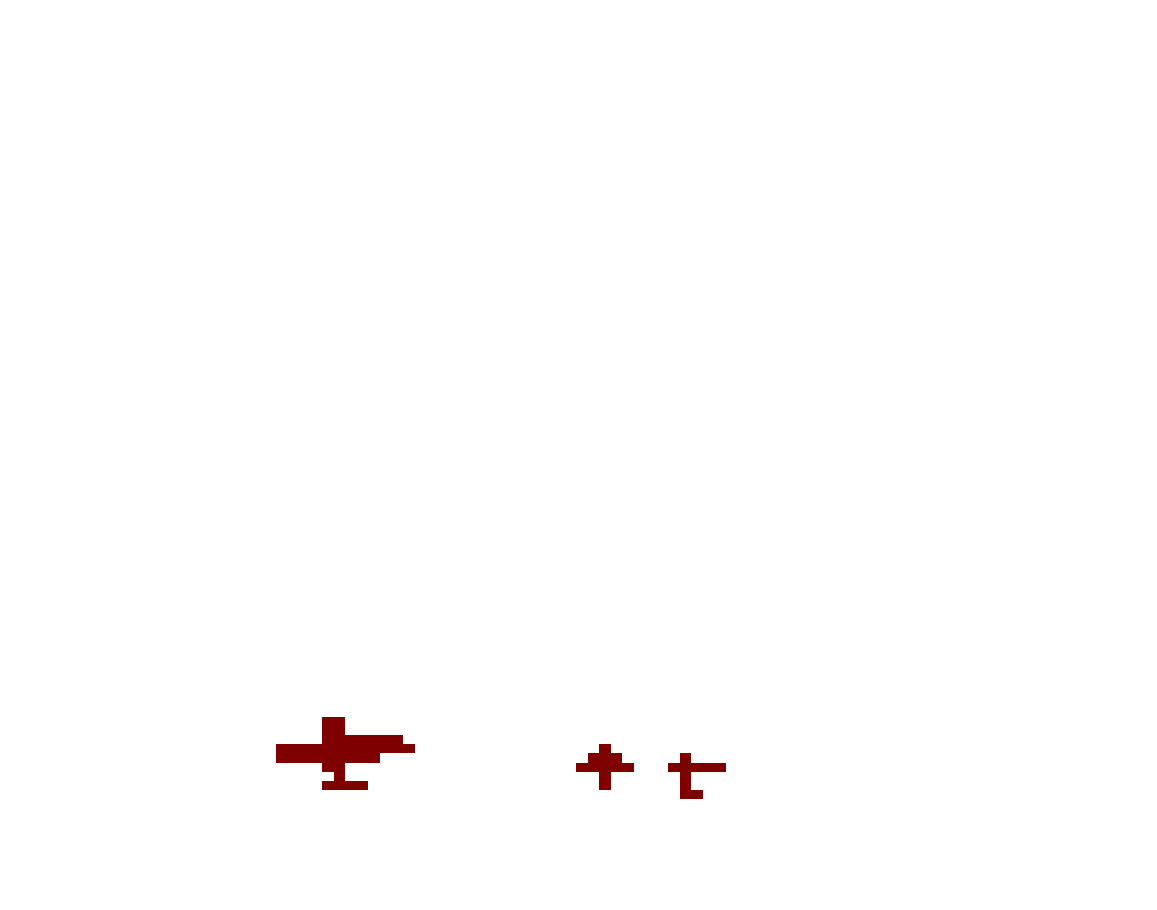}
&

\includegraphics[width = 3cm , height = 3cm]{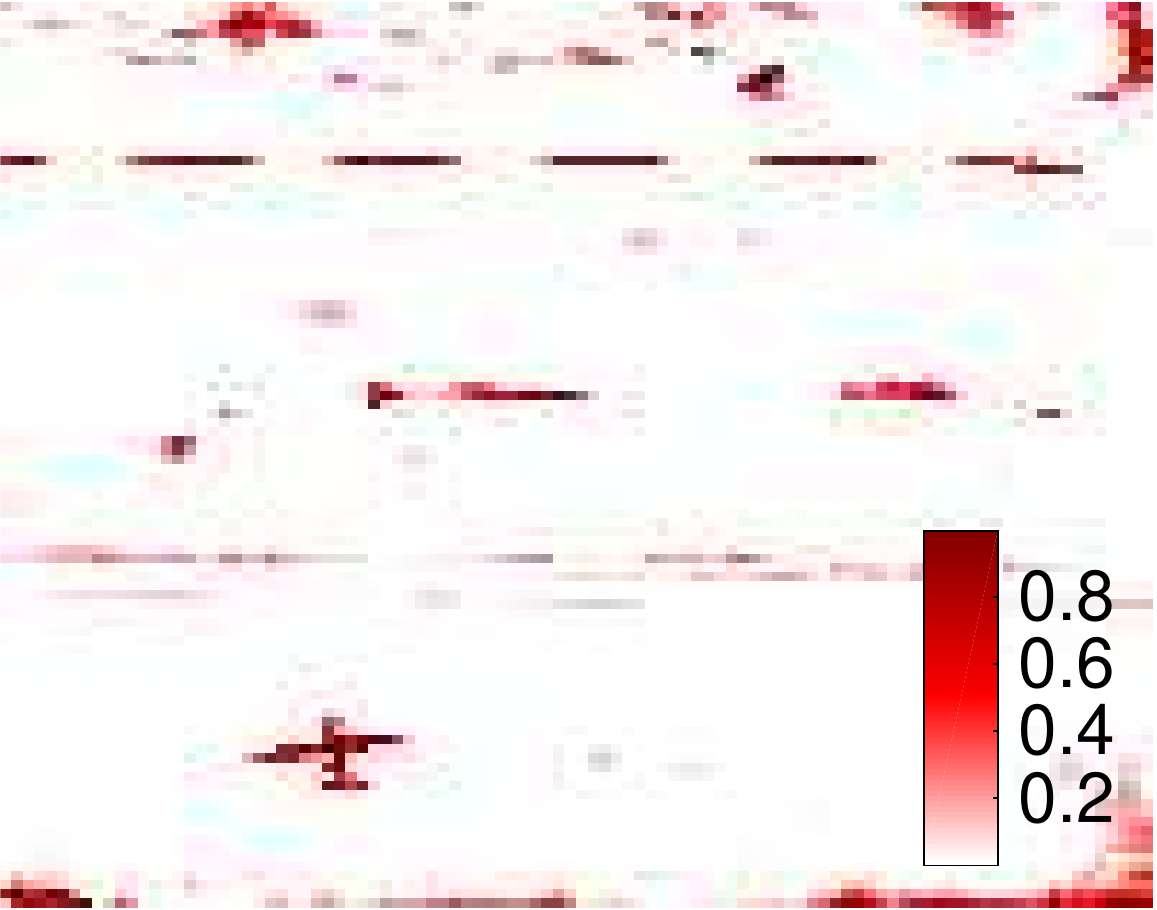}
&
\includegraphics[width = 3cm , height = 3cm]{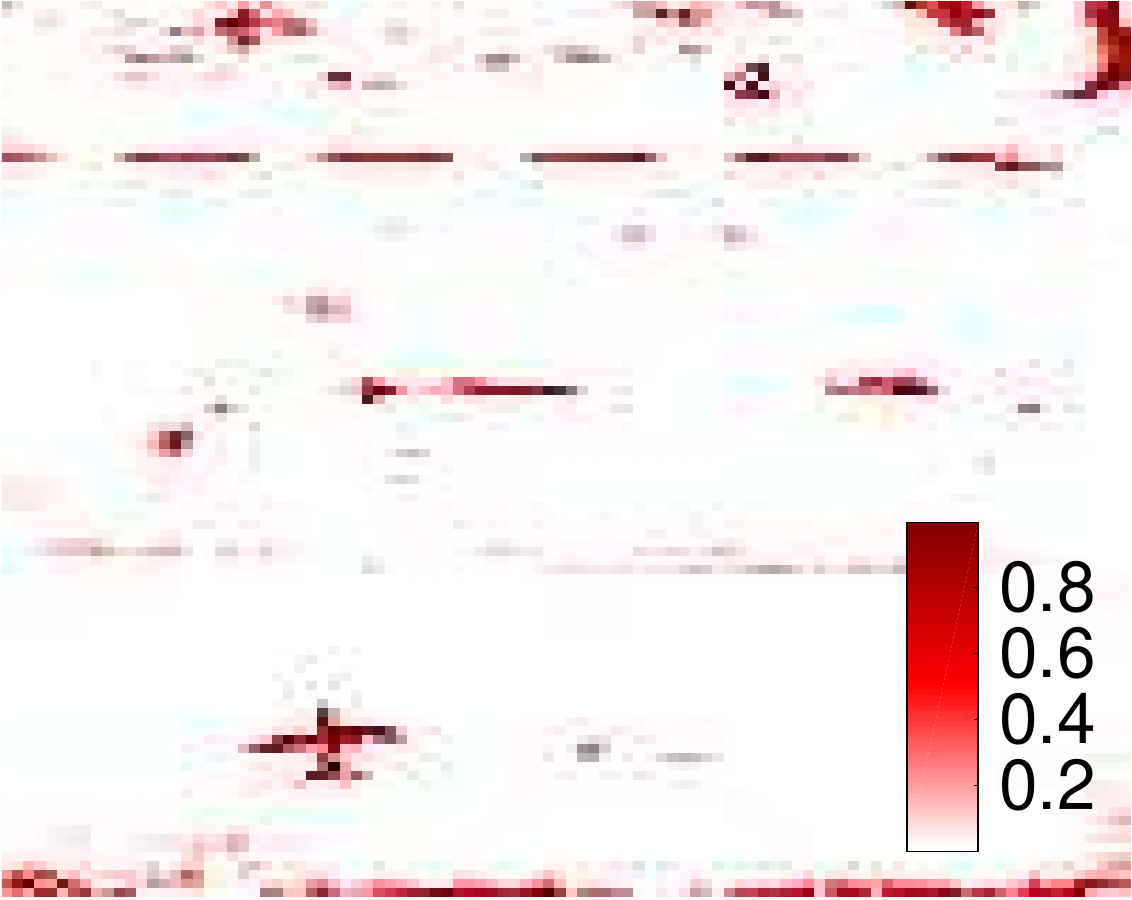}
& \includegraphics[width = 3cm , height = 3cm]{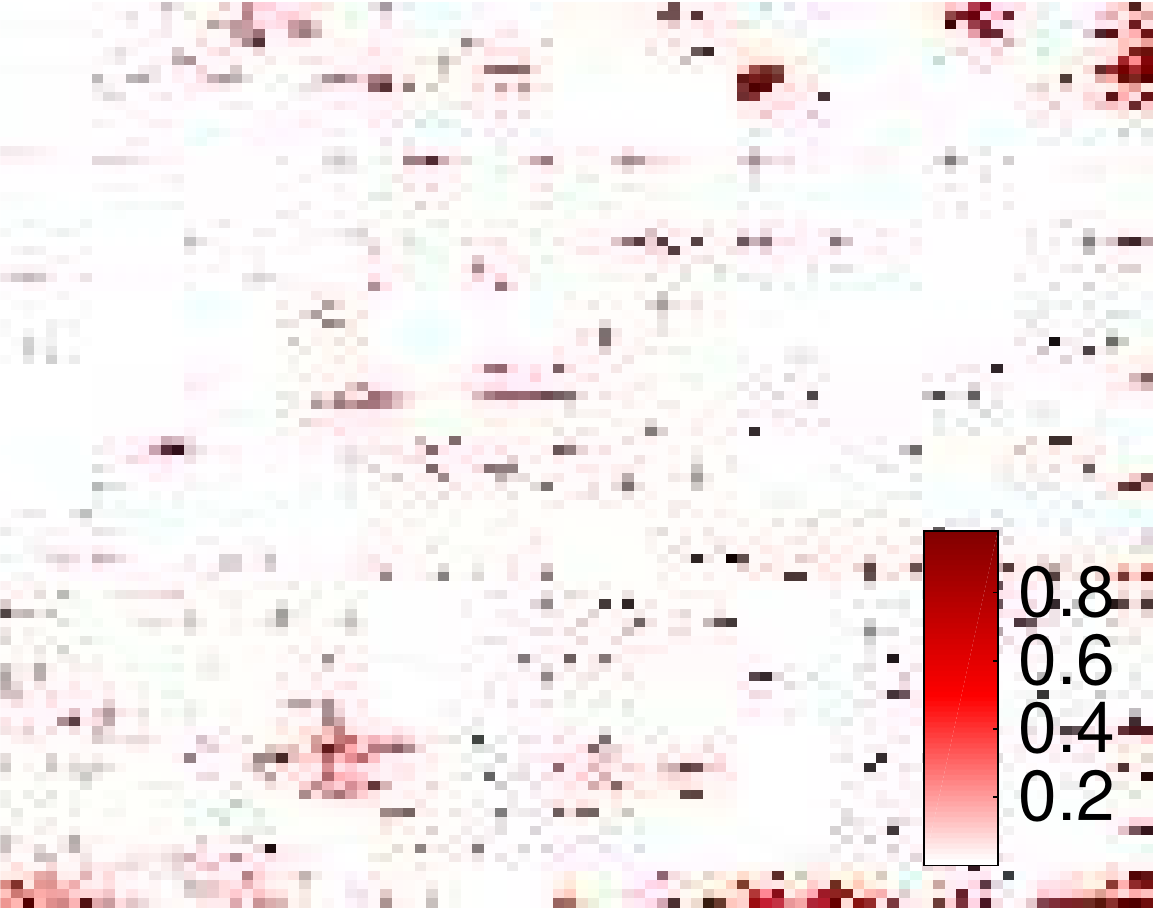}

& \includegraphics[width = 3cm , height = 3cm]{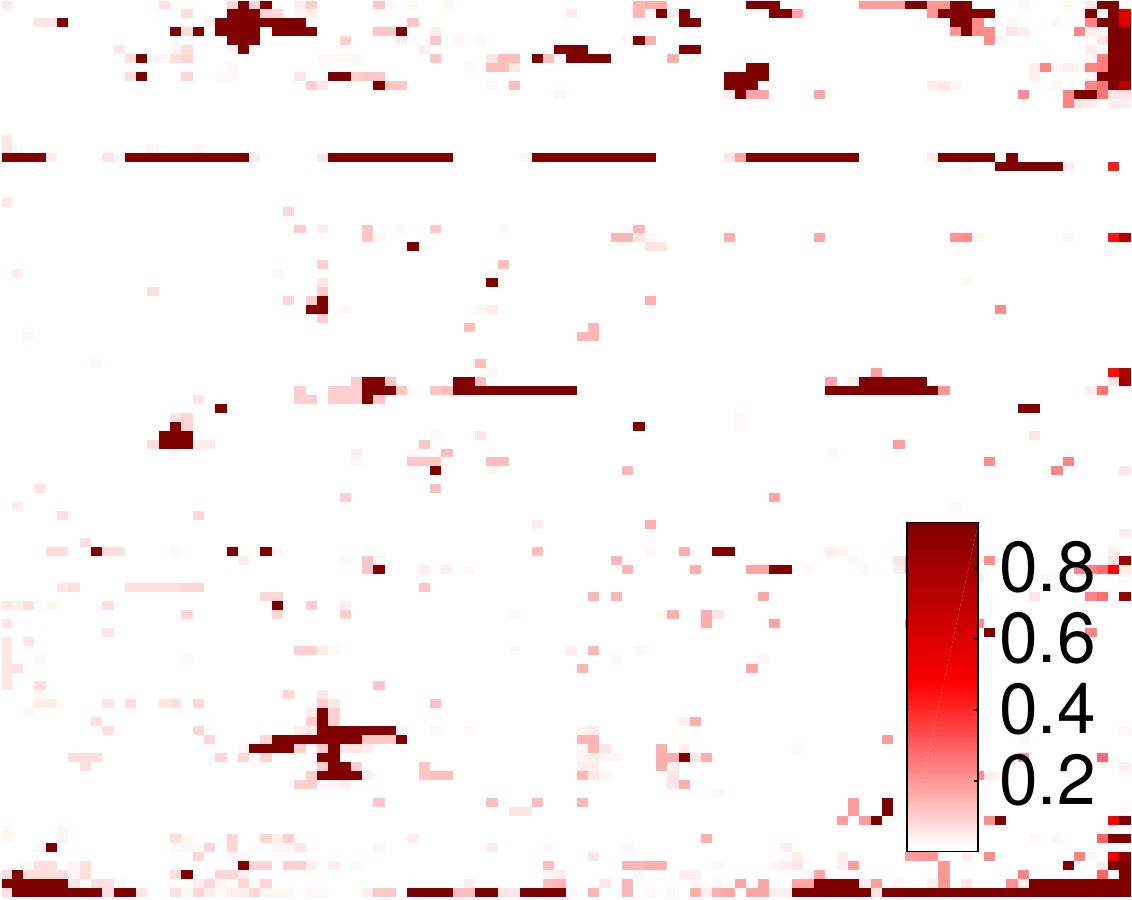}\\

  GulfPort & GT  & RX (0.90)& KRX (0.95)& RBIG  (0.95)& HYBRID (0.95)
 
\\

\\

    \end{tabular}
\end{center}
\caption{\small Anomaly detection predictions in four images (one per row). First column: Cat-Island, World Trade Center (WTC), Texas Coast and Pavia original datasets with anomalies outlined in green. Second column: represent the reference maps of each image. From third column to the last column: activation maps and the AUC values (in parenthesis) for the RX, {the best performance method among KRX, KDE and SVDD}, RBIG and the HYBRID model, respectively.}
\label{fig:AD}
\end{figure*}

We collected multispectral and hyperspectral images acquired by the AVIRIS and ROSIS-03 sensors.  Figure~\ref{fig:AD} showcases the scenes used in the experiments. The AD scenarios consider anomalies related to a diversity of problems: airplane, latent fires, urbanization and vehicle detection~\cite{Guo2016,ABU,jose}. 

\begin{table}[t]
\centering
\caption{Image attributes used for the experiments of anomaly detection (AD) and  change detection (CD). }\label{table:database}

\begin{tabular}{ |l|l|c|l|l|} 
\hline\hline
\rowcolor[HTML]{A9A9A9} {\bf Images} & {\bf Sensor}&{\bf Size} & {\bf Bands}  & {\bf SR [m]}\\ \hline \hline
\rowcolor[HTML]{D8D8D8} \multicolumn{5}{|l|}{AD}
\\ \hline \hline

Cat-Island & AVIRIS    & 150$\times$150      &  188  & 17.2 \\\hline
WTC & AVIRIS    & 200$\times$200     &  224  & 1.7\\\hline
Texas-Coast &   AVIRIS  & 100$\times$100     &  204  &17.2 \\\hline
GulfPort &  AVIRIS   & 100$\times$100      & 191  &3.4 \\\hline
\hline \hline

\rowcolor[HTML]{D8D8D8}\multicolumn{5}{|l|}{CD}\\ \hline \hline                   

Texas           & Cross-Sensor    & 301$\times$201      & 7    & 30
\\\hline
Argentina           & Sentinel-2    & 1257$\times$964     &12     & 10-60
\\\hline
Chile          & Landsat-8    &  201$\times$251     & 12     &   10-60       
\\\hline
Australia  & Sentinel-2       & 1175$\times$2031       &12   & 10-60
\\\hline            

\hline
\end{tabular}

\end{table}

\begin{figure*}[t]
    \centering
    
    \subfloat[Cat-Island]{\includegraphics[width = 4cm,height = 3.5cm]{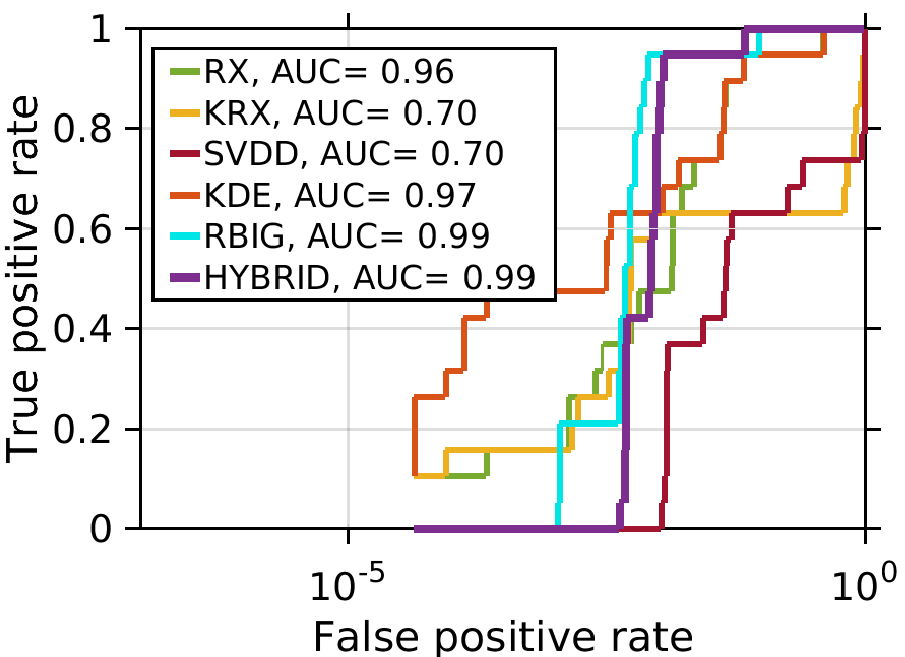}}
    \subfloat[WTC]{\includegraphics[width = 4cm,height = 3.5cm]{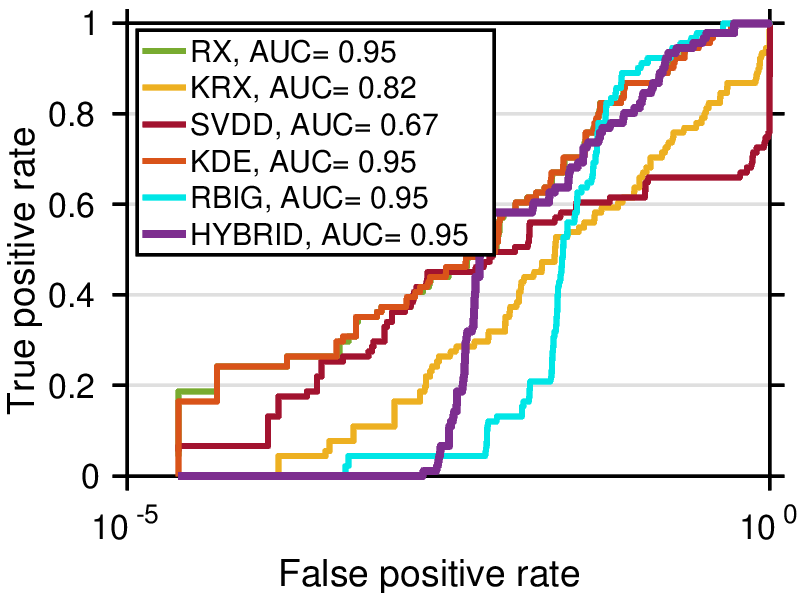}}
    \subfloat[Texas Coast]{\includegraphics[width = 4cm,height = 3.5cm]{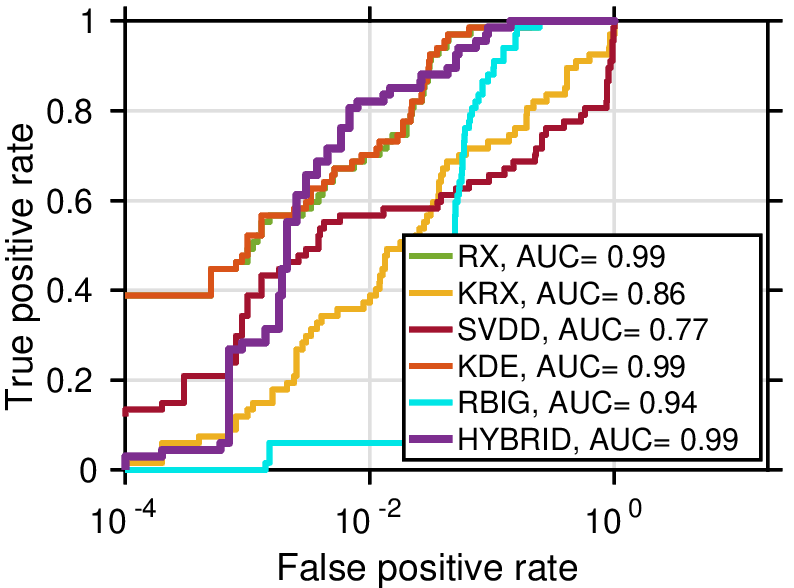}}
    \subfloat[GulfPort]{\includegraphics[width = 4cm,height = 3.5cm]{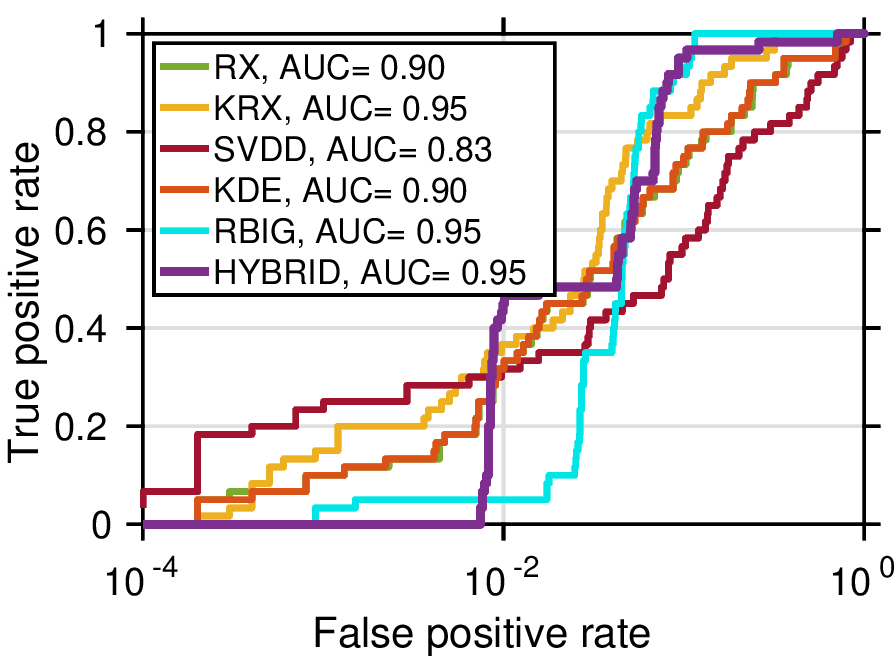}}
    
    \caption{Anomaly detection ROC curves in linear scale for all scenes. Numbers in legend display the AUC values for each method.}
    \label{fig:ROC_AD}
\end{figure*}

\begin{figure*}[]
    \centering
    
   \subfloat[Cat-Island]{\includegraphics[width = 4cm]{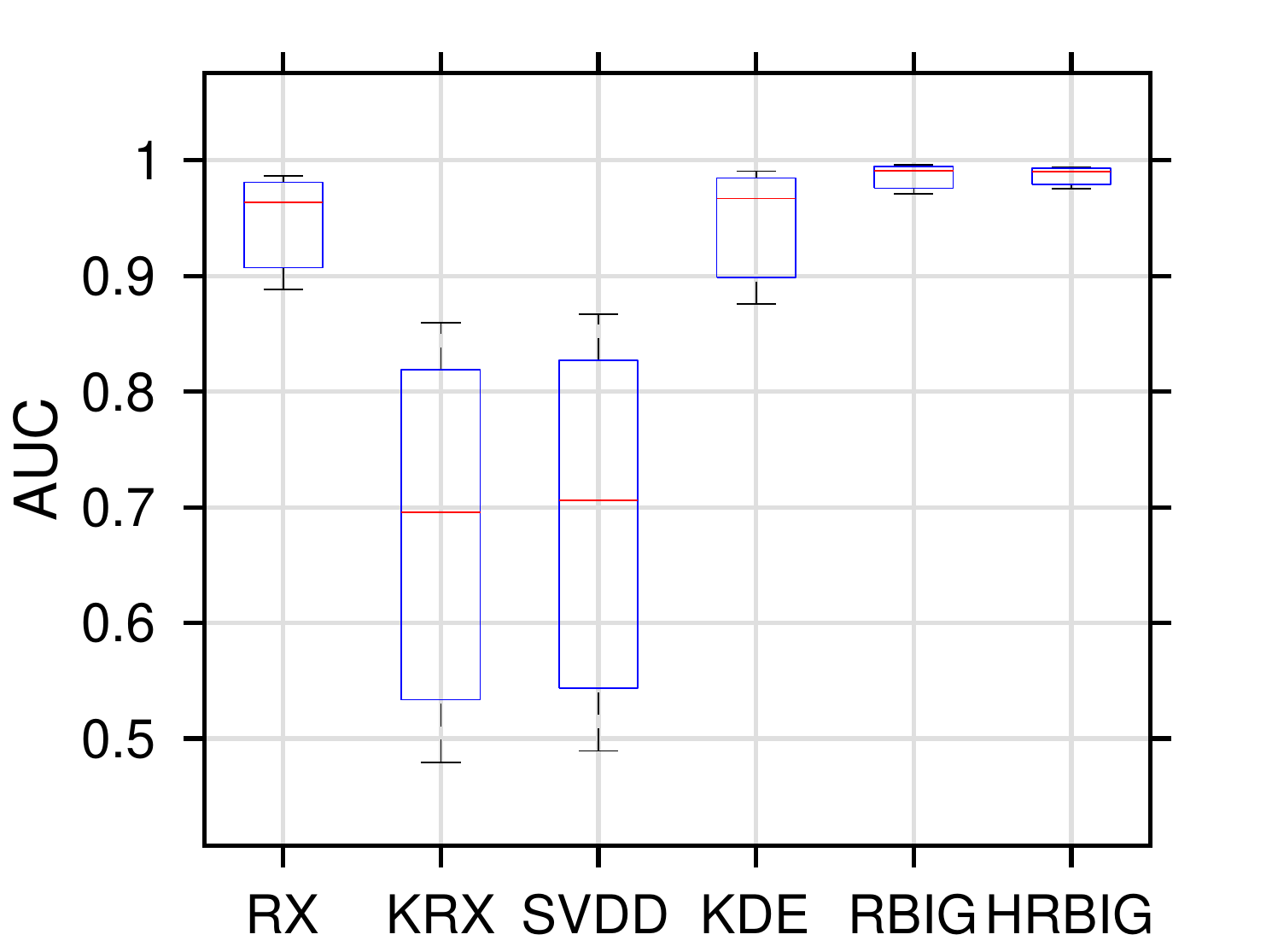}}
    \subfloat[WTC]{\includegraphics[width = 4cm]{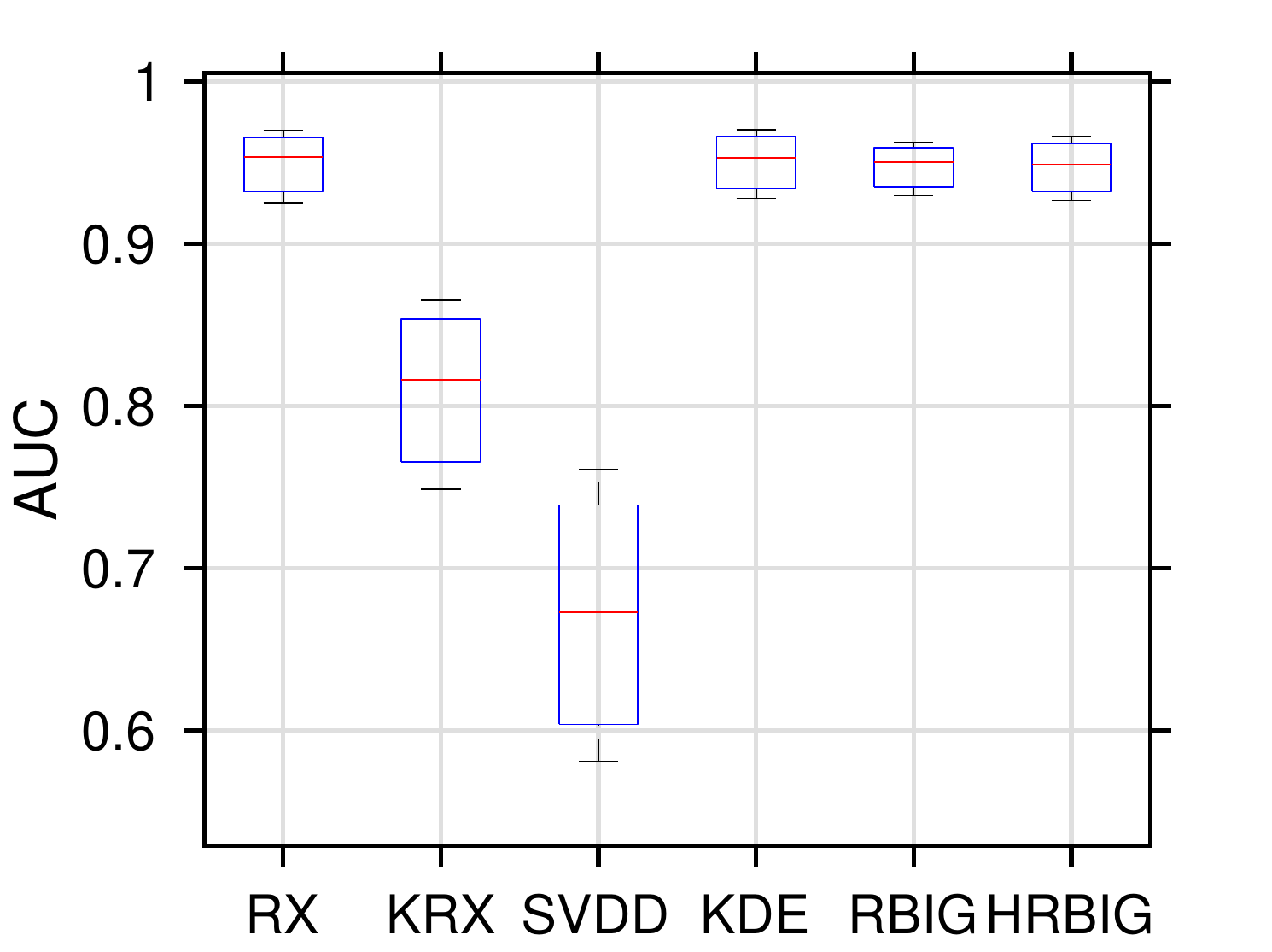}}
    \subfloat[Texas Coast]{\includegraphics[width = 4cm]{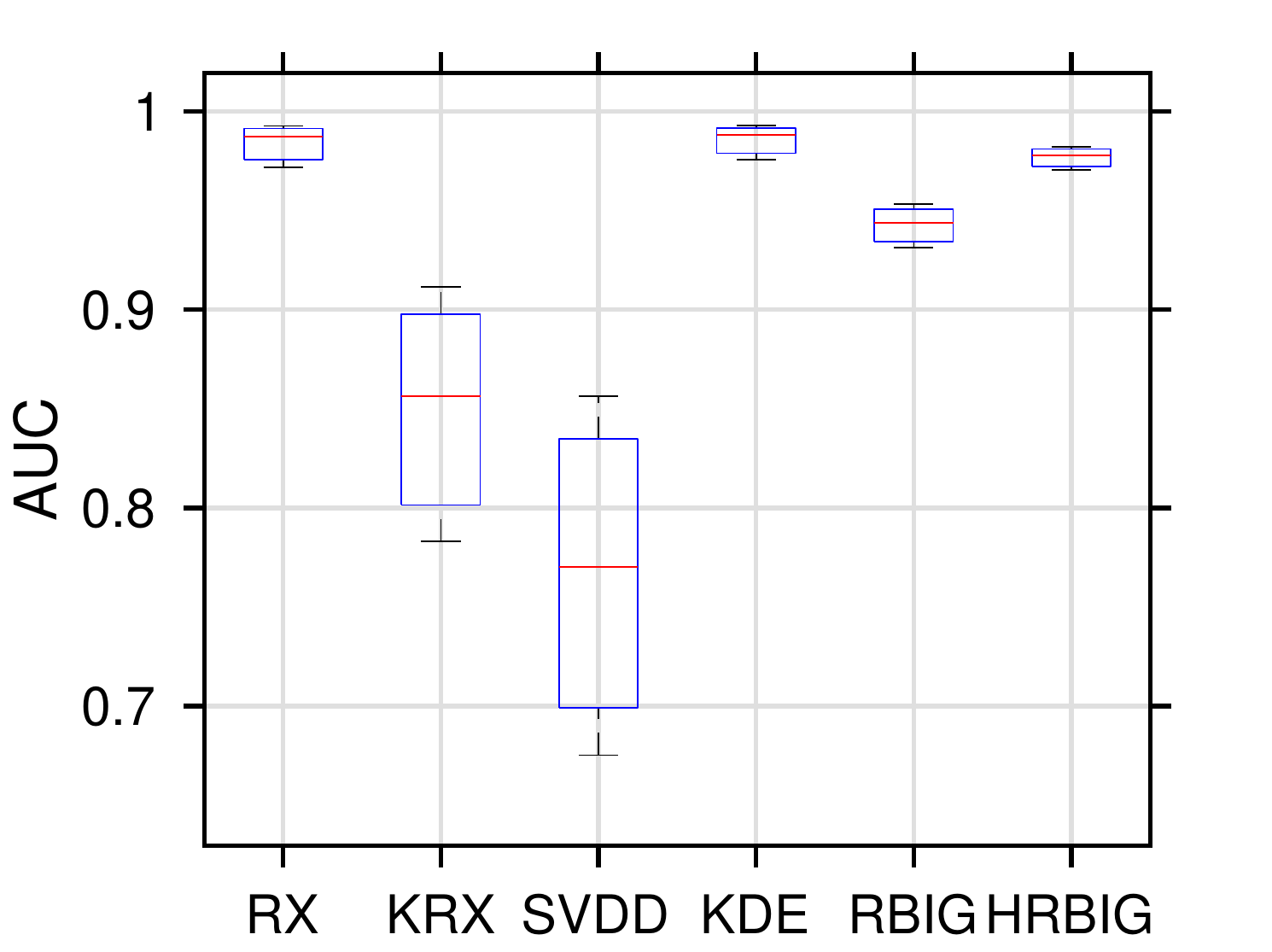}}
    \subfloat[GulfPort]{\includegraphics[width = 4cm]{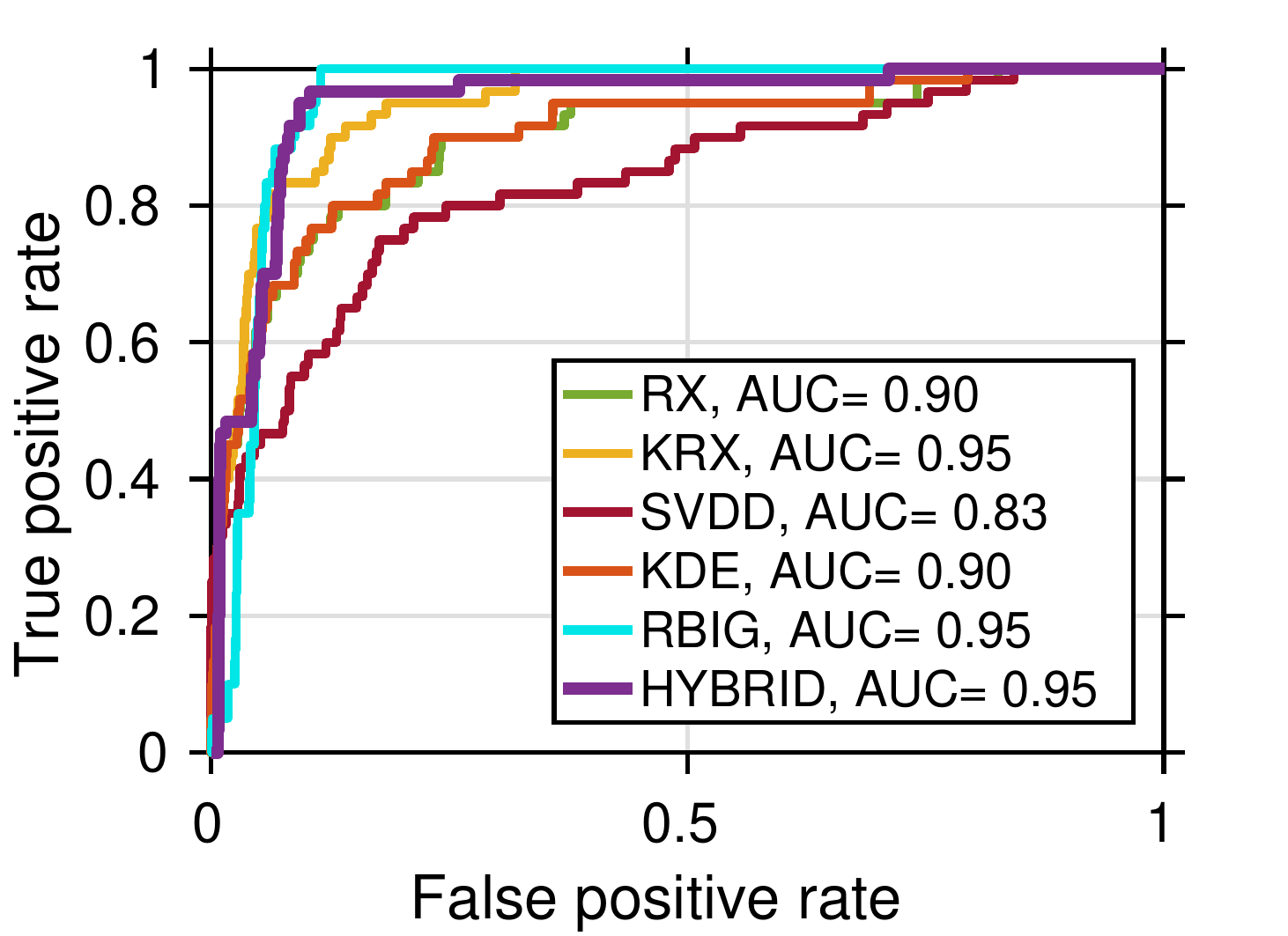}}
    
    \caption{Anomaly detection results of the bootstrap experiment for 1000 experiments. AUC values and standard deviation for each method are shown as boxplot, red line represent the median value, the blue box contains 95\% of the values, black lines represent the maximum and minimum values.}
    \label{fig:BOXPLOT_AD}
\end{figure*}

The Cat-Island dataset corresponds to the airplane captured flying over the beach and it is considered a strange object when compared to the rest of the image (a white spot in the middle of a beach) and the percentage of anomalies represent the 0.09\% of the scene. The World Trade Center (WTC) image was collected by the Airborne Visible Infra-Red Imaging Spectrometer (AVIRIS) over the WTC area in New York on 16 September 2001 (after the collapse of the towers in NY). The data set covered the hot spots corresponding to latent fires at the WTC, which can be considered as anomalies and it represent the 0.23\% of the scene. In the Texas Coast dataset, the anomalies represent the 0.67\% of the scene and the image contains roofs built on a wooded site and bright spots that reflect light which can be considered an anomaly. The GulfPort dataset correspond to a battery of airplanes taxied on the runway and the pecentage of anomalies represent the 0.60\% of the scene. 

\begin{figure*}[t!]
\setlength{\tabcolsep}{2pt}
\begin{center}
  \begin{tabular}{ccccc}
       
       \includegraphics[width = 3cm ,height = 3cm]{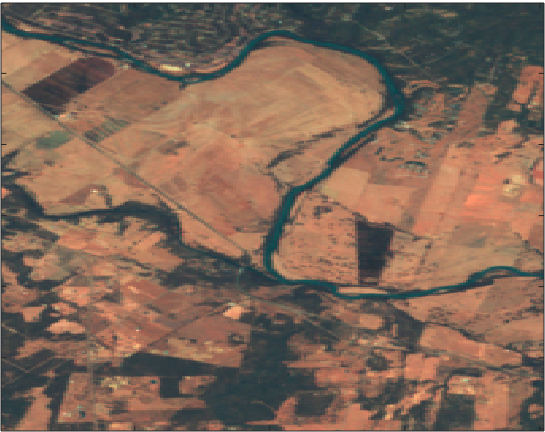}&
       \includegraphics[width = 3cm ,height = 3cm]{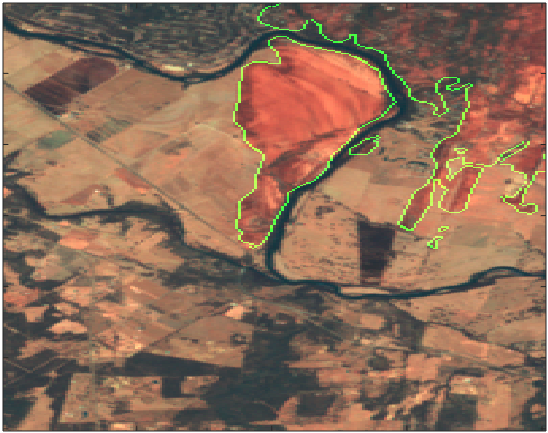}&
       \includegraphics[width = 3cm ,height = 3cm]{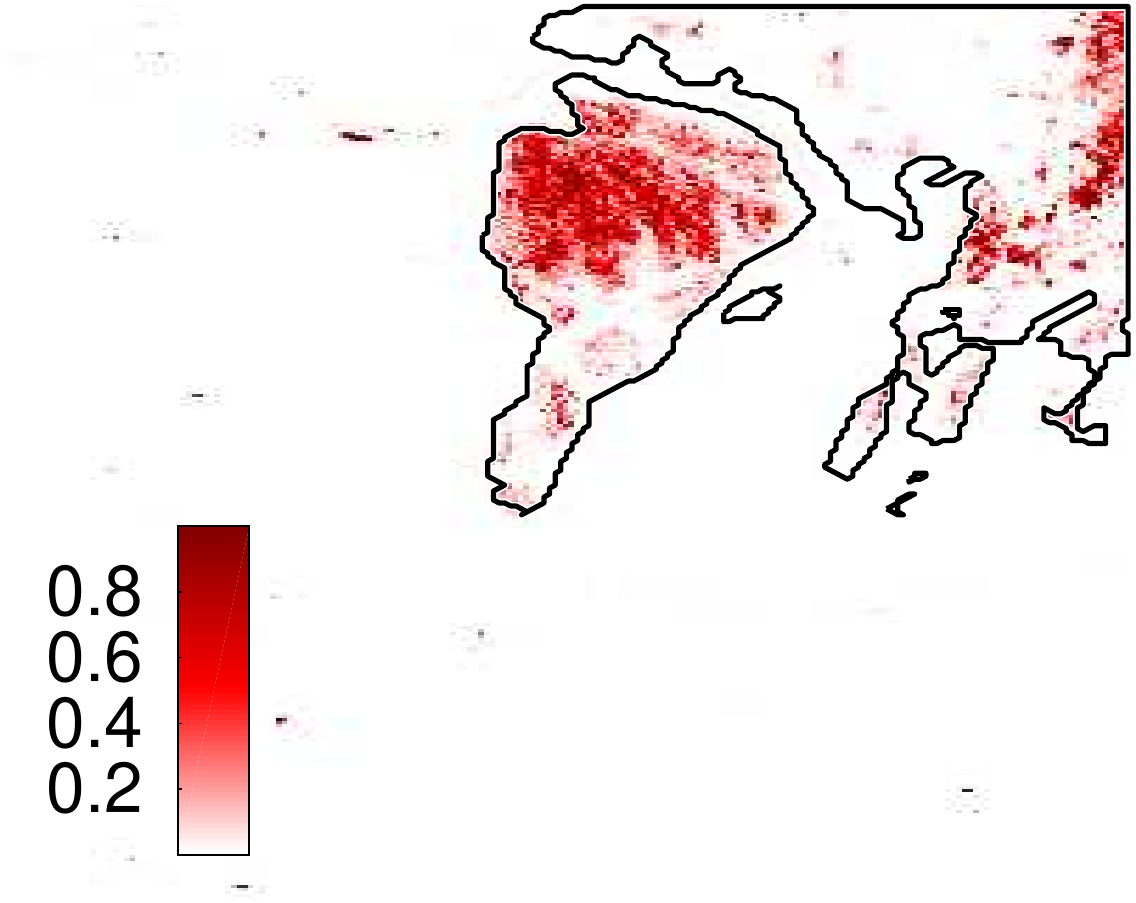}&
       \includegraphics[width = 3cm ,height = 3cm]{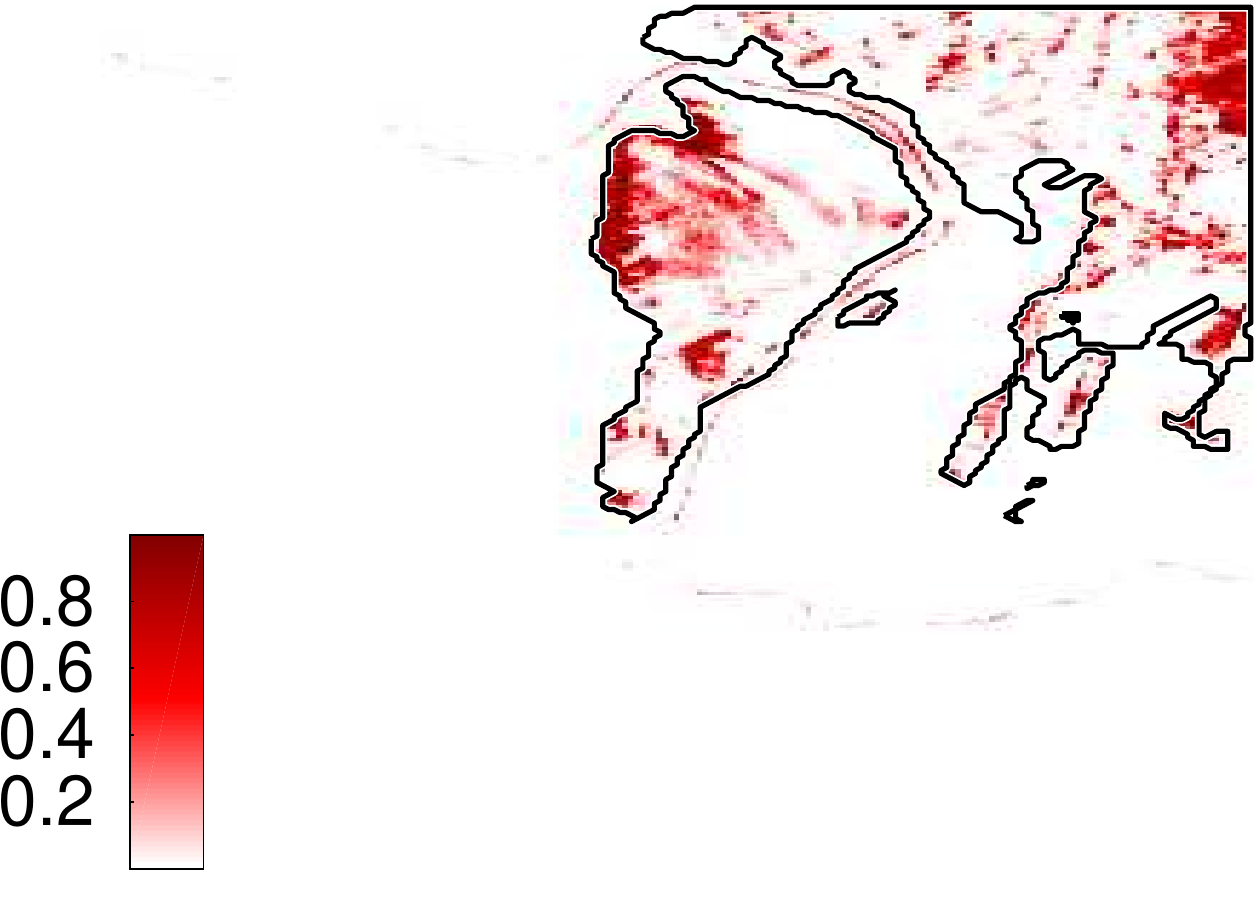}&
       \includegraphics[width = 3cm ,height = 3cm]{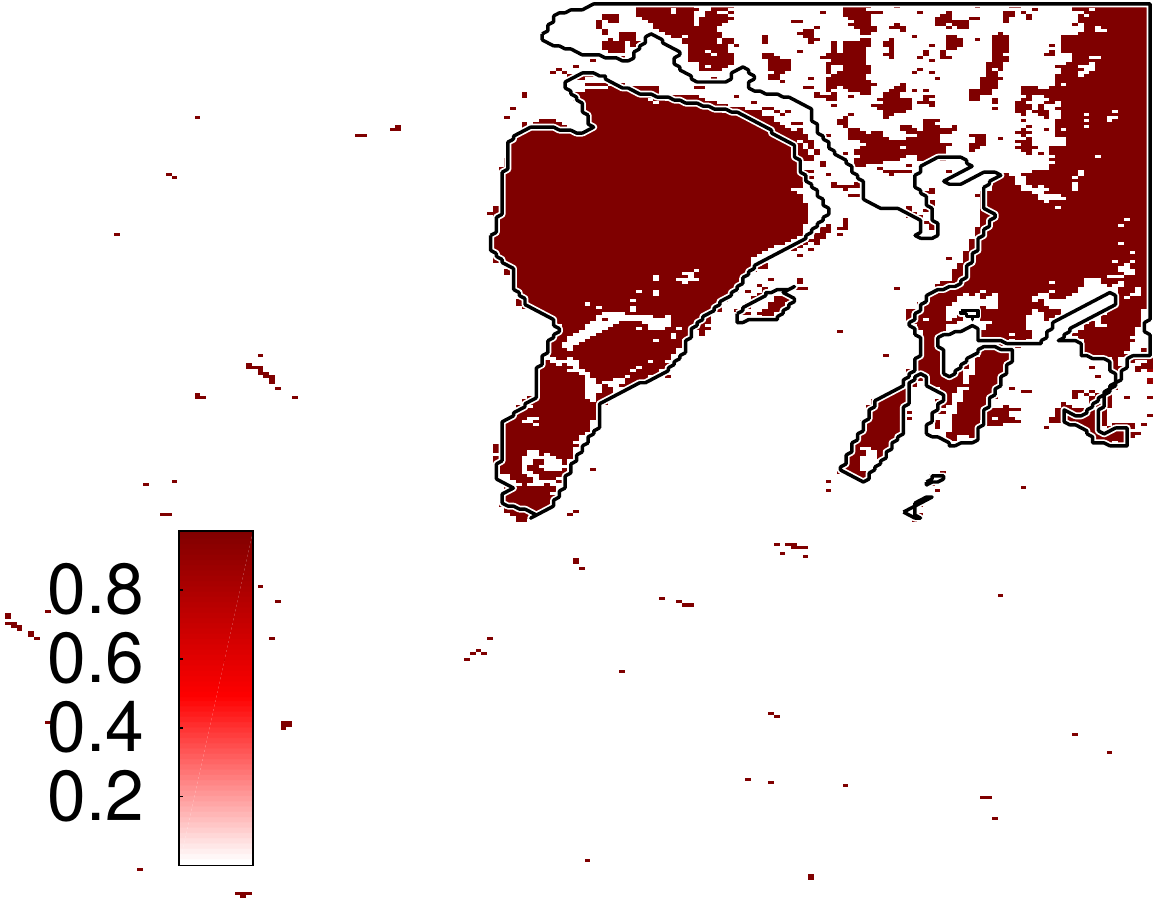}\\
       Texas ($t_1$) & Texas ($t_2$) & RX(0.91) & KSVDD(0.95) & RBIG(0.98) \\

\includegraphics[width = 3cm , height = 3cm]{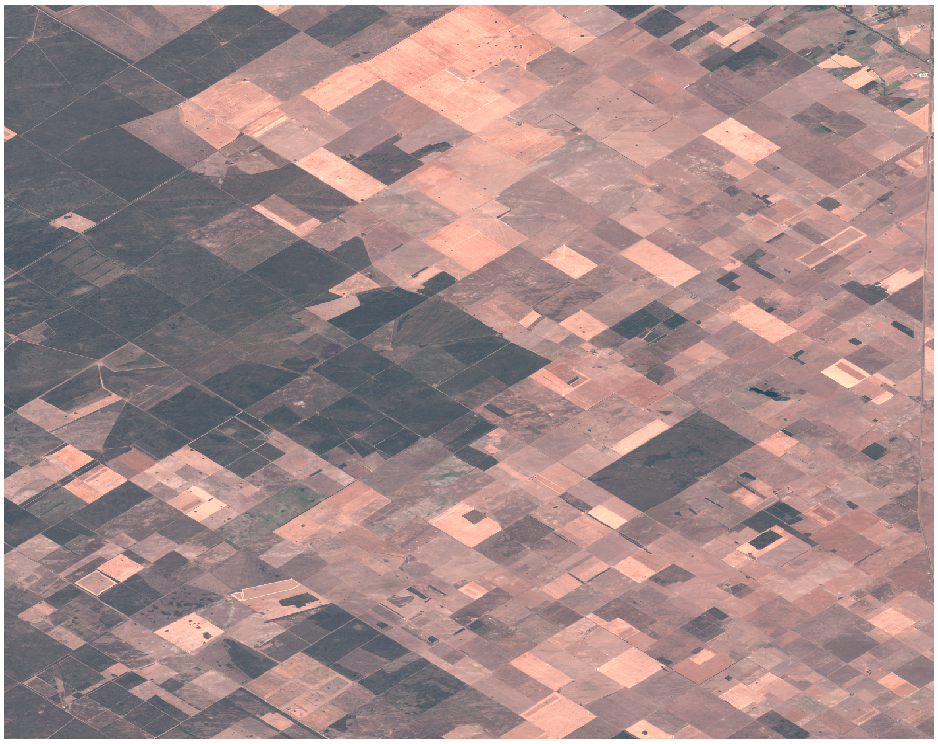} &
\includegraphics[width = 3cm , height = 3cm]{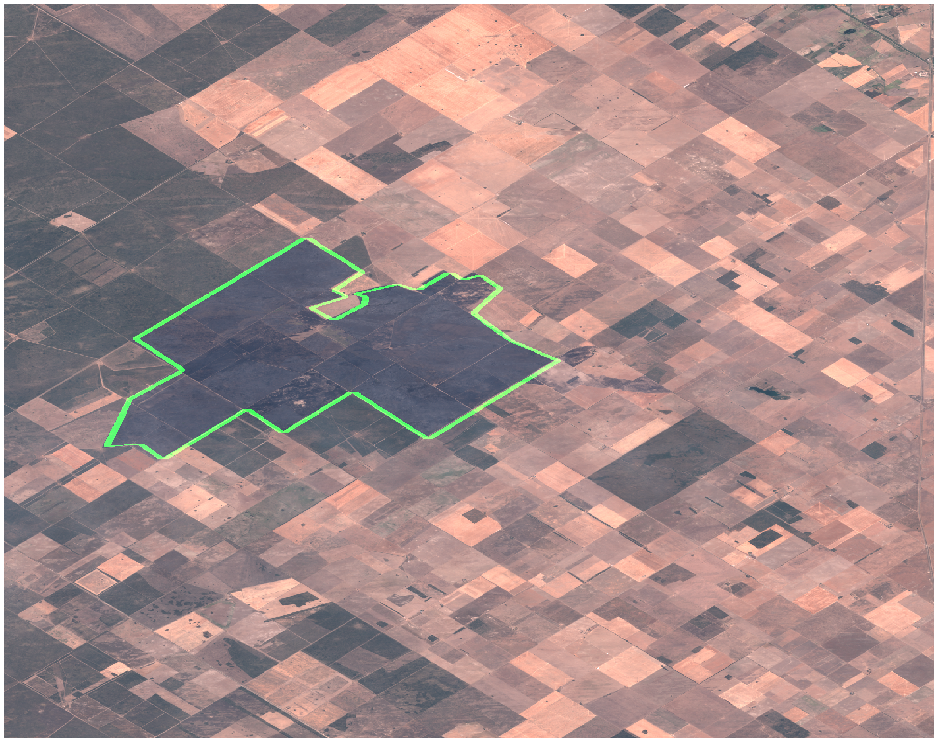}& 
\includegraphics[width = 3cm, height = 3cm]{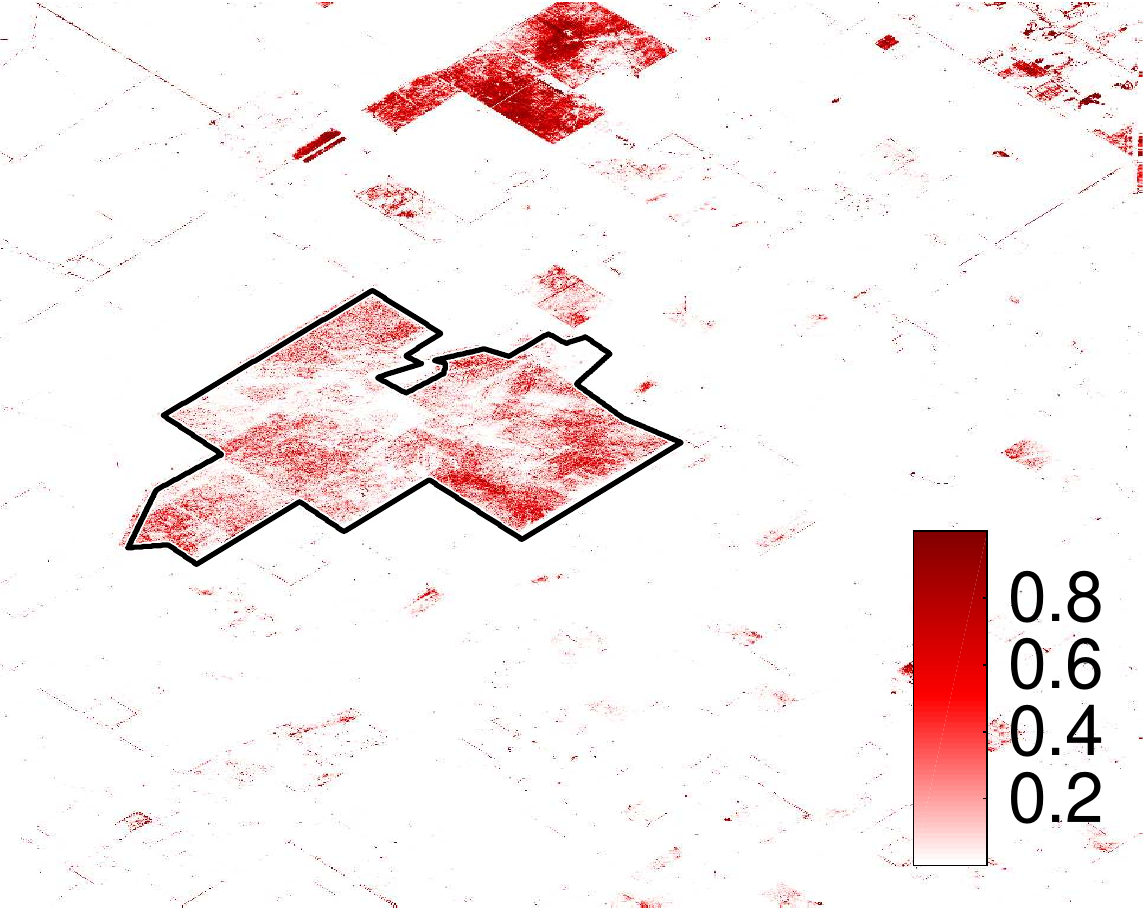}&
\includegraphics[width = 3cm, height = 3cm]{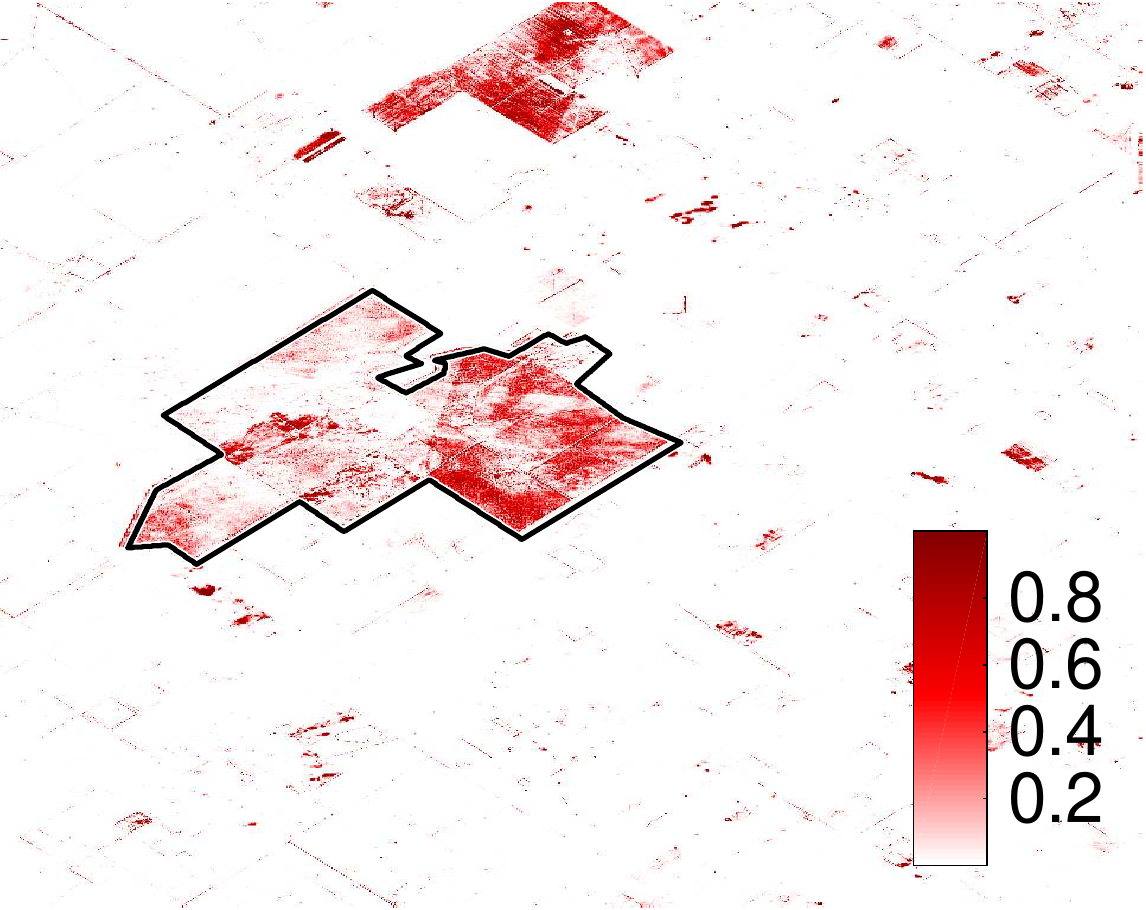}&
\includegraphics[width = 3cm , height = 3cm]{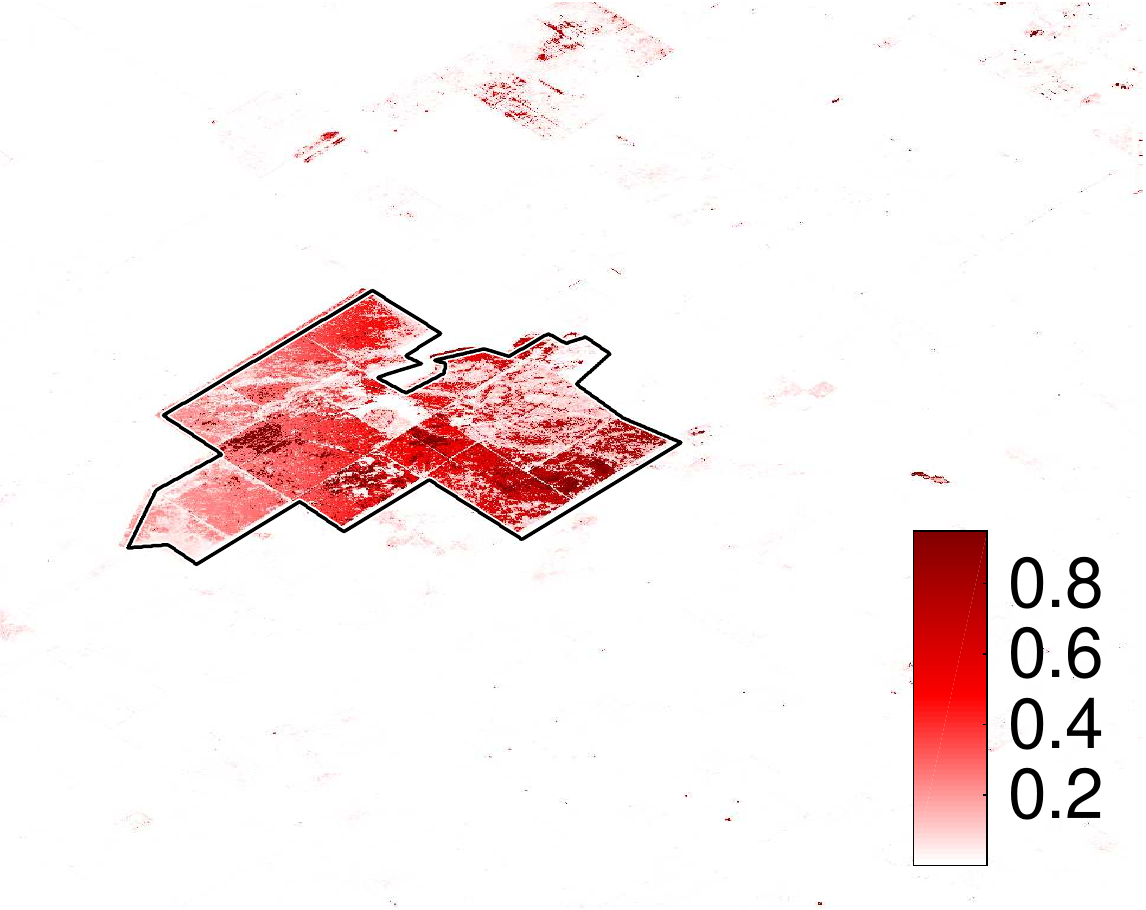}\\
Argentina ($t_1$) & Argentina ($t_2$) & RX (0.94) & K-RX (0.93) & RBIG  (0.97) \\

\includegraphics[width = 3cm , height = 3cm]{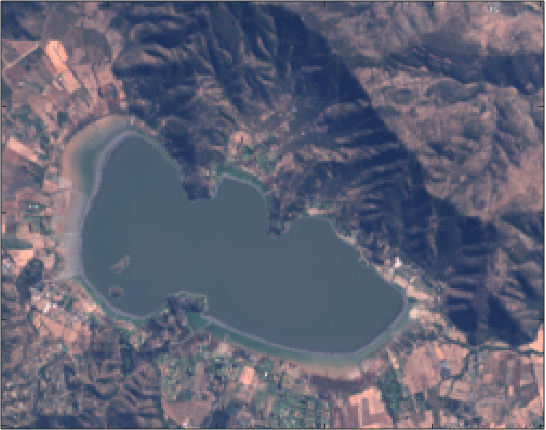}&
\includegraphics[width = 3cm , height = 3cm]{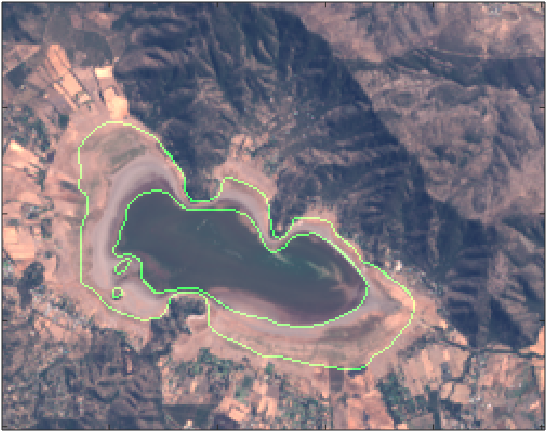}&
\includegraphics[width = 3cm , height = 3cm]{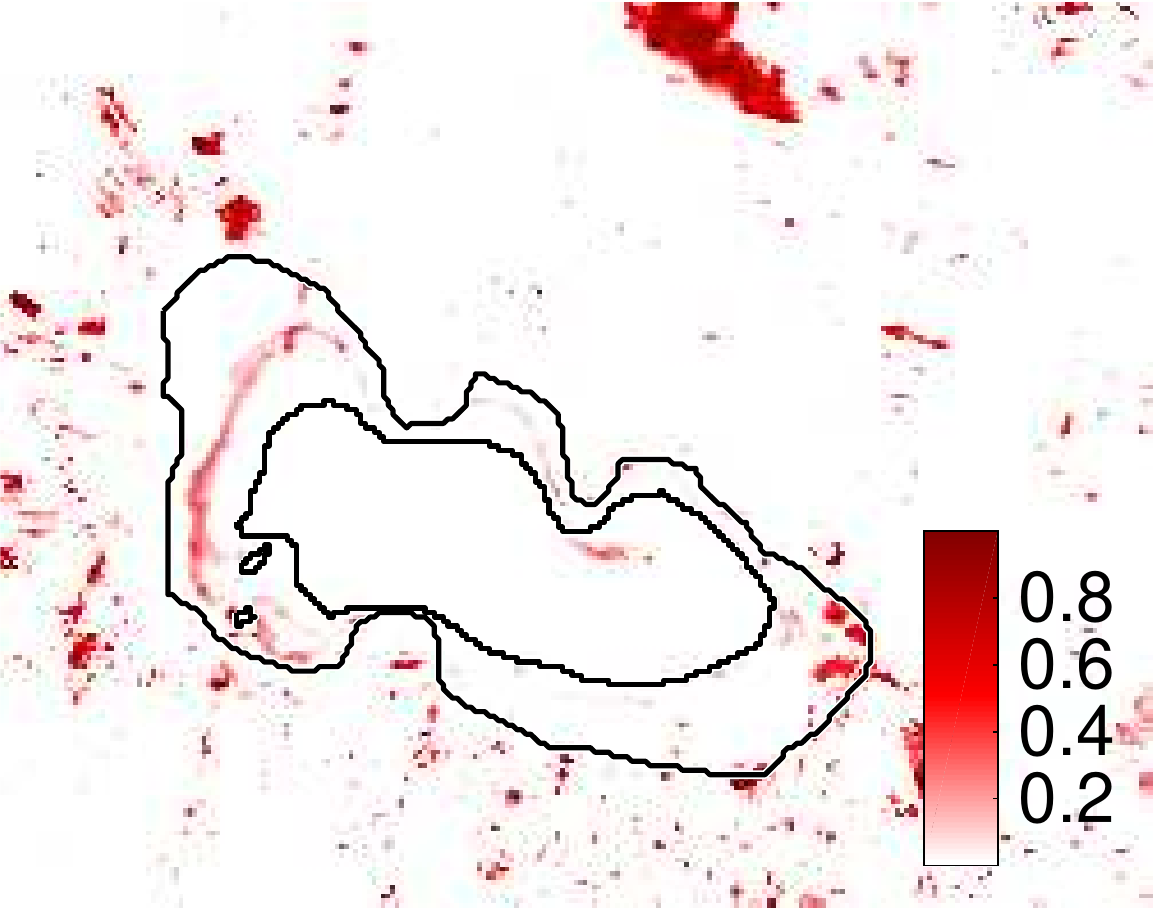}&
\includegraphics[width = 3cm , height = 3cm]{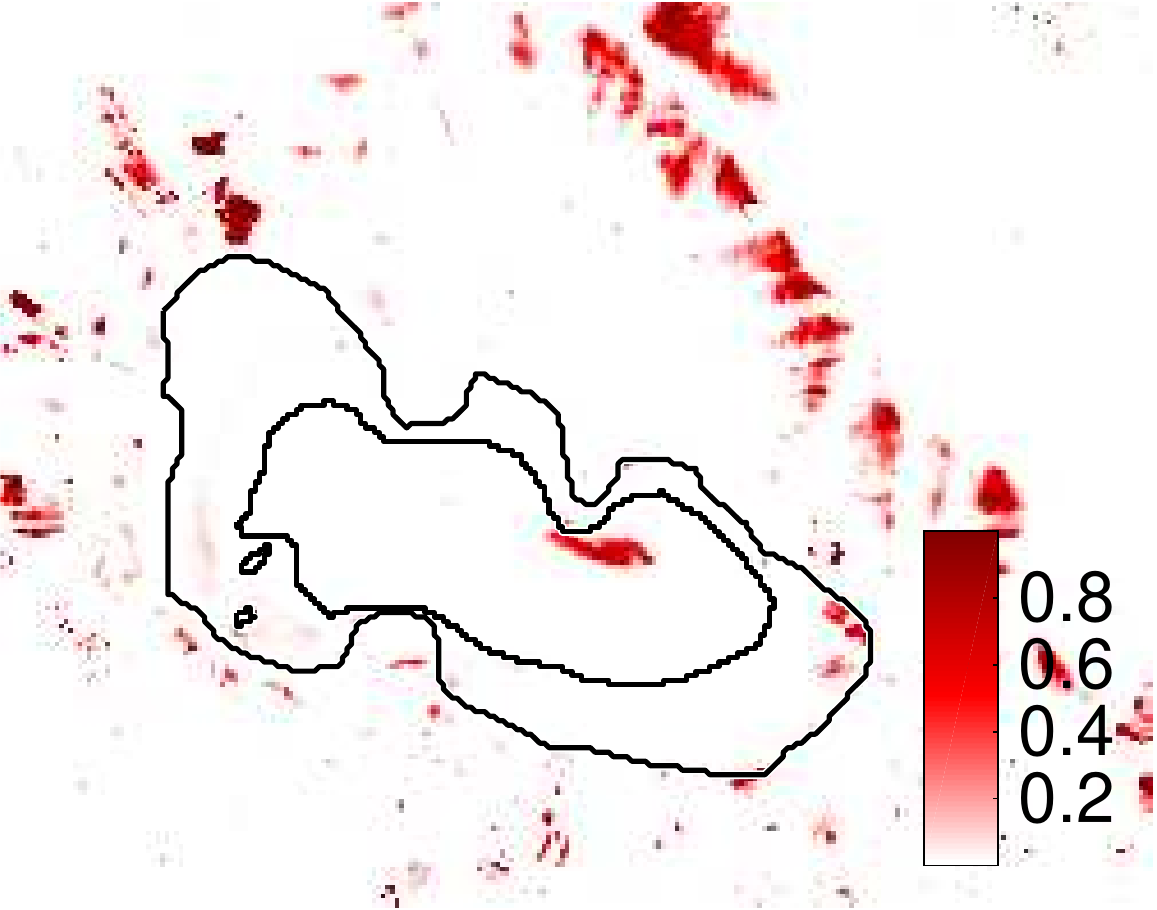}&
\includegraphics[width = 3cm , height = 3cm]{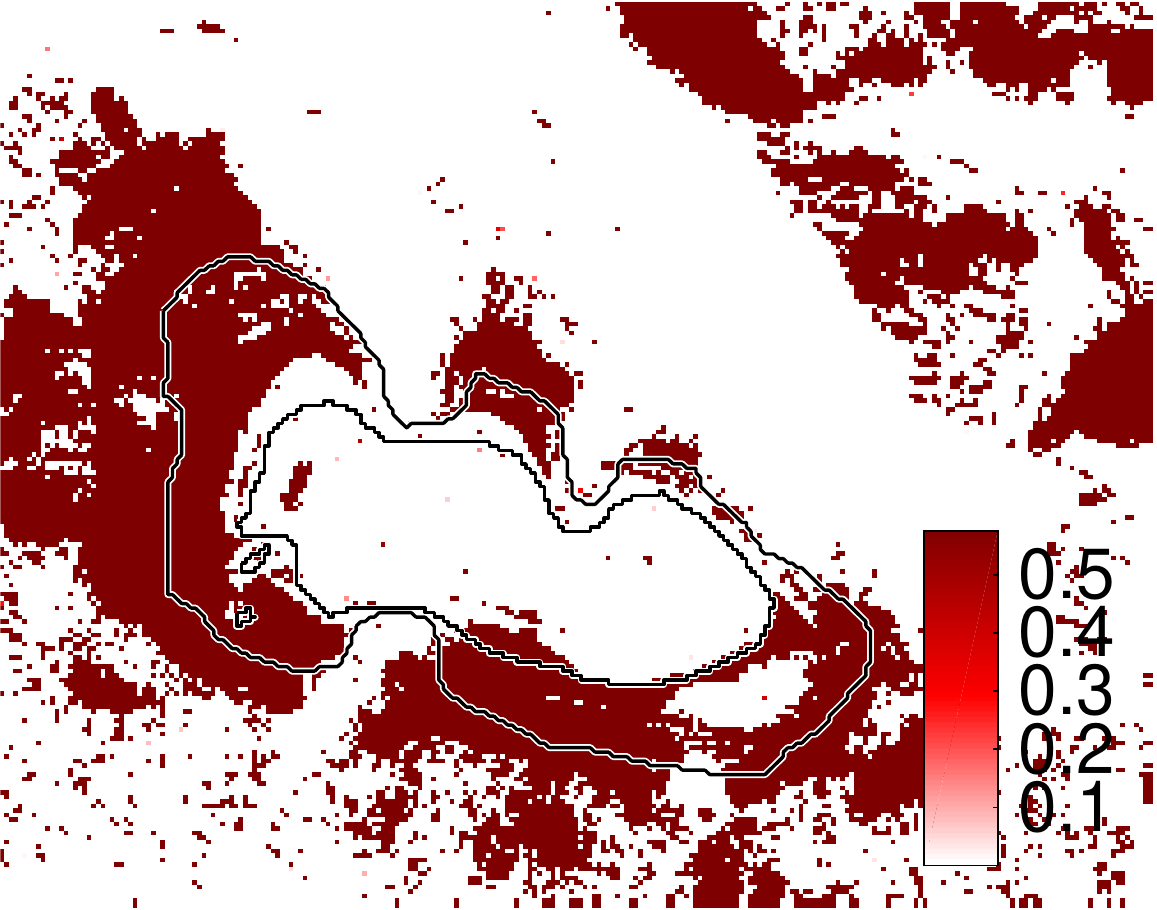} \\
Chile ($t_1$) & Chile ($t_2$) &RX  (0.64)&KRX (0.66)& RBIG (0.72)

\\

\includegraphics[width = 3cm , height = 3cm]{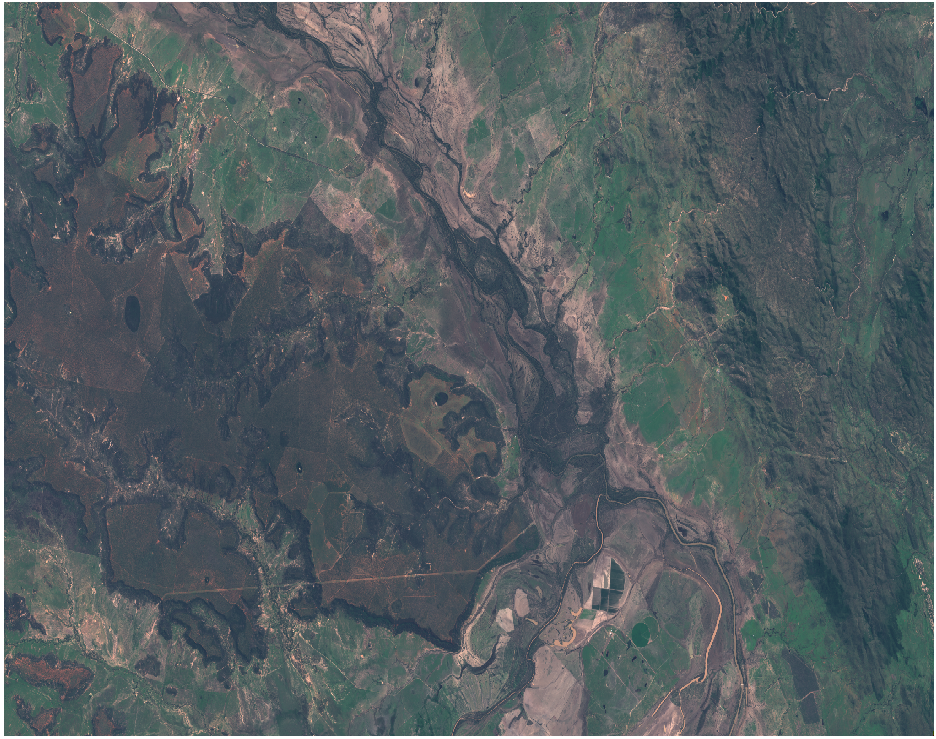}&
\includegraphics[width = 3cm , height = 3cm]{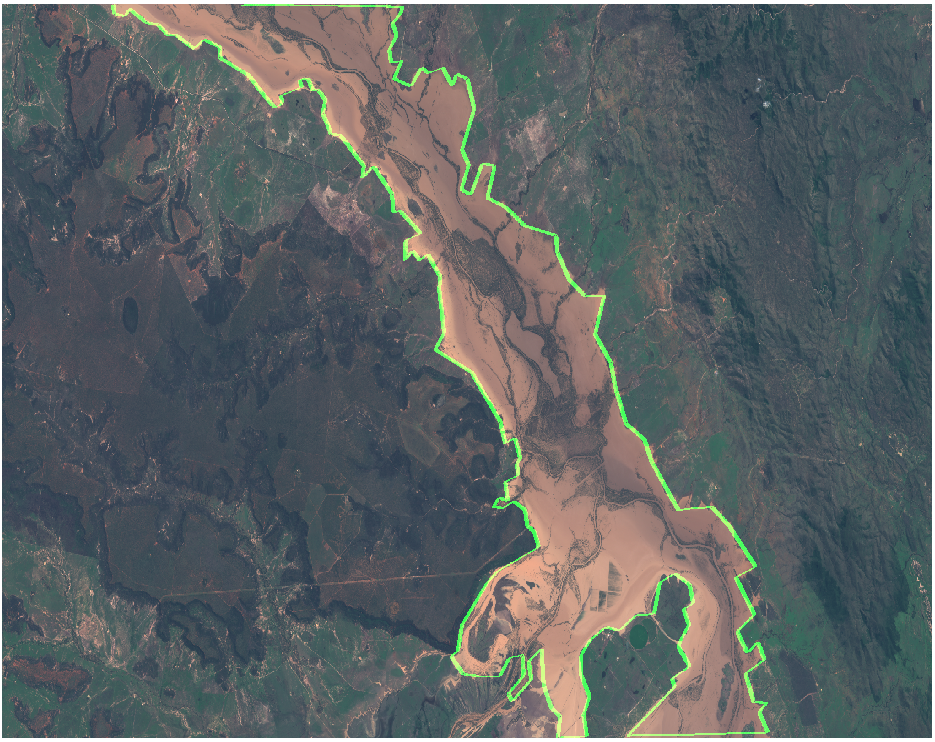}&
\includegraphics[width = 3cm , height = 3cm]{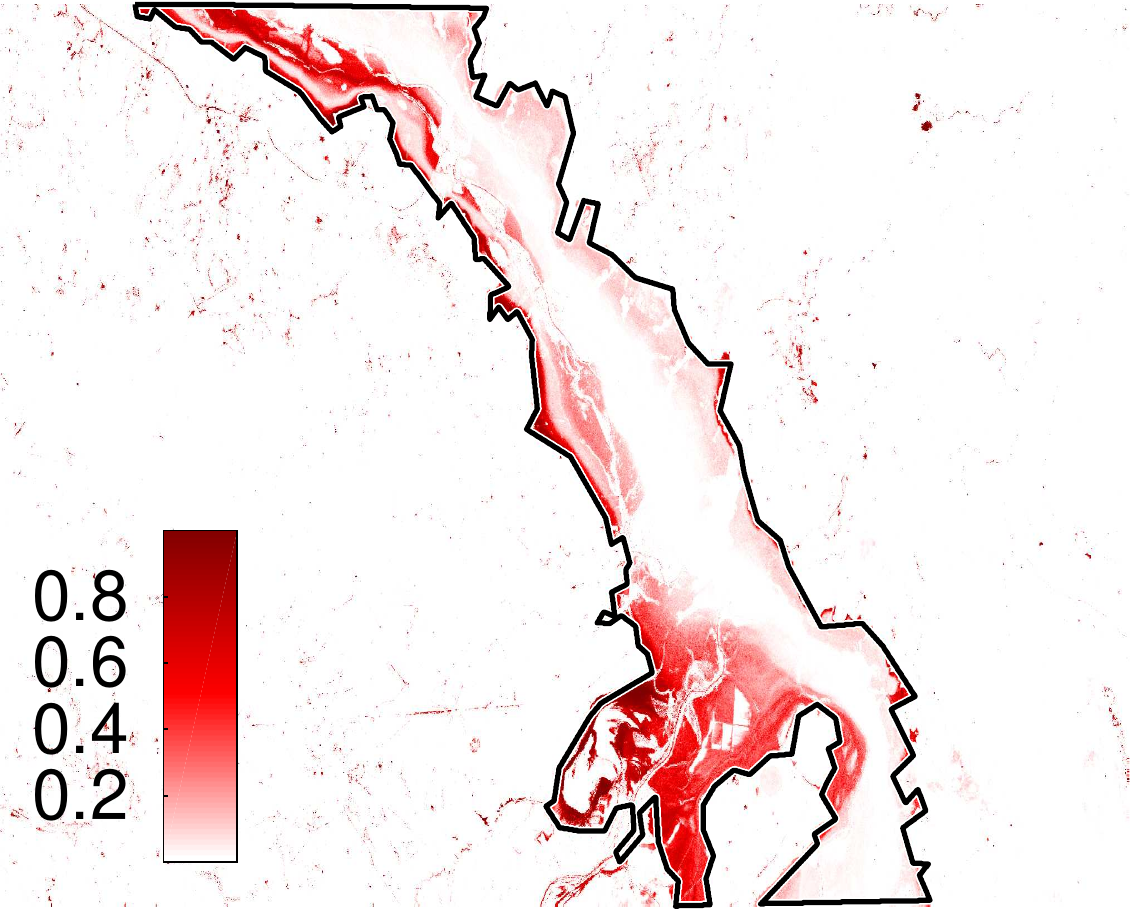}
&
\includegraphics[width = 3cm , height = 3cm]{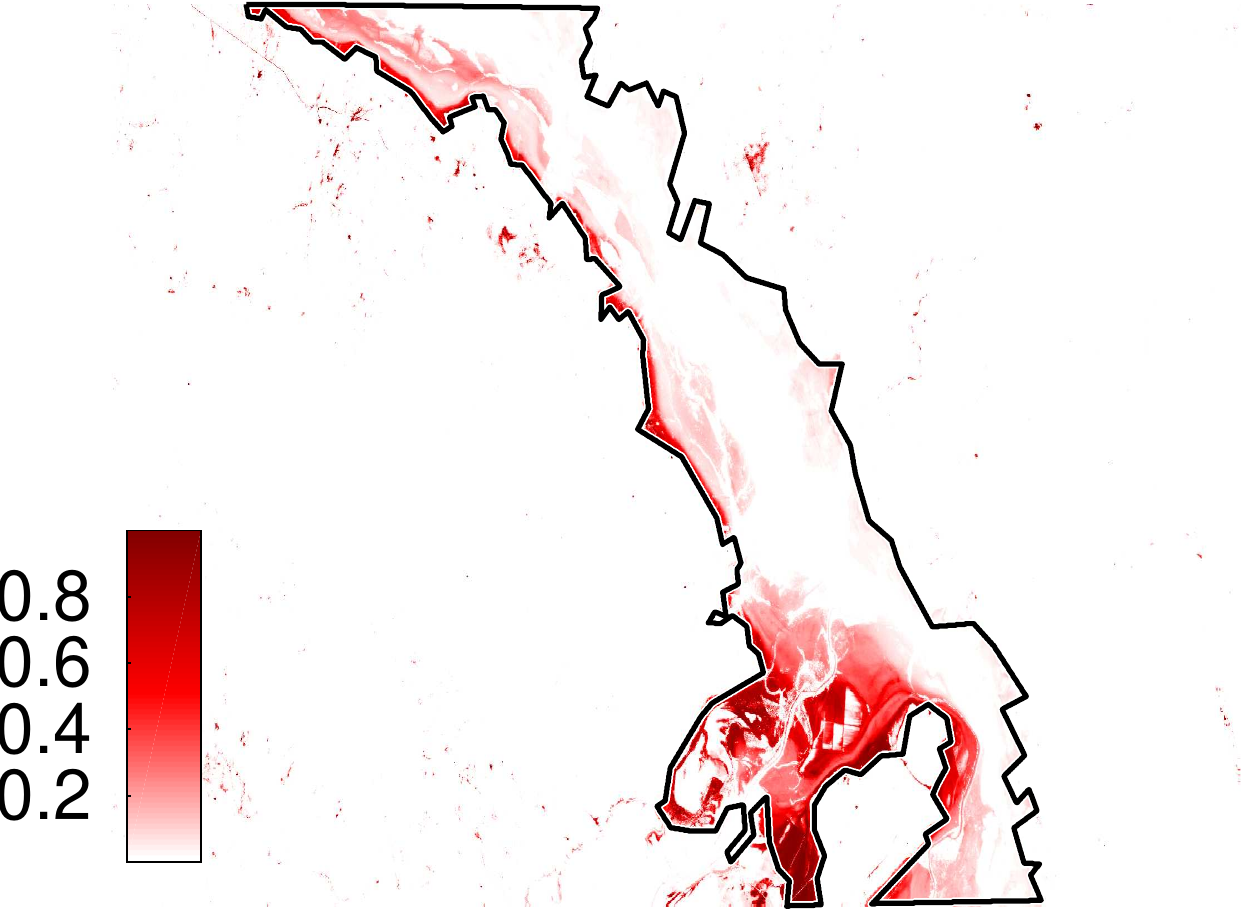}
& \includegraphics[width = 3cm , height = 3cm]{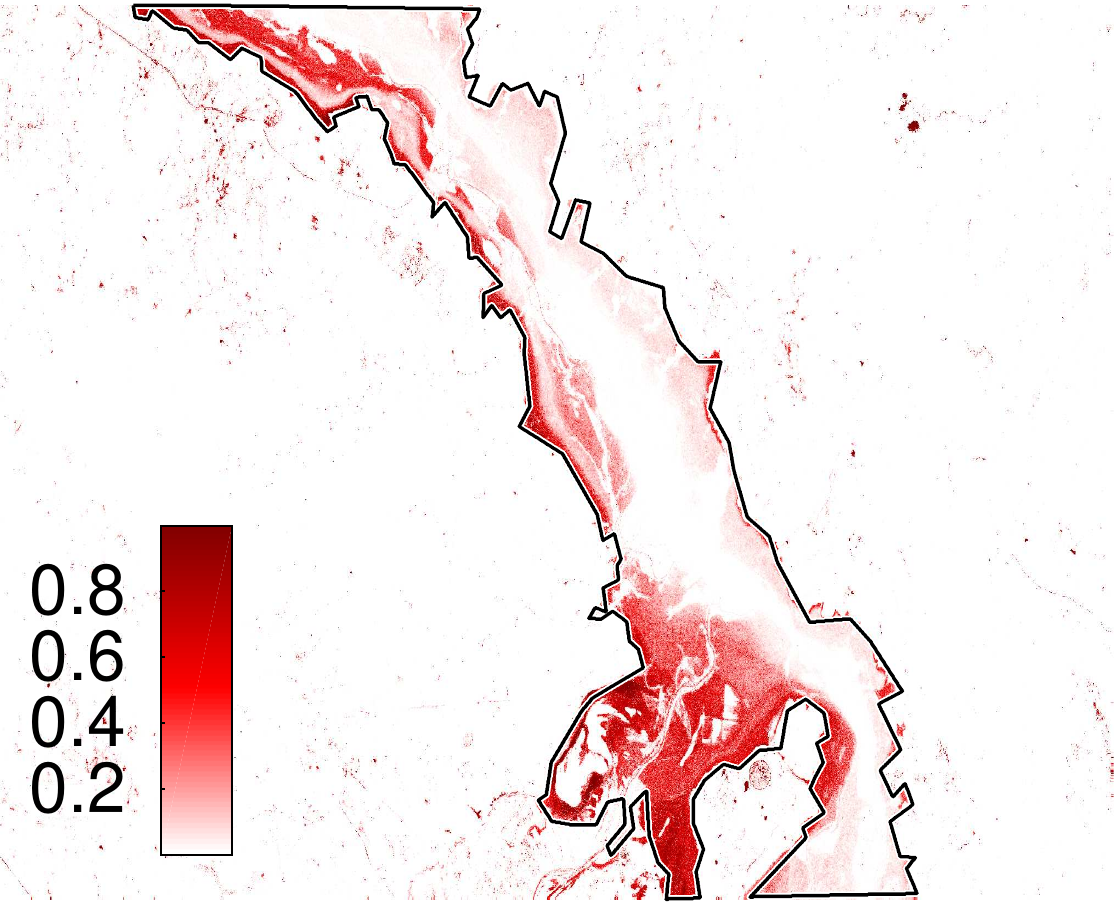}\\

Australia ($t_1$) &Australia ($t_2$) &RX (0.86)&KSVDD (0.93) &RBIG (0.93)

    \end{tabular}
\end{center}
\caption{Change detection results for different images. First two columns show the images before and after the change, with the changed region highlighted in green. Columns three to five show the prediction maps for the different methods, the amount of change detected in each pixel is colored from white (less) to red (more). AUC values are given in parenthesis. The changed region is outlined in black to facilitate the visual inspection.}
\label{fig:CD}
\end{figure*}

\subsubsection{Numerical and Visual Comparison}

It is important to take into consideration that KRX requires the selection of some hyperparameters, being the kernel parameter the most critical one. In order to perform a fair comparison while staying in an unsupervised learning setting, we use the standard RBF kernel function, $k({a},{b}) = \exp(-\|{a}-{b}\|^2/(2\sigma^2))$ and set the lengthscale parameter $\sigma$ to the median distance between all examples. 
 
\begin{table}[b!]
    \centering
     \caption{AUC results for Anomaly Detection images. The value for the best method for each image is in bold.}
     \scriptsize
    \begin{tabular}{|l|l|l|l|l|l|l|}
    \hline\hline
    \rowcolor[HTML]{A9A9A9}{{\bf METHODS}} &{\bf RX}&{\bf K-RX}&{\bf SVDD}&{\bf KDE}&{\bf RBIG}&{\bf HYBRID}\\ \hline\hline
       Cat-Island   &\it{0.96} &\it{0.70}&\it{0.70}&\it{0.97} &{\bf 0.99} &{\bf 0.99}  \\ \hline
                
        WTC &{\bf 0.95} &\it{0.82} &\it{0.67}&{\bf 0.95}&{\bf 0.95} &{\bf 0.95} \\ \hline
         
        Texas-Coast &{\bf 0.99} &\it{0.86} &\it{0.77}&{\bf 0.99}&\it{0.94} &{\bf 0.99} \\ \hline
          GulfPort &\it{0.90} &{\bf 0.95}&\it{0.83}&\it{0.90} &{\bf 0.95} &{\bf 0.95}\\  \hline
         \hline
         
    \end{tabular}
   
    \label{tab:AD}
\end{table}

\begin{figure*}[h]
\captionsetup[subfloat]{position=top}
    \centering
    \subfloat[Texas]{\includegraphics[width = 4cm]{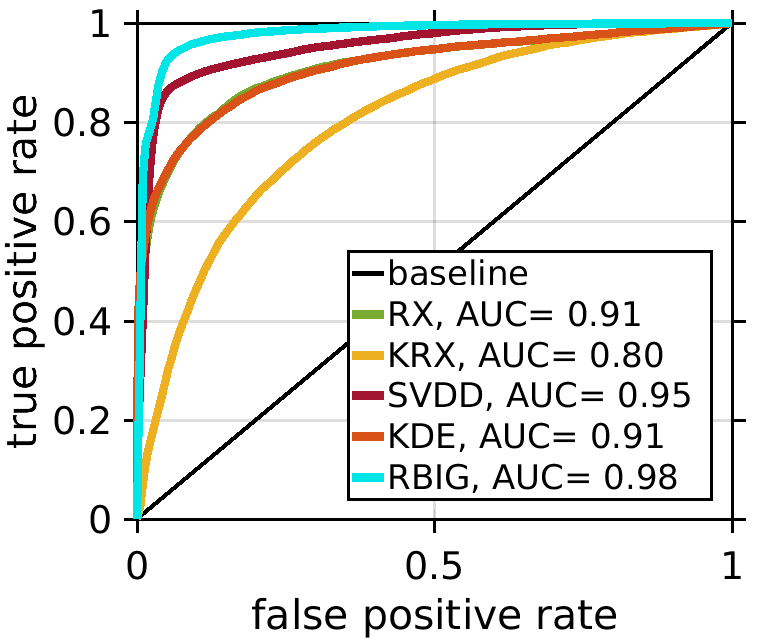}} 
    \subfloat[Argentina]{\includegraphics[width = 4cm]{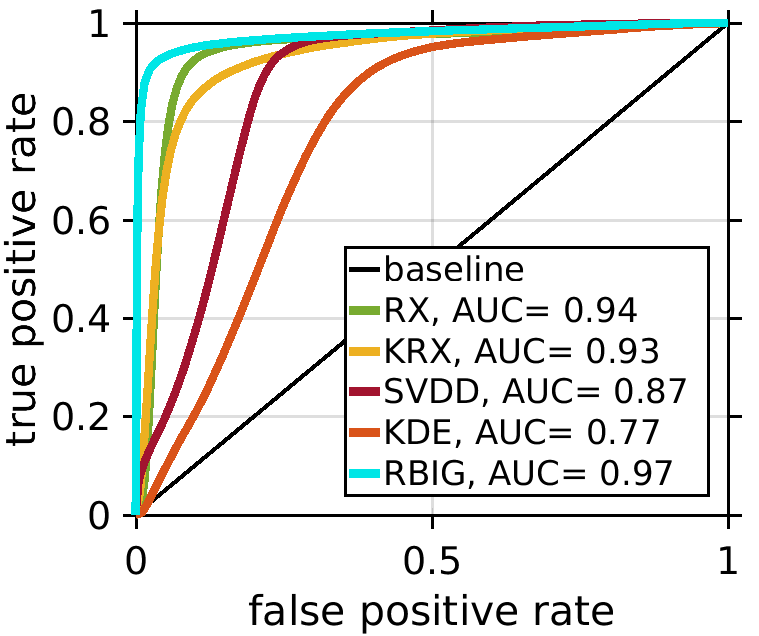}}
    \subfloat[Chile]{\includegraphics[width = 4cm]{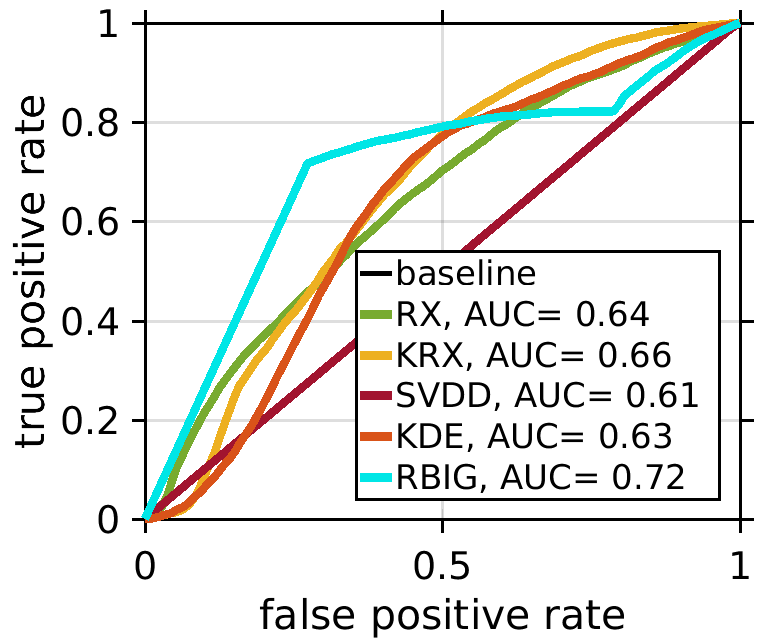}}
    \subfloat[Australia]{\includegraphics[width = 4cm]{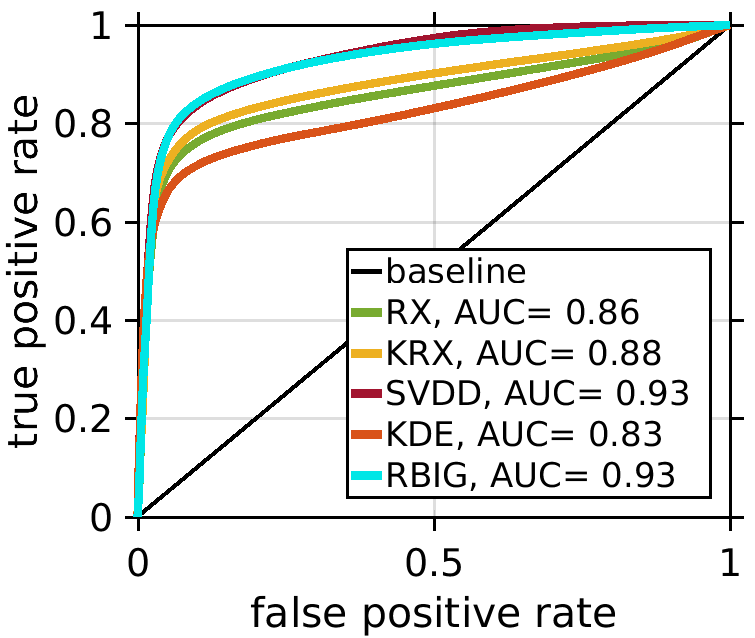}}
    \\
    
    {\includegraphics[width = 4cm]{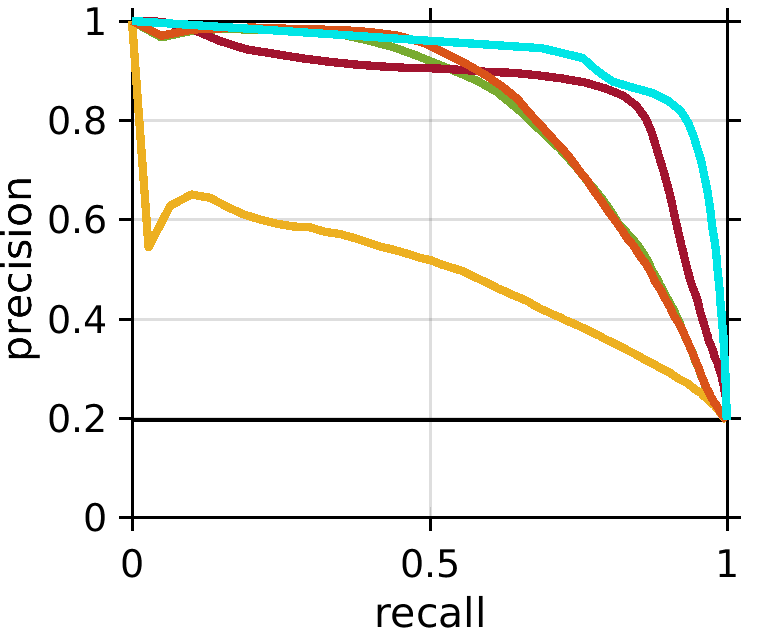}} 
    {\includegraphics[width = 4cm]{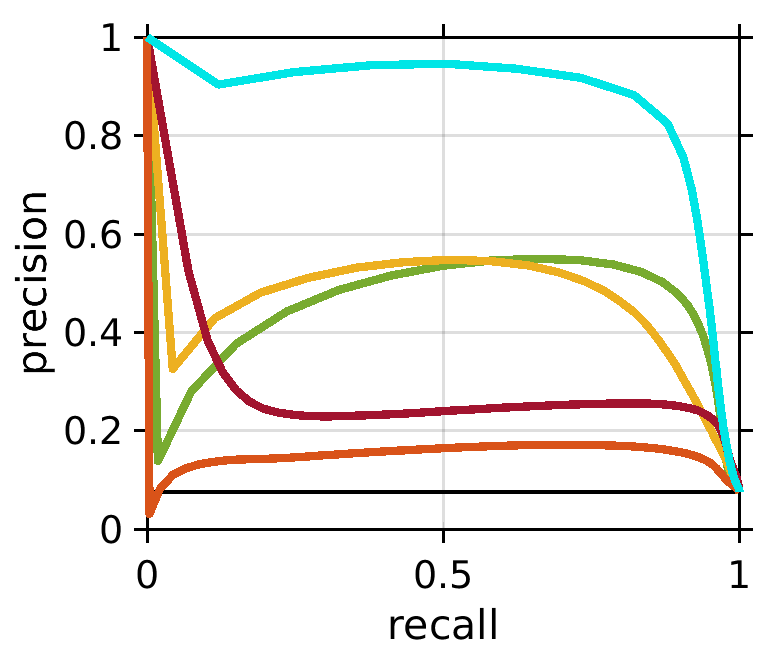}}
    {\includegraphics[width = 4cm]{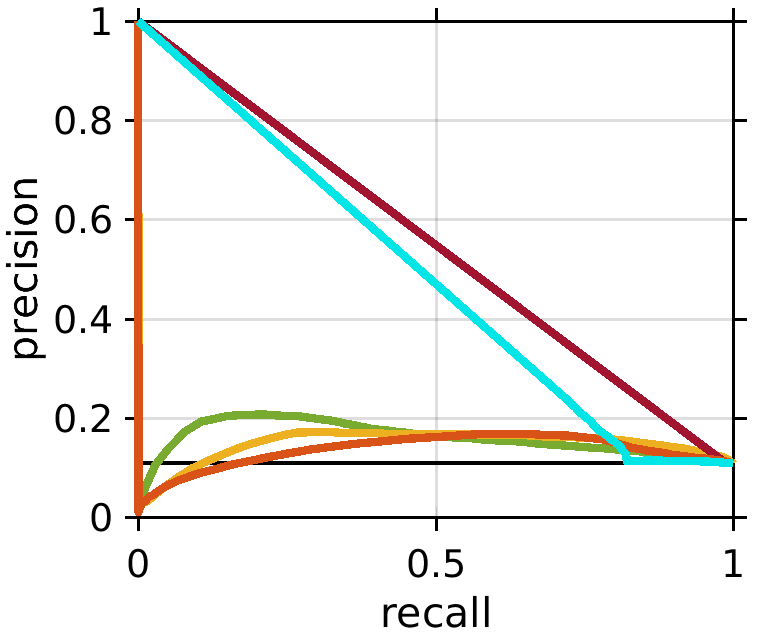}}
    {\includegraphics[width = 4cm]{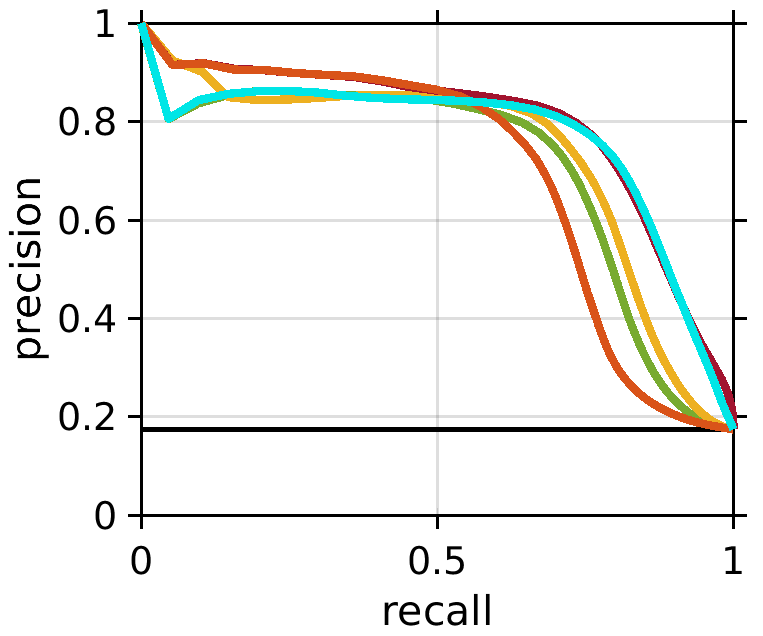}}
    
    \caption{ROC (top row) and Precision-Recall (bottom row) curves for change detection problems.}
    \label{fig:ROC_CD}
\end{figure*}

A visual comparison of the results in terms of activation maps for all methods is given in Fig.~\ref{fig:AD}. They display the predictions given to each sample. The prediction maps show a binary representation between change and non-change samples obtained from the model subject to a threshold. Results in all scenes demonstrate that (1) RX is a competitive method for detection, (2) KRX struggles to obtain reasonable results mainly due to the problem of hyperparameter tuning, (3) RBIG alone excels in all cases, while the hybrid approach (i.e. RX followed by RBIG) refines the results and yields clearer activation maps with sharper spatial detections.

Additionally, for a quantitative assessment of the results, it is customary to provide the ROC curves and to derive scores like the AUC from it. 
Figure~\ref{fig:ROC_AD} shows the ROC curves and Table~\ref{tab:AD} summarizes all AUC values for all images and methods. For each experiment, we performed 1000 runs for testing the significance of the methods based on the ROC profiles. The results are shown in Figure~\ref{fig:BOXPLOT_AD}. 
Although the RBIG model achieves good results, RX model is able to compete and achieve results as good as RBIG for some images. The HYBRID model is able to keep the properties of the above mentioned models obtaining results equal or better than any other method. While KRX obtains a reasonable performance in some images, it clearly fails in some situations like the Cat-Island image. The low standard deviations show that all methods but the KRX are clearly robust with a little bit bigger standard deviation for the RX method in most cases.

\subsection{Experiment 3: Real and Natural Changes}

This section reports an experiment to analyze the performance of the proposed methods in change detection problems. The database is composed of different scenes with natural changes, whose characteristics are summarized in Table~\ref{table:database}.


\subsubsection{Data collection}

We collected pairs of multispectral images in such a way that they coincide at the same spatial resolution but at different acquisition time, the images are co-registered. We selected the images in such a way that an anomalous change happened between the two acquisition times. We manually labeled all the images finding the changed pixels. Labeling considered avoiding shadows, changes in lighting and natural changes in vegetation which could compromise results evaluation. All images contain changes of different nature, which allows us to analyze and study how the algorithms perform in heterogeneous realistic scenarios. 
The Texas wildfire dataset is composed by a set of four images acquired by different sensors over Bastrop County, Texas (USA), and is composed by a Landsat 5 TM as the pre-event image and a Landsat 5 TM plus an EO-1 ALI and a Landsat 8 as post-event images. This phenomenon is considered the most destructive wildland-urban interface wildfire in Texas history and the interest region represent the 19.54\%. The Argentina image represents an area burned between the months of July and August 2016 due to the high temperatures in these crop areas, the change region representing the 7.5\% of the whole scene. 
The Chile dataset represents the Aculeo lake in central part of this country, which has now dried up completely. These images contrast the lake in 2014, when it still contained substantial water, and 2019, when it consisted of dried mud and green vegetation. Scientists attribute the lake's decline to an unusual decade-long drought, coupled with increased water consumption from a growing population, and the changed region represents a relevant 10.81\% of the whole scene. The last dataset labeled as Australia shows the natural floods caused by Cyclone Debbie in Australia 2017. Storm damage resulted from both the high winds associated with the cyclone, and the very heavy rain that produced major riverine floods. The change samples represent an important portion of the scene, the 17.35\% of pixels affected. Since our RBIG approach only takes the time $$t_1$$ image, these big changes do not have a critical impact on method's performance.

\subsubsection{Numerical and Visual Comparison}

Figure~\ref{fig:CD} shows the RGB composites of the pairs of images, the corresponding reference map and activation maps obtained. 
RBIG obtains clearly better results than the other methods in all cases; very good  performance in three out of the four scenarios and a clear advantage in the most difficult one (Chile image). 

When dealing with highly skewed datasets, PR curves give a more informative picture of an algorithm's performance compared to ROC. 
Figure \ref{fig:ROC_CD} shows both the ROC and the PR curves results for all methods and all the images. In all cases RBIG outperforms the other methods largely, thus suggesting  the suitability of adopting a more direct approach of density estimation in the change detection problems too. 
A summary of the AUC values of all methods and scenarios is shown in Table~\ref{tab:ACD}. The RBIG approach is to able to estimate the change samples with a high accuracy overtaking in 7\%, 3\%, 6\% and 5\% respectively with respect the second best method. 
{While AUC returns an overall measure of method's robustness, in (remote sensing) anomaly detection problems one typically cares  about the low false alarm rate regime. We study the performance of the methods looking at different false positive rates (FPRs) in Table~\ref{tab:FPRAUC}. The proposed RBIG approach consistently reports the best performance, especially in very low FPR regimes.}

\begin{table}[h!]
    \centering
     \caption{AUC results for Change Detection images. The best value for each image are in bold }
     \scriptsize
    \begin{tabular}{|l|l|l|l|l|l|}
    \hline\hline
    \rowcolor[HTML]{A9A9A9}{{\bf METHODS}} &{\bf RX}&{\bf K-RX}&{\bf K-SVDD}&{\bf KDE}&{\bf RBIG}\\ \hline\hline
        Texas   &\it{0.91} &\it{0.80} &\it{0.95}&\it{0.91}&{\bf 0.98}  \\ \hline
                
        Argentina &\it{0.94} &\it{0.93}&\it{0.87}&\it{0.77} &{\bf 0.97}\\ \hline
         
        Chile &\it{0.64} &\it{0.66}&\it{0.61}&\it{0.63} &{\bf 0.72}  \\ \hline
        Australia &\it{0.86} &\it{0.88} &{\bf 0.93} &\it{0.83} &{\bf 0.93}  \\ \hline
         
    \end{tabular}
   
    \label{tab:ACD}
\end{table}

\begin{table}[h!]
    \centering
     \caption{{Low false alarm rate regime. First row correspond to false positive rate (FPR) and columns represent the AUC values from rigorous to moderate FPR.}  }
     \scriptsize
    \begin{tabular}{|l|l|l|l|l|l|l|}
    \hline\hline
    \rowcolor[HTML]{A9A9A9}{{\bf FPR}} &{\bf 0.1}&{\bf 0.2}&{\bf 0.3}&{\bf 0.1}&{\bf 0.2}&{\bf 0.3}\\ \hline\hline
    \rowcolor[HTML]{D8D8D8} & \multicolumn{3}{|l|}{Texas} & \multicolumn{3}{|l|}{Argentina}\\ \hline \hline 

        {\bf RX}    &  \it{0.78} &\it{0.87} &\it{0.91} &  \it{0.93} &{\bf 0.96} &{\bf 0.97} \\ \hline
        {\bf KRX}   &  \it{0.46} &\it{0.65} &\it{0.76} &  \it{0.85} &\it{0.92} &\it{0.95} \\ \hline
        {\bf SVDD}  &  \it{0.90} &\it{0.93} &\it{0.95} &  \it{0.37} &\it{0.84} &{\bf 0.97} \\ \hline
        {\bf KDE}   &  \it{0.78} &\it{0.86} &\it{0.90}  &  \it{0.20} &\it{0.48} &\it{0.75} \\ \hline
         {\bf RBIG} &  {\bf 0.96 } &{\bf 0.98} &{\bf 0.99} &  {\bf 0.95 } &{\bf 0.96} &{\bf 0.97} \\ \hline          \hline
            
        \rowcolor[HTML]{D8D8D8} & \multicolumn{3}{|l|}{Chile} & \multicolumn{3}{|l|}{Australia} \\ \hline\hline
     
        {\bf RX}    &  \it{0.21} &\it{0.37} &\it{0.49} &  \it{0.76} &\it{0.81} &\it{0.83} \\ \hline
        {\bf KRX}   &  \it{0.09} &\it{0.34} &\it{0.50} &  \it{0.79} &\it{0.83} &\it{0.86} \\ \hline
        {\bf SVDD}  &  \it{0.07} &\it{0.08} &\it{0.45} &  {\bf 0.84} &{\bf 0.89} &{\bf 0.93} \\ \hline
        {\bf KDE}   &  \it{0.06} &\it{0.23} &\it{0.47} &  \it{0.71} &\it{0.76} &\it{0.78} \\ \hline
         {\bf RBIG} &  {\bf 0.38} &{\bf 0.72} &{\bf 0.73} &  {\bf 0.84} &{\bf 0.89} &{\bf 0.93} \\ \hline \hline
    \end{tabular}
      
    \label{tab:FPRAUC}
\end{table}

\section{Conclusions}\label{sec:conclusions}
We introduced a novel detector based on multivariate Gaussianization. The methodology copes with anomaly and change detection problems in remote sensing image processing, and meets all requirements of the problems: is an unsupervised method with no parameters to fit, it can deal with large amount of data, and it is more accurate to competing approaches. The model assumption is based on detecting anomalies by estimating probabilities of pixels.
The proposed method excelled quantitatively (AUC, ROC and PR curves) and qualitative based on visual inspection over the rest of the implementations, in both anomaly and change detection. The evaluation considered a wide range of remote sensing images, in a  diversity of problems, dimensionality and number of examples. We also suggested a hybrid approach where the Gaussianization method is applied after a regular anomaly detector: this facilitates the density estimation and improves the results notably. 
Future work will consider 
exploiting the information-theoretic properties of RBIG \cite{laparra2020information} which opens alternatives to identify changes in image time series.


\newpage
 \begin{wrapfigure}{l}{25mm} 
    \includegraphics[width=1in,height=1.25in,clip,keepaspectratio]{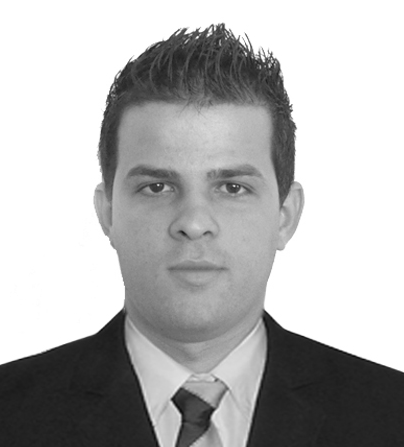}
  \end{wrapfigure}\par
  
\textbf{Jos\'e A. Padr\'on Hidalgo} received the B.Sc degree in Telecommunications and Electronics from the University of Pinar del R\'io, Cuba, in 2015. He received the Ph.D. degree in electronics engineering from the Universitat de Val\`encia, Valencia, in 2021.  

His current research interest includes developing algorithms in order to detect anomaly and anomalous changes for Remote Sensing imagery with the Image and Signal Processing Group with the Universitat de Val\`encia.

 \begin{wrapfigure}{l}{25mm} 
    \includegraphics[width=1in,height=1.25in,clip,keepaspectratio]{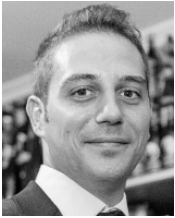}
  \end{wrapfigure}\par
 
\textbf{Valero Laparra} was born in Valencia, Spain, in 1983. He received the B.Sc degree in telecommunications engineering and the B.Sc. degree in the electronics engineering from the Universitat de Val\`encia, Valencia, in 2005 and 2007, respectively, the B.Sc. degree in mathematics from the Universidad Nacional de Educaci\'on a Distancia, Madrid, Spain, in 2010, and the Ph.D. degree in computer science and mathematics from the Universitat de Val\`encia, in 2011. He is currently an Assistant Professor with the Escuela T\'ecnica Superior de Ingener\'ia, Universitat de Val\`encia, where he is also a Researcher with the Image Processing Laboratory

 \begin{wrapfigure}{l}{25mm} 
    \includegraphics[width=1in,height=1.25in,clip,keepaspectratio]{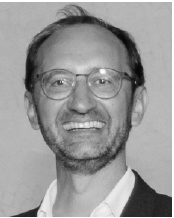}
  \end{wrapfigure}\par
 
\textbf{Gustau Camps-Valls }(IEEE Fellow'18, ELLIS Fellow, IEEE Distinguished lecturer, PhD in Physics) is a Full professor in Electrical Engineering and head of the Image and Signal Processing (ISP) group, http://isp.uv.es, at the Universitat de Val\`encia. He is interested in the development of machine learning algorithms for geosciences and remote sensing data analysis. He is an author of around 250 journal papers, more than 300 conference papers, 25 international book chapters, and editor of 6 books on kernel methods and deep learning. He holds a Hirsch's index h=75 (Google Scholar), entered the ISI list of Highly Cited Researchers in 2011, and Thomson Reuters ScienceWatch identified one of his papers on kernel-based analysis of hyperspectral images as a Fast Moving Front research. He received two European Research Council (ERC) grants: an ERC Consolidator grant on "Statistical learning for Earth observation data analysis" (2015) and an ERC Synergy grant on "Understanding and Modelling the Earth system with machine learning" (2019). In 2016 he was included in the prestigious IEEE Distinguished Lecturer program of the GRSS.

\end{document}